\documentclass{article}

\usepackage[final,nonatbib]{neurips_data_2023}

\usepackage[utf8]{inputenc} 
\usepackage[T1]{fontenc}    
\usepackage{hyperref}       
\usepackage{url}            
\usepackage{booktabs}       
\usepackage{amsfonts}       
\usepackage{nicefrac}       
\usepackage{microtype}      
\usepackage{xcolor}         
\usepackage{graphicx}
\usepackage{amsmath}
\usepackage{pifont}
\newcommand{\cmark}{\ding{51}}
\newcommand{\xmark}{\ding{55}}

\usepackage{pbox}
\newcommand{\specialcell}[2][c]{%
  \begin{tabular}[#1]{@{}c@{}}#2\end{tabular}}

\newcommand{\ie}{\textit{i}.\textit{e}.}
\newcommand{\eg}{\textit{e}.\textit{g}.}

\title{FORB: A Flat Object Retrieval Benchmark for Universal Image Embedding}

\author{%
  Pengxiang Wu, Siman Wang, Kevin Dela Rosa, Derek Hao Hu \\
  Snap Inc.\\
  \texttt{\{pwu,swang7,kevin.delarosa,hao.hu\}@snap.com} \\
}

\begin{document}

\maketitle

\begin{abstract}
  Image retrieval is a fundamental task in computer vision. Despite recent advances in this field, many techniques have been evaluated on a limited number of domains, with a small number of instance categories.
  Notably, most existing works only consider domains like 3D landmarks, making it difficult to generalize the conclusions made by these works to other domains, \eg, logo and other 2D flat objects.
  To bridge this gap, we introduce a new dataset for benchmarking visual search methods on flat images with diverse patterns. Our flat object retrieval benchmark (FORB) supplements the commonly adopted 3D object domain, and more importantly, it serves as a testbed for assessing the image embedding quality on out-of-distribution domains.
  In this benchmark we investigate the retrieval accuracy of representative methods in terms of candidate ranks, as well as matching score margin, a viewpoint which is largely ignored by many works.
  Our experiments not only highlight the challenges and rich heterogeneity of FORB, but also reveal the hidden properties of different retrieval strategies.
  The proposed benchmark is a growing project and we expect to expand in both quantity and variety of objects.
  The dataset and supporting codes are available at https://github.com/pxiangwu/FORB/.
\end{abstract}

\section{Introduction}
\label{sec:intro}

Image retrieval is a fundamental and long-standing task in computer vision. Given a query image, this task aims to search for the most similar images from a large database.
Recent methods have achieved remarkable performance on certain domains, such as 3D landmark \cite{revisited_oxford_paris,gldv2} and clothes \cite{liu2016deepfashion}.
To perform image retrieval, the prevailing practice is to map the query image into a compact embedding space, where similar images are close to each other while dissimilar ones are separated away.
This embedding space can be handcrafted and one classic design is the Bag of Words (BoW) \cite{rootsift,bow}. A more effective idea is to learn the embedding automatically, based on deep neural netowrks \cite{gordo2016deep,babenko2015aggregating,delf,radenovic2018fine}.
However, all these methods have only been evaluated on a limited number of domains (\eg, 3D landmarks), and as a result, it remains unclear if the embedding of one method is more general than the others. In particular, for learning-based methods, since they are usually trained on a specific restricted dataset with limited object classes (\eg, ImageNet \cite{deng2009imagenet} and Open Images \cite{OpenImages}), their feature embeddings could be not universal enough to generalize to various open-world objects. Therefore, it is necessary to have benchmarks supplementary to the existing ones for a more comprehensive evaluation of the embeddings, especially in terms of their out-of-distribution (OOD) generalization ability. 

In particular, existing image retrieval benchmarks mainly involve domains of 3D objects. 
Examples of the commonly considered objects include 3D landmarks, clothes, natural living things and online products.
While many recent benchmarks that curate the images of these objects have sufficiently large query image sets, they are typically limited to a small number of object categories or instances. 
Moving beyond 3D objects, there are several datasets focusing on 2D flat objects. However, these datasets are mostly small in size and related to one particular type of object, \ie, logo \cite{FlickrLogos,opensetlogo}. Besides, their query images tend to be in canonical pose without much distraction from the background, making the retrieval less challenging.

\begin{figure}[t!]
	\begin{center}
		\begin{minipage}[b]{.15\linewidth}
			\centering
			\includegraphics[width=\textwidth]{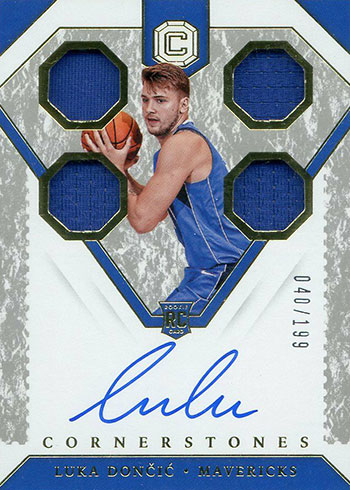}
			\scriptsize
			(a) Database
		\end{minipage}
            \begin{minipage}[b]{.21\linewidth}
			\centering
			\includegraphics[width=\textwidth]{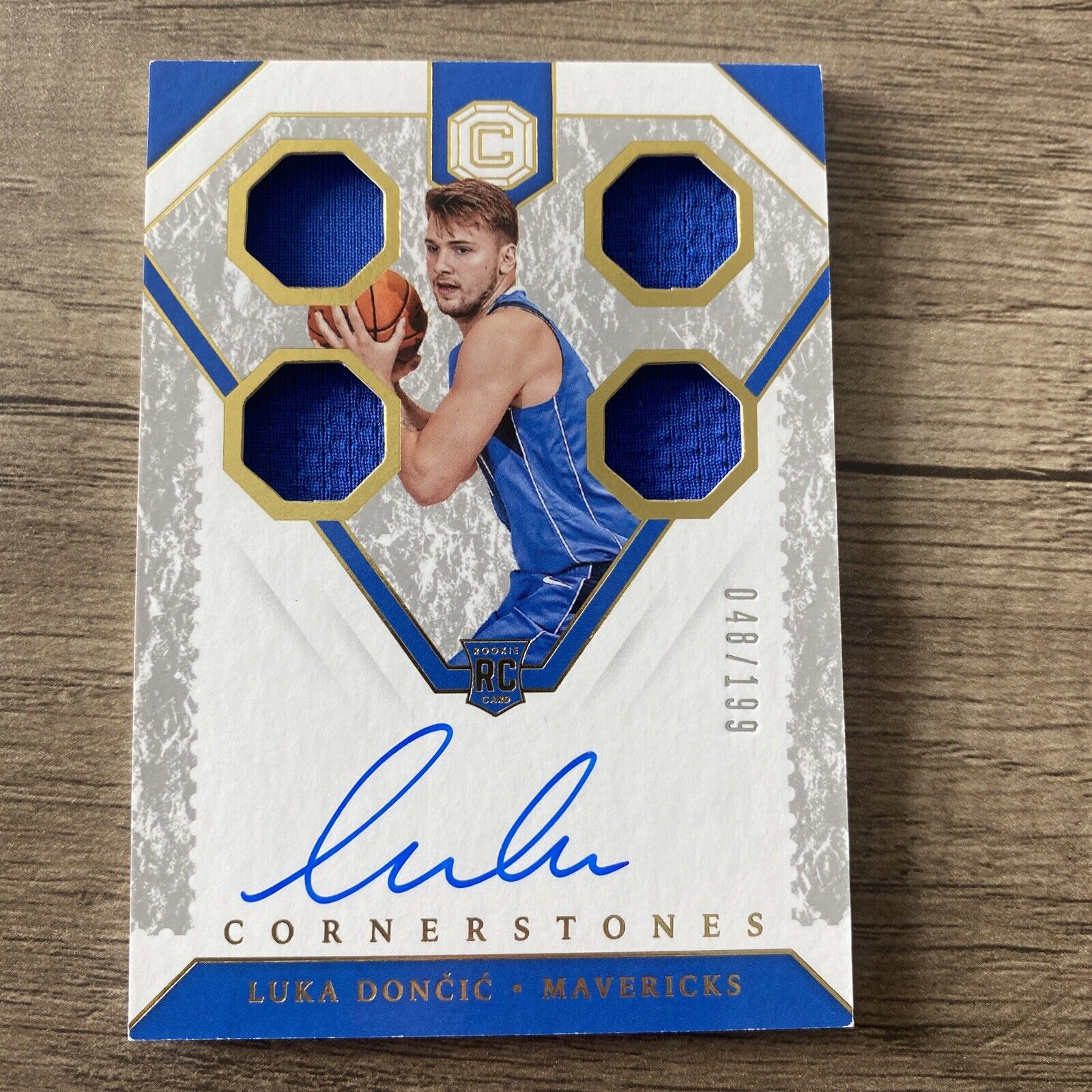}
			\scriptsize
			(b) Query (easy)
		\end{minipage}
            \begin{minipage}[b]{.1445\linewidth}
			\centering
			\includegraphics[width=\textwidth]{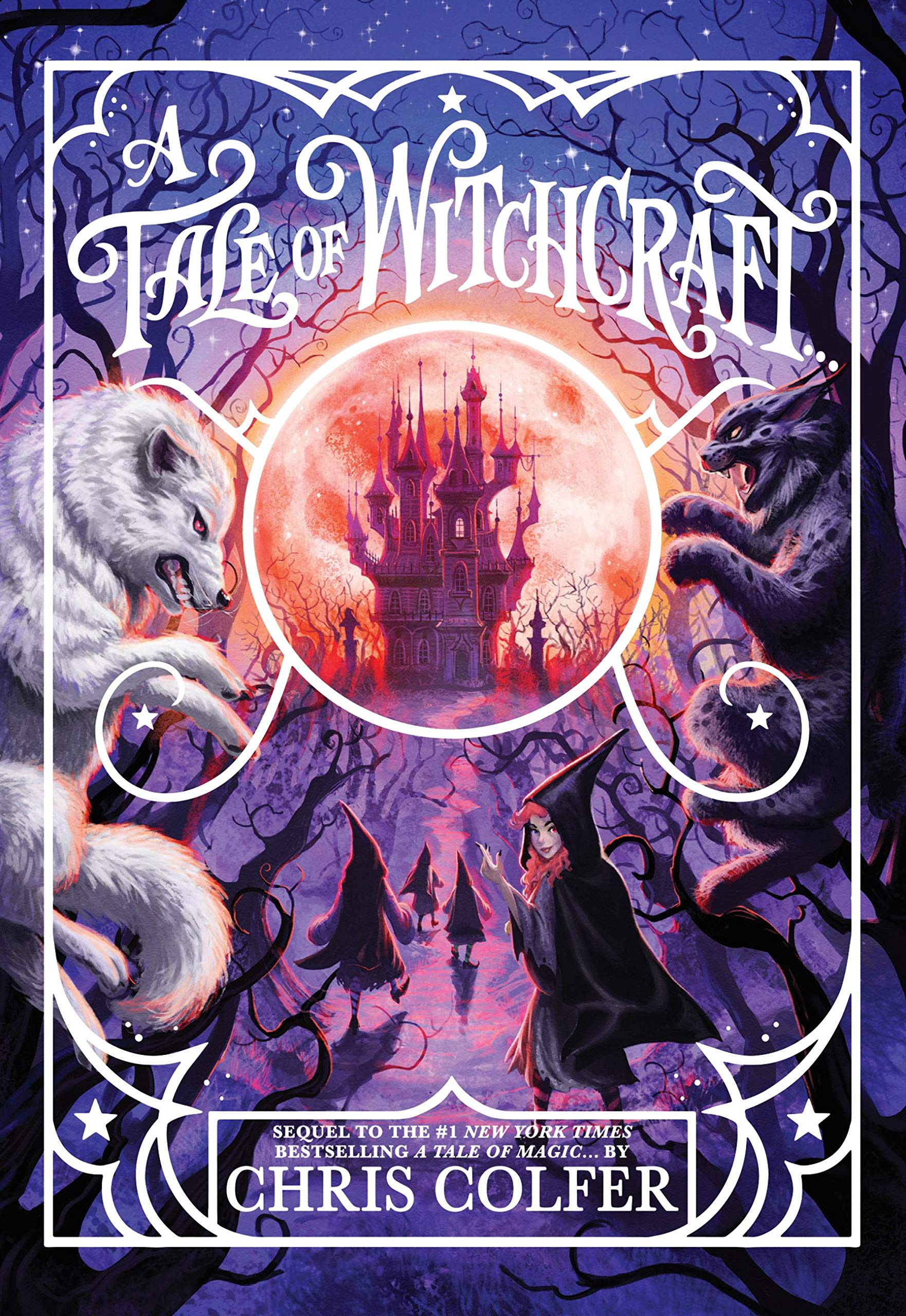}
			\scriptsize
			(c) Database
		\end{minipage}
            \begin{minipage}[b]{.1682\linewidth}
			\centering
			\includegraphics[width=\textwidth]{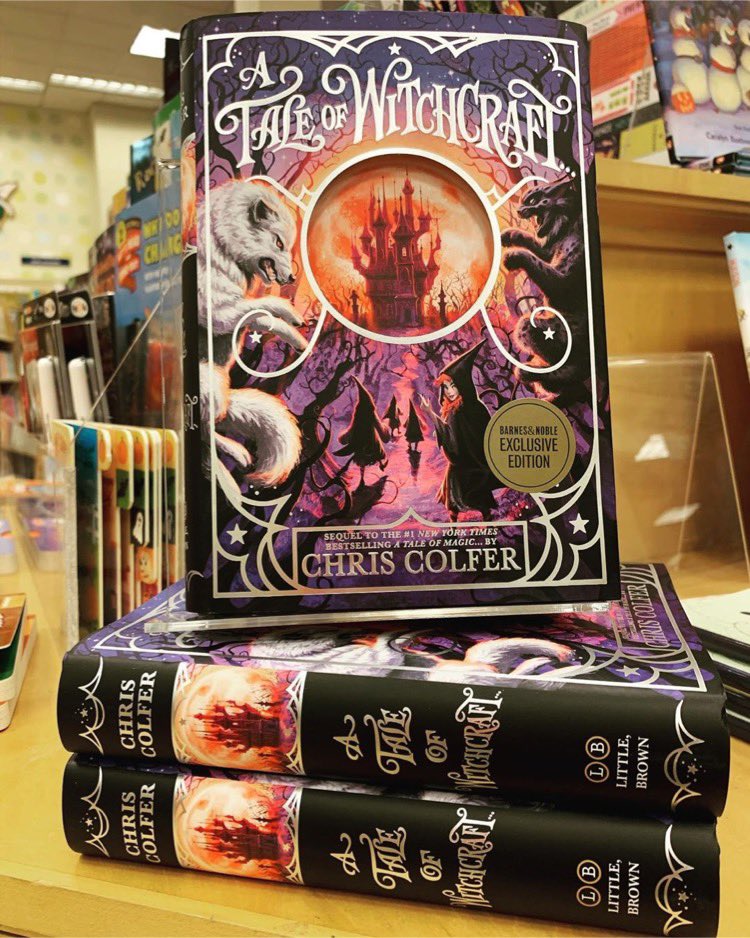}
			\scriptsize
			(d) Query (Medium)
		\end{minipage}
            \begin{minipage}[b]{.1115\linewidth}
			\centering
			\includegraphics[width=\textwidth]{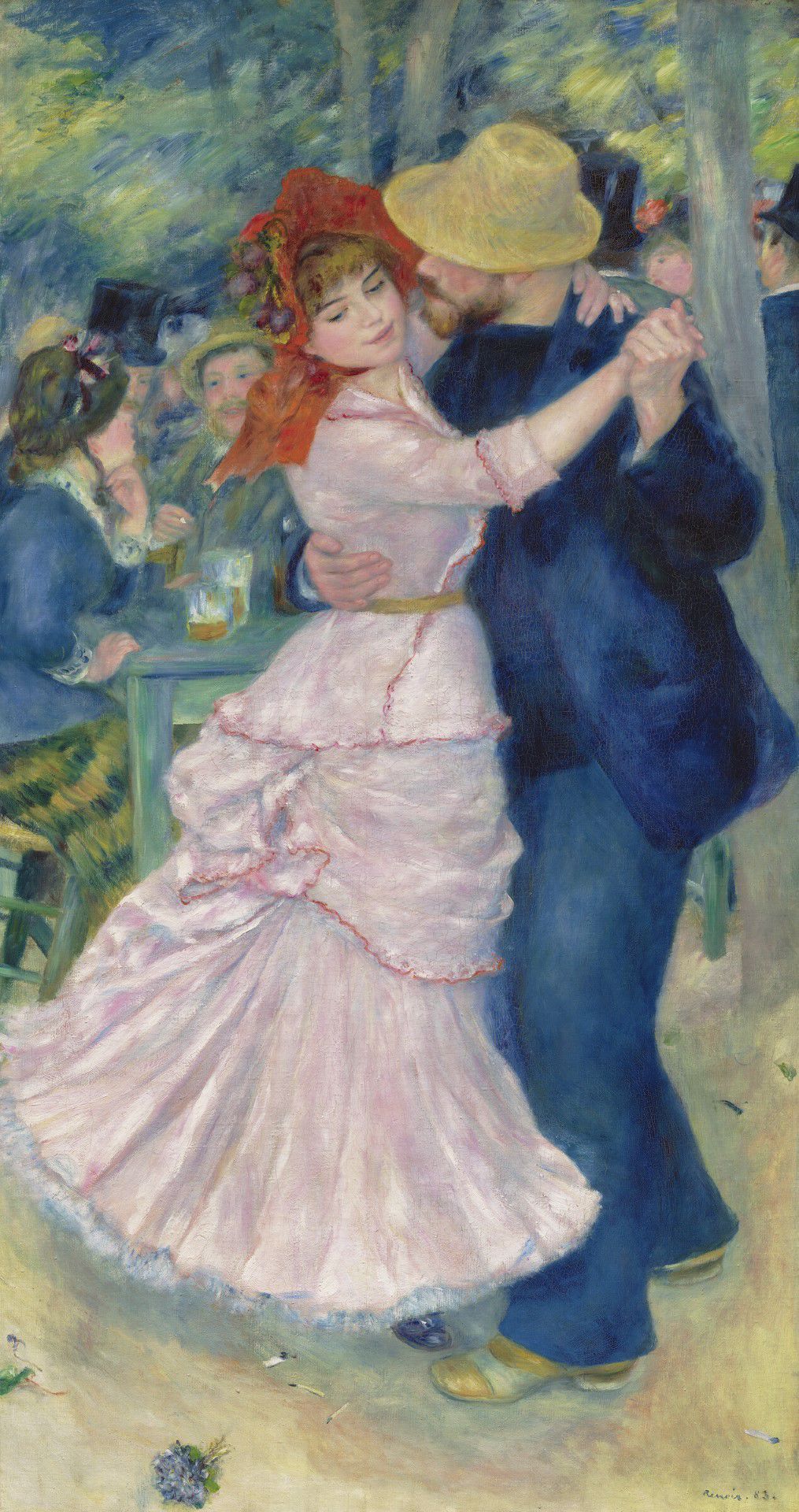}
			\scriptsize
			(e) Database
		\end{minipage}
            \begin{minipage}[b]{.169\linewidth}
			\centering
			\includegraphics[width=\textwidth]{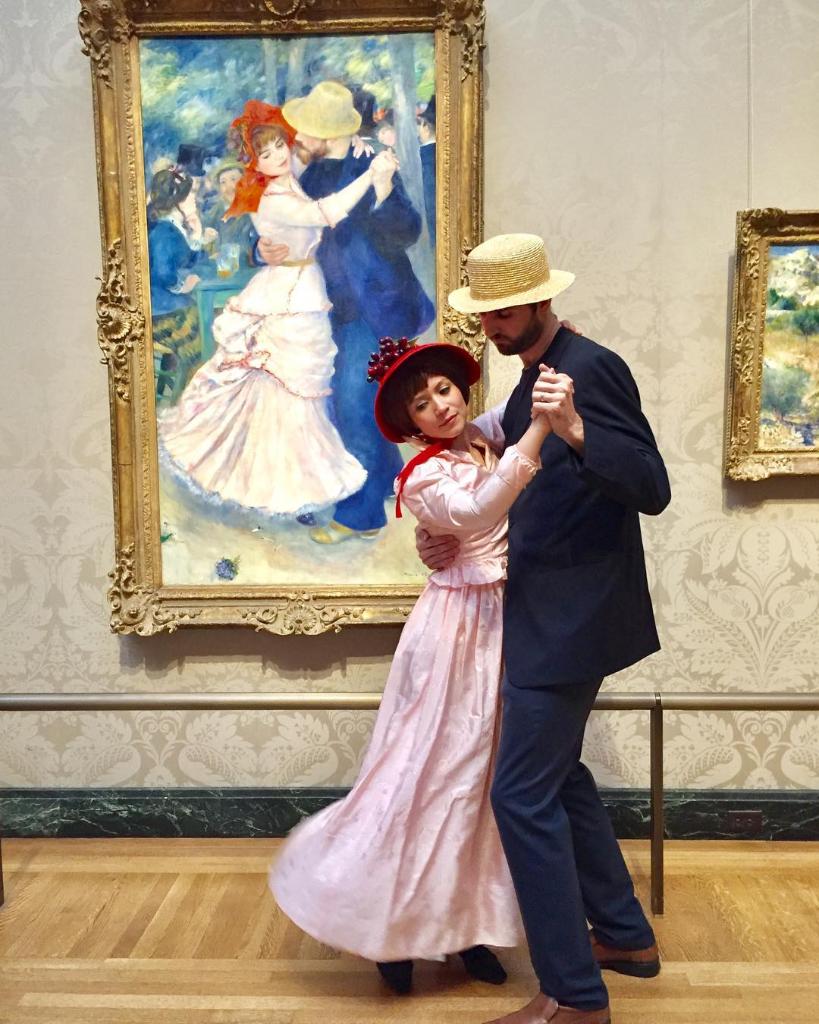}
			\scriptsize
			(f) Query (medium)
		\end{minipage}
            \\
            \vspace{0.1cm}
            \begin{minipage}[b]{.143\linewidth}
			\centering
			\includegraphics[width=\textwidth]{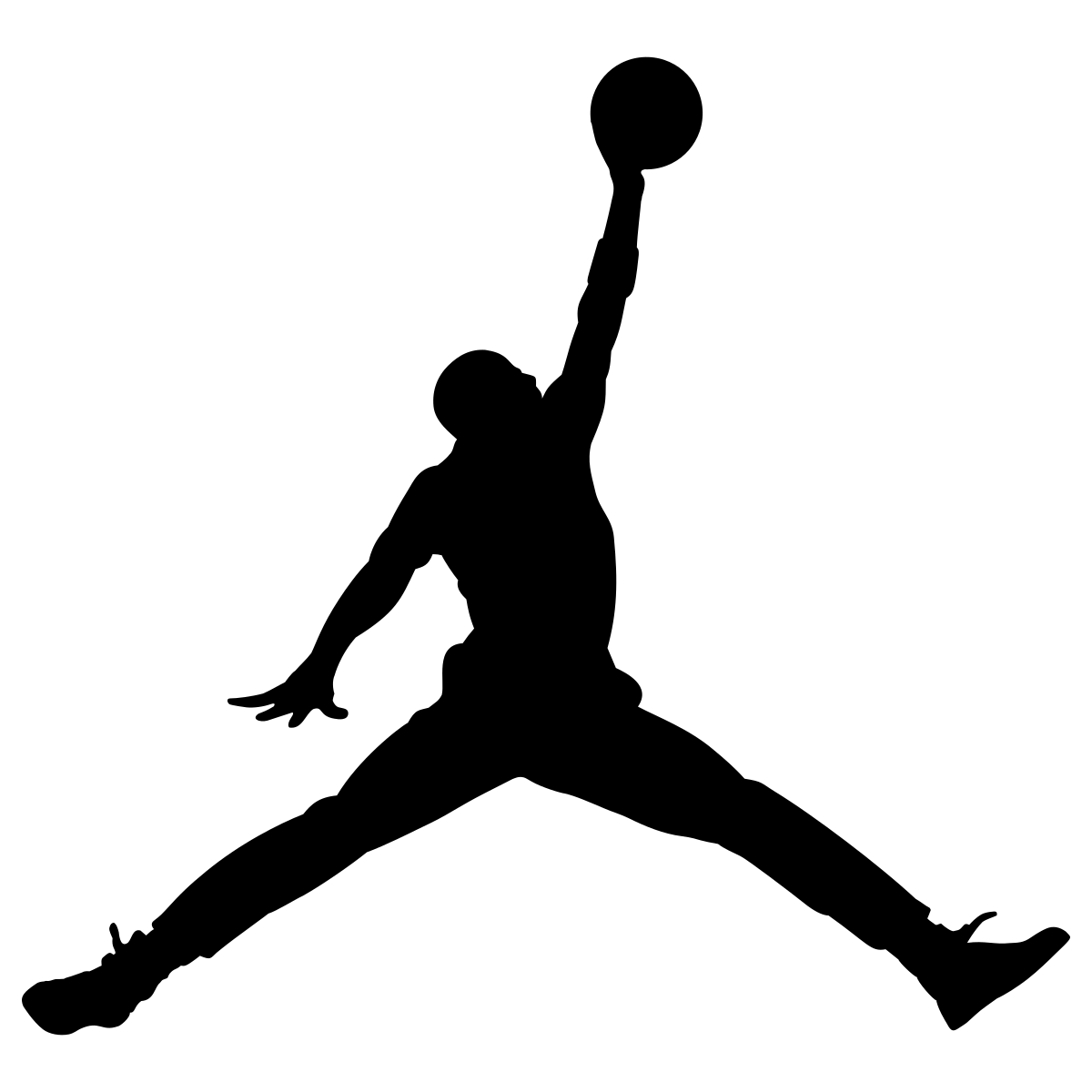}
			\scriptsize
			(g) Database
		\end{minipage}
            \begin{minipage}[b]{.177\linewidth}
			\centering
			\includegraphics[width=\textwidth]{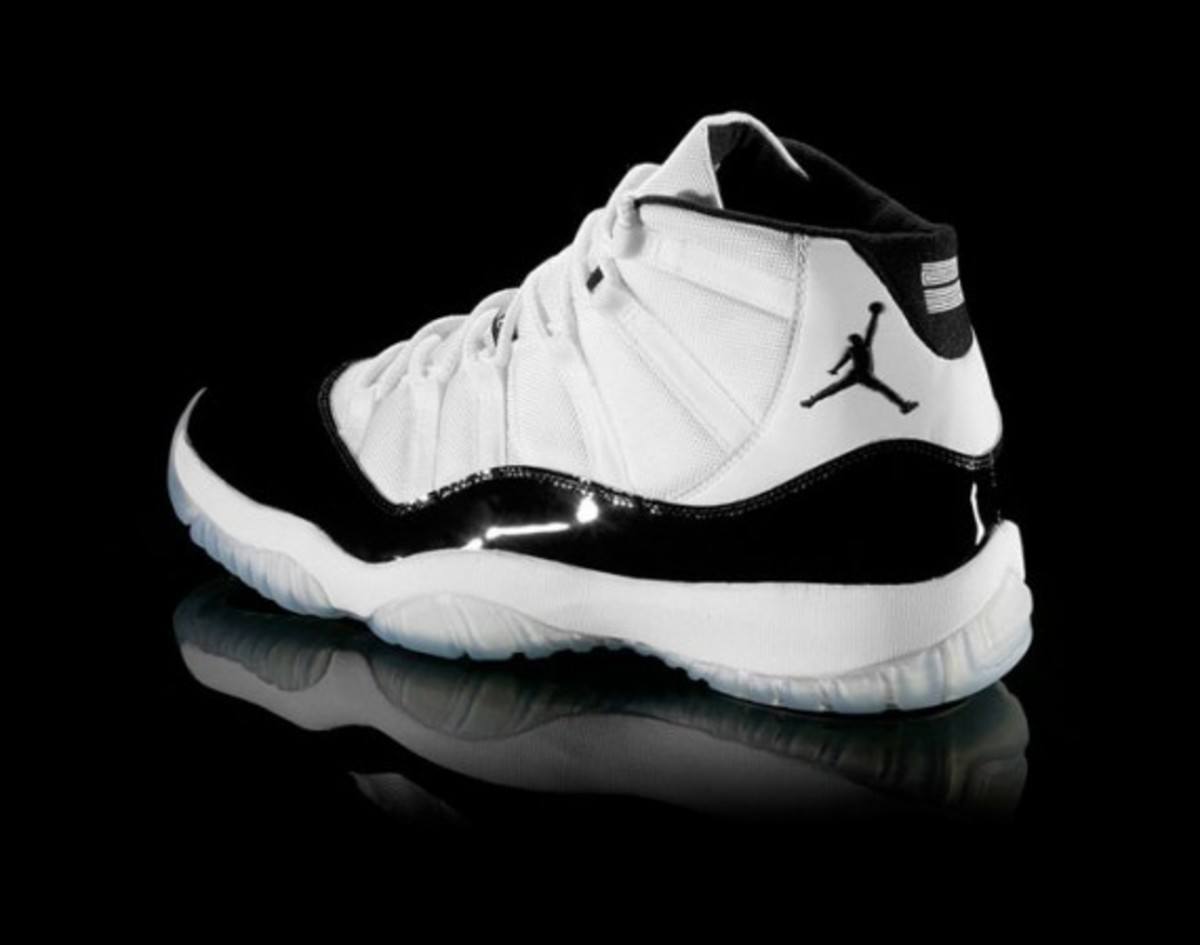}
			\scriptsize
			(h) Query (hard)
		\end{minipage}
            \begin{minipage}[b]{.199\linewidth}
			\centering
			\includegraphics[width=\textwidth]{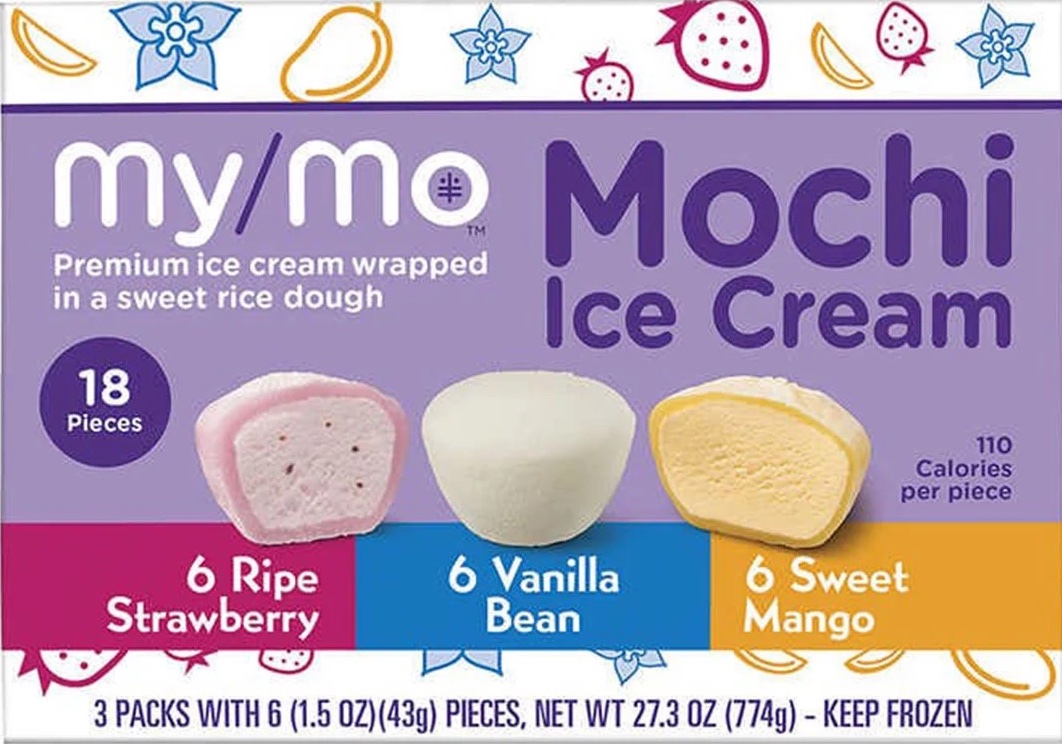}
			\scriptsize
			(i) Database
		\end{minipage}
            \begin{minipage}[b]{.185\linewidth}
			\centering
			\includegraphics[width=\textwidth]{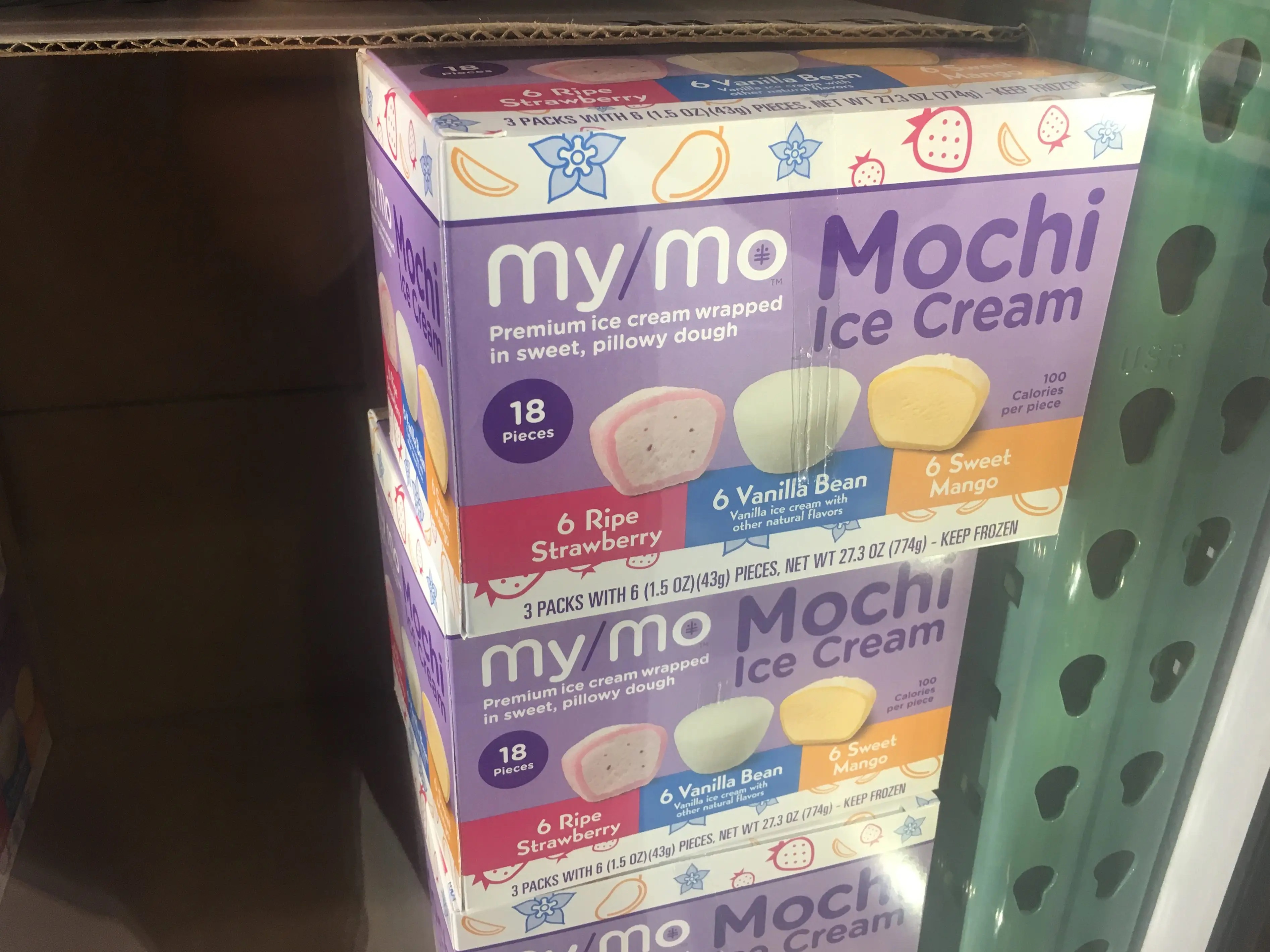}
			\scriptsize
			(j) Query (medium)
		\end{minipage}
            \begin{minipage}[b]{.0962\linewidth}
			\centering
			\includegraphics[width=\textwidth]{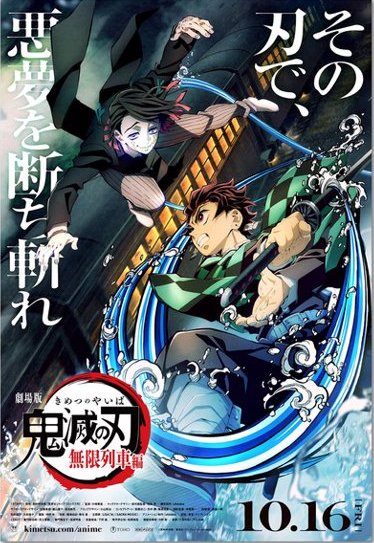}
			\scriptsize
			(k) Database
		\end{minipage}
            \begin{minipage}[b]{.152\linewidth}
			\centering
			\includegraphics[width=\textwidth]{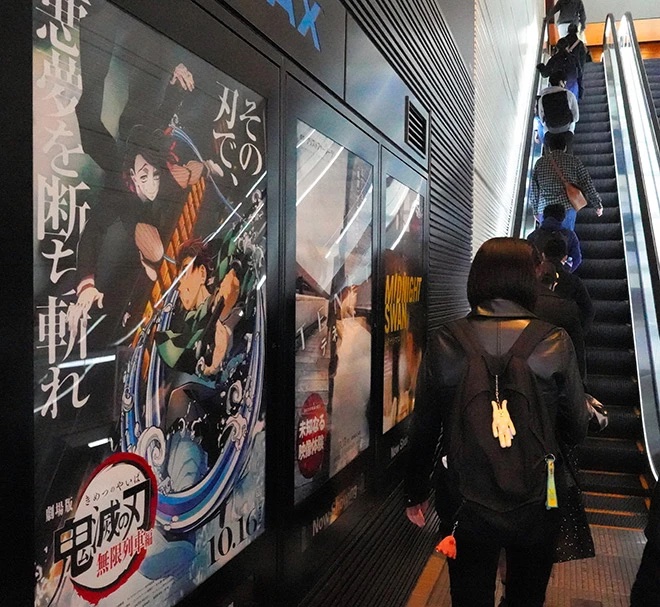}
			\scriptsize
			(l) Query (medium)
		\end{minipage}
            \\
            \vspace{0.1cm}
            \begin{minipage}[b]{.137\linewidth}
			\centering
			\includegraphics[width=\textwidth]{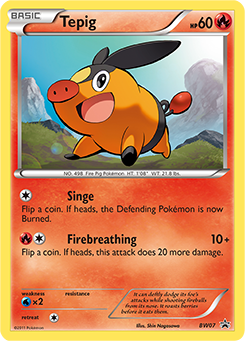}
			\scriptsize
			(m) Database
		\end{minipage}
            \begin{minipage}[b]{.338\linewidth}
			\centering
			\includegraphics[width=\textwidth]{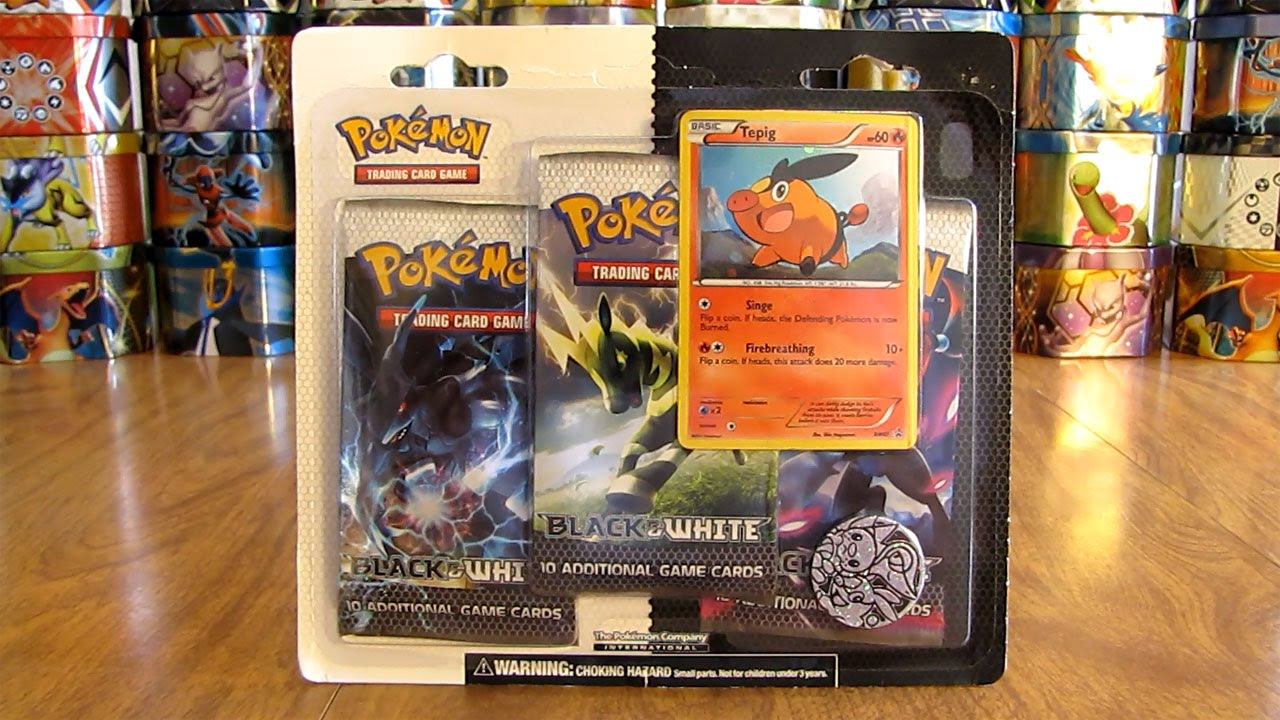}
			\scriptsize
			(n) Query (hard)
		\end{minipage}
            \begin{minipage}[b]{.088\linewidth}
			\centering
			\includegraphics[width=\textwidth]{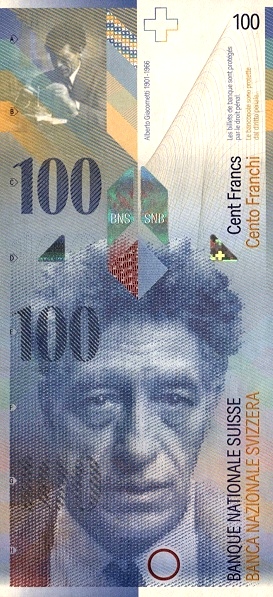}
			\scriptsize
			(o) Database
		\end{minipage}
            \begin{minipage}[b]{.4\linewidth}
			\centering
			\includegraphics[width=\textwidth]{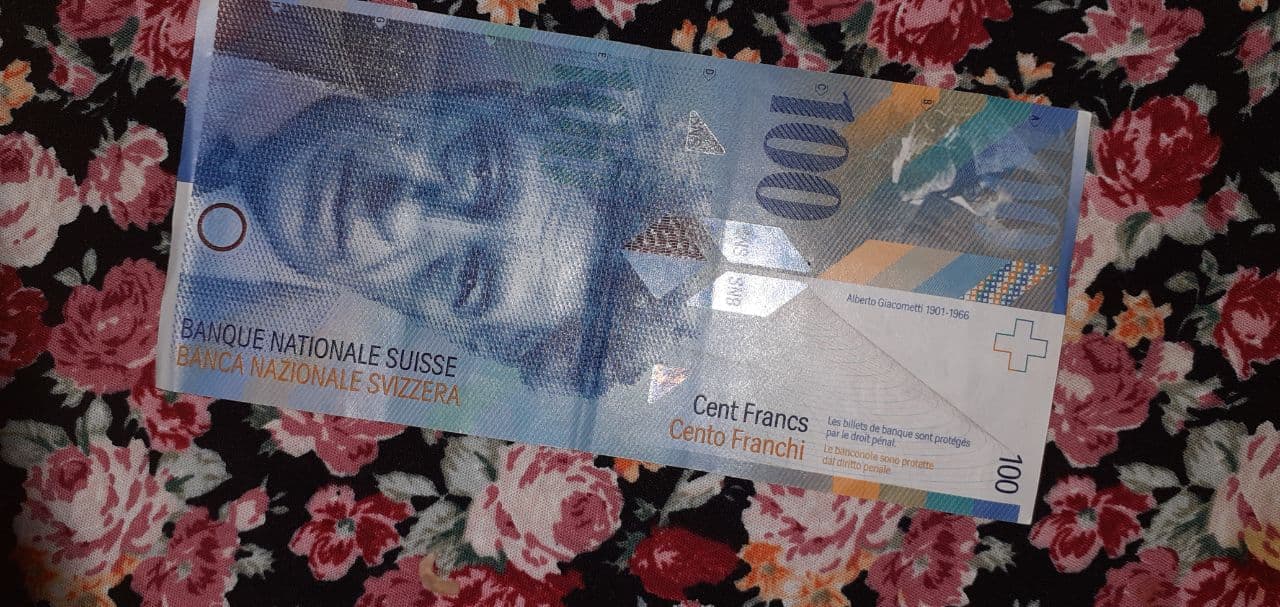}
			\scriptsize
			(p) Query (medium)
		\end{minipage}
	\end{center}
	\caption{Example database and query images from our FORB benchmark. For each query image, we show its corresponding index image and the retrieval difficulty. The images are from different content domains: (a)(b) photorealistic trading card; (c)(d) book cover; (e)(f) painting; (g)(h) logo; (i)(j) packaged goods; (k)(l) movie poster; (m)(n) animated trading card; (o)(p) currency.
    }
	\label{fig:eg_images}
        \vspace{-0.1cm}
\end{figure}

In order to fill the domain gap of existing benchmarks and to encourage future research in this area, we present a Flat Object Retrieval Benchmark (FORB) which contains diverse flat objects with different query difficulties. The flat objects are those with 2D surface only, which bears the textures and patterns of the object (\eg, painting and logo; see Figure~\ref{fig:eg_images}). Despite being one dimension less than 3D objects, such flat surfaces still pose many challenges for image retrieval. In particular, there can be large variations between the query and database images, due to surface and color distortions, perspective transformation, view occlusion, and illumination change. Our benchmark takes into account all these challenges and covers objects with a variety of textures (see Section~\ref{sec:dataset}). Notably, these objects are common in daily life and our benchmark could benefit diverse real-world visual search applications, such as recognizing logos for brand promotion, augmenting artwork exhibits in a museum, online shopping and more.

To understand how different image embeddings perform on our benchmark, we evaluate the retrieval accuracy from two perspectives: 
(1) Candidate rank, which corresponds to the sorted order of database images based on their similarities to the query image. The correctness of ranks
reflects the discriminative ability of image embedding and can be measured with mean Average Precision (mAP). (2) Matching score margin.
For a query image, ideally its matching scores against ground-truth database images should be high (\eg, assuming cosine similarity), while the scores against non-relevant images should be low. Therefore, the degree of compliance with this ideal margin also delineates the quality of image embedding, a viewpoint which is largely ignored by previous works. To measure this margin, we propose to query the given image against distractor images, giving \textit{false positive} candidates. By thresholding the matching scores, we can compute a specific false positive rate (FPR) and an updated mAP, which together quantify the margin of image embeddings. In particular, an ideal embedding should have a low FPR while keeping a high mAP.

To establish baselines on our benchmark, we evaluate a series of representative methods, including both learning-based models and handcrafted designs. Our results reveal intriguing properties of embeddings built from different feature levels. Specifically, we show that even a model is trained on 3D objects, its embedding induced from low- or mid-level image features can still be universal enough to distinguish diverse flat objects. Moreover, for feature-scarce images, embeddings based on high-level features tend to achieve better accuracy.

Our contributions include: (1) We introduce FORB, a new visual search benchmark for evaluating image embeddings on flat objects. FORB supplements the commonly used 3D object benchmarks and essentially provides a platform for assessing the OOD generalization ability of an embedding method. 
(2) We propose a new evaluation metric motivated by matching score margin. This metric is complementary to mAP and offers a new perspective on image embedding quality.
(3) We conduct comprehensive comparisons for different representative methods, providing solid baselines for future method developments.
(4) Our evaluation results reveal the hidden properties of different retrieval strategies as well as their limitations, providing insights into the development of new techniques.

\begin{table}[t!]
    \caption{Comparison of our benchmark against existing image retrieval datasets.}
    \label{tab:related_work}
    \resizebox{1.0\columnwidth}{!}{
        \centering
        \begin{tabular}{l|ccccc }
        \hline
        Dataset & Domain & \# Query & \# Database & Has distractor & Has difficulty label \\
        \hline
        Oxford \cite{oxford} & 3D landmark & 55 & 5K & \xmark & \xmark \\
        Paris \cite{paris} & 3D landmark & 55 & 6K & \xmark & \xmark \\
        $\mathcal{R}$-Oxford \cite{revisited_oxford_paris} & 3D landmark & 70 & 5K + 1M & \cmark & \cmark  \\
        $\mathcal{R}$-Paris \cite{revisited_oxford_paris} & 3D landmark & 70 & 5K + 1M & \cmark & \cmark\\
        GLD \cite{gld} & 3D landmark & 118K & 1.1M & \cmark & \xmark \\
        GLDv2 \cite{gldv2} & 3D landmark & 118K & 762K & \cmark & \xmark \\
        CUB \cite{welinder2010caltech} & Bird & 6K & 6K & \xmark & \xmark \\
        Cars196 \cite{krause20133d} & Car & 8K & 8K & \xmark & \xmark \\
        SOP \cite{sop} & 3D product & 60K & 60K & \xmark & \xmark \\
        DeepFashion \cite{liu2016deepfashion} & Clothes & 14K & 13K & \xmark & \xmark \\
        VehicleID \cite{liu2016deepvehicle} & Vehicle & 35.6K & 4.8K & \xmark & \xmark \\
        iNaturalist \cite{van2018inaturalist} & Plant \& Animal & 136K & 136K & \xmark & \xmark \\
        \hline
        FlickrLogos \cite{FlickrLogos} & Flat object (logo) & 4K & 320 & \cmark & \xmark \\
        \hline
        FORB & Flat object & 14K & 54K & \cmark & \cmark \\
        \hline
        \end{tabular}
    }
\end{table}

\section{Related Work}
\label{sec:related_work}

\subsection{Existing Datasets and Benchmarks for Image Retrieval}

There has been a long history of developing benchmarks for image retrieval. For example, to promote research in instance-level recognition and search, Oxford \cite{oxford} and Paris \cite{paris} datasets were introduced and have motivated a wealth of innovations in this field. With a similar motivation, researchers curated CUB \cite{welinder2010caltech} and Cars196 \cite{krause20133d} to facilitate fine-grained object matching.
Despite the popularity of these datasets, they are small in size and only involve a limited number of instances and categories. To further enrich the object domains for image retrieval and increase the size and complexity of the task, several more challenging datasets were constructed, such as SOP \cite{sop}, DeepFashion \cite{liu2016deepfashion}, VehicleID \cite{liu2016deepvehicle}, iNaturalist \cite{van2018inaturalist}, and Google Landmarks dataset v2 (GLDv2) \cite{gldv2}. In particular, GLDv2 has gained widespread attention since being introduced due to its significant scale and variability, and serves as a solid benchmark for testing emerging retrieval techniques.

One limitation of these datasets is that they only focus on the task of 3D object retrieval, involving a restricted number of object domains (\eg, 3D landmarks). In fact, compared to 3D objects, there exist few benchmarks on other domains, especially 2D flat objects. In real-world visual search applications, flat objects also make up a large fraction of queries. 
However, there are only few benchmarks on such objects and most of them are for logo \cite{FlickrLogos,opensetlogo}.
To fill this domain gap, our FORB benchmark includes a variety of flat objects and supplements existing 3D object benchmarks. In particular, FORB effectively serves as an OOD query set for evaluating the embeddings trained on 3D objects. In Table~\ref{tab:related_work} we compare  FORB against existing image retrieval datasets in detail.

It is worth mentioning that there exists another similar benchmark for assessing the generalization abilities of image embeddings, \ie, Google Universal Image Embedding Challenge\footnote{https://www.kaggle.com/competitions/google-universal-image-embedding}. However, this benchmark mainly involves 3D objects and its evaluation data is kept private. We believe our FORB supplements this benchmark and will facilitate the development of visual search applications, such as organizing photo collections, visual commerce and more.

\subsection{Out-of-Distribution Query}

Most existing benchmarks only have ``on-topic'' queries without considering the out-of-distribution ones. As a result, they fail to present real-world challenges and are not enough to fully evaluate the quality of an image embedding. Notably, in a generic visual search app, the system tends to be queried with a large number of irrelevant queries, \ie, OOD queries, for which it is expected to not yield any results. Therefore, OOD queries provide an additional important view into the robustness of image embeddings.
This issue of lacking OOD queries in existing benchmarks was recognized in GLDv2 \cite{gldv2} and addressed with plenty of non-landmark queries. In practice, to assess the discriminative ability of image embeddings between true positive and false positive candidates, GLDv2 employs micro Average Precision ($\mu$AP), which both measures ranking performance and penalizes false positive predictions. Our FORB benchmark shares a similar motivation to GLDv2, but with a few key differences: (1) We do not provide additional OOD queries with respect to the database images. Instead, we split database into index images and distractors, and query the images against distractors. In this way we effectively turn all the query images into OOD queries. (2) Instead of using $\mu$AP, we propose a new metric, $t$-mAP, which computes an averaged mAP over different confidence thresholds. The thresholds are determined through quantiles of false positive rates. Compared to $\mu$AP and mAP, our $t$-mAP takes into account the matching score margin, which directly reflects the discriminability of image embeddings.

\subsection{Universal Image Embedding}

The quality of image embeddings determines the performance of modern image retrieval methods. 
Based on the design of image features, existing embeddings can be divided into two categories: handcrafted and learning-based. The former one builds image embeddings based on handcrafted low-level features (\eg, SIFT \cite{lowe2004distinctive}), using a bag of words (BoW). This design paradigm dominates many classic methods, such as \cite{philbin2007object,mikulik2010learning,rootsift,vlad,asmk}, and usually leads to embeddings that generalize well over various domains. With the rapid advancement of deep learning, such handcrafted embeddings have been replaced with the learning-based ones in the community.
The learning of image embeddings is commonly conducted in a supervised manner, on crowd-labeled datasets \cite{he2016deep,huang2017densely,hu2018squeeze,dosovitskiy2020image}. However, supervised learning is not scalable since manual annotation of large-scale training data is time-consuming and costly. As a result, the training data usually contains limited pre-defined object classes (\eg, ImageNet \cite{deng2009imagenet} and Open Images \cite{OpenImages}), and embeddings learned from these data are not universal enough to generalize to various open-world objects \cite{an2023unicom}. In recent years, self- and weakly-supervised learning have gained extensive attention due to their less reliance on labeled data. By designing appropriate pre-text tasks and training strategies (\eg, image-text matching), these learning paradigms can easily leverage a large number of unlabeled or noisy data, producing image embeddings of greater generality than supervised learning \cite{he2020momentum,he2022masked,chen2021exploring,simclr,byol,clip,align}.

\section{The FORB Benchmark}
\label{sec:dataset}

Our FORB benchmark only provides testing query images without training data. It serves as a testbed supplementary to existing benchmarks, with the following goals.

\paragraph{Goals} Our proposed benchmark aims to enrich the object domains considered in image retrieval tasks and  measure the generalization ability of embedding models with respect to out-of-distribution queries.
Besides, we also seek to understand the effects of image features from different levels on the embedding quality, thereby shedding light on future development of embedding models.

\subsection{Data Collection}

There are 8 different types of flat objects involved in our benchmark: (1) \textit{Animated trading card}. We consider one particular type of card, \ie, Pokemon trading card. (2) \textit{Photorealistic trading card}. We consider cards for different sports, such as baseball, basketball, and football. (3) \textit{Book cover}, which comes from books in different languages, such as English and Chinese. (4) \textit{Painting}, which involves various styles, such as impressionism and baroque, etc. (5) \textit{Currency}, which involves banknotes of modern and antique designs from different countries. We consider both the front and back of a banknote. (6) \textit{Logo}. We consider common logos (\eg, Nike) as well as long-tailed logos (\eg, brands of local small businesses). (7) \textit{Packaged goods}. We only consider products for which the corresponding index images are displayed on flat surface. (8) \textit{Movie poster}. We consider posters from different countries, such as America and Japan. In Figure~\ref{fig:eg_images} we show examples for each object. As can be seen, these objects have diverse textures, involving animation and artificial patterns, etc, and thus offer various retrieval challenges. Also, they are common in daily life and retrieving such objects serves as a practical use case in real applications. For example, eBay builds an image retrieval system\footnote{https://pages.ebay.com/scantolist/} for trading cards to facilitate the sales of cards.

To build our benchmark, we collected the query and index images mainly via Google Images. Specifically, before collecting images, for each type of objects we firstly curated a list of object names. Their names can be obtained from dedicated websites, such as TCGplayer\footnote{https://www.tcgplayer.com/} for animated trading card and Wikimedia Commons\footnote{https://commons.wikimedia.org/} for painting. Next, we queried Google Images with each of the names and retrieved the corresponding query and index images. The returned results were typically noisy and we manually filtered out the irrelevant images as well as those that could be copyright protected. In this way, we effectively matched each index image with diverse query images, giving image-level ground truths. Note that our collected query images are in the wild whereas the database images are in canonical pose (see Figure~\ref{fig:eg_images}). Besides Google Images, we also leveraged some other sources to further augment the benchmark, such as Google Lens API, eBay, and Amazon. 

To increase retrieval difficulty and challenge, similar to previous works \cite{revisited_oxford_paris,gldv2} we also introduced distractors to the benchmark. The distractors are images that share similar semantics, contents, or textures with the index images. They can be from the same domains as the index images, or from other domains. Distractors are primarily introduced to increase the retrieval difficulty, as they would bring perplexing features that deceive retrieval algorithms and reduce the accuracy of retrieval results. Ideally, a strong retrieval algorithm should be robust against distractors. In our benchmark, the distractor images were all from the 8 object domains and crawled from different specific websites, such as TCGplayer and Wikimedia Commons. See supplementary material for some examples. The details of our benchmark can be found in Table \ref{tab:data_stat}.

\begin{table}[t!]
  \caption{Overview of the proposed FORB benchmark.}
  \label{tab:data_stat}
  \resizebox{1.0\columnwidth}{!}{
      \centering
      \begin{tabular}{l|rrr|rrr}
        \hline
        Object Type             & \# Query  & \# Index & \# Distractor & \# Easy & \# Medium & \# Hard\\
        \hline
        Animated trading card & 6,025  & 1,392 & 11,137 & 714   & 4,868  & 443  \\
        Photorealistic trading card & 2,187  & 484   & 521    & 67    & 2,039  & 81 \\
        Book cover           & 1,461  & 470   & 10,739 & 66    & 1,277  & 118  \\
        Painting     & 988    & 430   & 615    & 119   & 710    & 159   \\
        Currency             & 758    & 395   & 1,188  & 112   & 576    & 70    \\
        Logo                 & 1170   & 535   & 174    & 24    & 957    & 189   \\
        Packaged goods       & 800    & 476   & 2,382  & 24    & 727    & 49    \\
        Movie poster         & 512    & 403   & 23,094 & 49    & 426    & 37    \\
        \hline
        Total                & 13,901 & 4,585 & 49,850 & 1,175 & 11,580 & 1,146 \\
        \hline
      \end{tabular}
  }
\end{table}

\subsection{Data Annotation and Metadata}

As mentioned above, we provide image-level retrieval ground truths for each query image. To enable a more detailed evaluation on the quality of image embeddings, we also offer annotations on the retrieval difficulties for each query image. Specifically, we break down difficulty into three levels: easy, medium, and hard. The specific difficulty level for a query image is subject to the following factors: (1) occlusion; (2) blur; (3) truncation; (4) color distortion; (5) perspective distortion; (6) texture complexity; (7) area of the object in the query image. For example, if the target object only occupies a small area in the image, we tag ``hard'' for the given query image due to the distraction of background; see Figure~\ref{fig:eg_images}(h)(n). Similarly, if the object does not bear severe perspective distortion or truncation, we tag “easy”; see Figure~\ref{fig:eg_images}(b). In practice, assigning difficulty levels to query images can be a subjective process. To reduce bias and ensure precise difficulty assessment, we involve different annotators in manually labeling the difficulty of each image and then use majority voting to determine the final difficulty level. As shown in Table~\ref{tab:exp_results}, the annotated difficulty levels are quite consistent with retrieval accuracies for all methods, \ie, the accuracies are high on easy queries, whereas they are low on hard queries.

We store the annotations with a newline delimited JSON file, where each line contains the metadata corresponding to a query image. Specifically, each line is comprised of the following information: (1) query image ID; (2) the file name of query image; (3) the source URL of query image; (4) the file names of ground-truth index images; (5) the source URLs of ground-truth index images; (6) difficulty level. Here the source URL corresponds to where we downloaded the image. Apart from this annotation information, we also provide newline delimited JSON files for tracking the set of query and database images, respectively, where each line contains information regarding the image file name and source URL.

We host the metadata files at https://github.com/pxiangwu/FORB/, which is publicly accessible. As for the query and database images, they can be downloaded via the provided source image URLs. Alternatively, these images are also accessible from a Google drive, where we snapshot all the images from source URLs. Both the metadata and image files are licensed under CC BY-NC-SA. 

\subsection{Metrics}

Our FORB benchmark uses the commonly adopted mAP metric, as well as a new one that takes into account the matching score margin.

\paragraph{mAP} The mean Average Precision metric considers both the true positives and false positives in the ranked retrieval results. The metric is defined as follows:
\begin{equation}
    \textrm{mAP}@k = \frac{1}{Q}\sum_{q=1}^{Q}\textrm{AP}@k(q),~~\textrm{AP}@k(q) = \frac{1}{\min(m_q, k)}\sum_{k=1}^{\min(n_q, k)}\textrm{P}_q(k)\textrm{rel}_q(k),
\end{equation}
where $Q$ is the total number of query images; $m_q$ is the number of ground-truth index images matched with query image $q$; $n_q$ is the number of predictions made by the retrieval method; $\textrm{P}_q(k)$ is the precision at rank $k$ for query image $q$; and $\textrm{rel}_q(k)$ is a relevance indicator function which equals 1 if the result at rank $k$ is relevant and equals to 0 otherwise. Note that for some query images (\eg, OOD images) they do not have associated index images to retrieve, and mAP does not penalize the method even if it retrieves some results for the query images.

\paragraph{$t$-mAP} To take into account OOD queries and false positive results, we introduce thresholded mAP, \ie, $t$-mAP. This metric measures the matching score margin with the aid of OOD queries, and is computed as below:
\begin{equation}
\label{eq:2}
    t\textrm{-mAP} = \frac{1}{\tau(1)}\int_{0}^{\tau(1)}\textrm{mAP}(t)dt,
\end{equation}
where $\tau(x)$ is the threshold that leads to a false positive rate of $1 - x$ on OOD queries after thresholding the retrieved candidates with respect to their matching scores; formally, $\tau(x) = \min \{\Tilde{x} \mid \textrm{FPR}(\Tilde{x}) = 1 - x \}$, where $\textrm{FPR}(\Tilde{x})$ is the false positive rate at threshold $\Tilde{x}$. 
$\textrm{mAP}(t)$ is the mAP computed after the retrieval results are suppressed at threshold $t$.
Note that $\textrm{mAP}(t)$ tends to decrease with increasing threshold $t$. 
However, for an ideal universal image embedding, it is expected to still have a high mAP even at threshold $\tau(1)$, due to its strong discriminability between true positives and false positives.

In practice, to numerically compute Equation~(\ref{eq:2}), we uniformly sample 11 thresholds and average $\textrm{mAP}(t)$ over them:
\begin{equation}
    t\textrm{-mAP} = \frac{1}{11}\sum_{t\in\{0, \tau(0.1), \dots, \tau(1.0)\}}\textrm{mAP}(t).
\end{equation}
As can be seen, $t\textrm{-mAP}$ takes value from $[0, 1]$, with higher value indicating better performance.

\section{Experiments}
\label{sec:exp}

In this section, we evaluate several representative image retrieval methods on our FORB benchmark. Based on the evaluation results, we also provide a detailed analysis on the behavior of different image embeddings and their intriguing properties. 

\subsection{Baseline Methods}

We consider 10 existing image retrieval methods as baselines and investigate their image embedding qualities. According to how the embedding is built, these methods can be categorized into 3 groups.

\paragraph{Bottom-up} This strategy builds a global image embedding based on local image features. The related methods include: (1) BoW \cite{rootsift}. This method extracts RootSIFT \cite{rootsift} local features from the given image, which are then quantized using a codebook and finally assembled into a sparse feature vector, \ie, image embedding. Since BoW only relies on handcrafted low-level image features, the produced embedding tends to have better generalization ability than learning-based ones that fit to certain domains. (2) FIRe \cite{fire}, which extracts mid-level image features and then aggregates them in a manner similar to BoW. However, different from BoW, FIRe is deep learning-based and the feature extraction needs to be learned with certain training data, \eg,  SfM-120k \cite{sfm-120}.

\paragraph{Top-down} Contrary to the bottom-up approach, this strategy learns to extract local image features through image-level supervision on global image embeddings. The local features typically correspond to the convolutional feature maps and are used for feature matching or reranking. In contrast, the global image embeddings are used in the first stage of a retrieval system to efficiently select the most similar images. In our experiment, we consider one representative approach, DELG \cite{delg}, which jointly extracts deep local features and global image embeddings.

\paragraph{Top-only} This strategy performs image retrieval with learned global image embeddings directly, without the need of extracting and using local image features. The global image embeddings are typically produced from a deep model that is trained on a large dataset, in a supervised or self- / weakly-supervised manner. In the experiment, we consider the following state-of-the-art methods: (1) CLIP \cite{clip}; (2) SLIP \cite{slip}; (3) BLIP \cite{blip}; (4) BLIP2 \cite{blip2}; (5) DINO \cite{dino}; (6) DINOv2 \cite{dinov2}; (7) DiHT \cite{diht}. Note that apart from the design differences, another major distinction among these methods lies in their training data; see Table~\ref{tab:train_data} for more details. In Table~\ref{tab:speed} we also show the specific neural network model used in each method. 

It is worth mentioning that for some top-only methods, their training data may overlap with our FORB benchmark. In particular, we find a few images from FORB are also included in LAION-5B \cite{laion}, and therefore training data based on the subset of LAION-5B (\eg, LAION-438M \cite{laion} and 129M \cite{blip}) may also share duplicate images with FORB. In addition, since the training set of CLIP are collected from web, it may overlap with FORB as well. This test set overlap issue has been discussed in previous works \cite{clip, laion} and is considered to have little impact on the validity of performance evaluations. In the supplementary material we perform extra experiments on a deduplicated version of FORB and observe the evaluation results closely resemble those from the original FORB (see Section A.4)

\begin{table}
  \caption{The training data used by different image retrieval methods. ``Web images'' means the training data are sourced from the Internet and typically comprise various 3D objects along with some flat objects. We use the generic term ``3D objects'' to indicate the training data involve diverse 3D objects, such as 3D landmarks, plants, and animals, etc.}
  \label{tab:train_data}
  \resizebox{1.0\columnwidth}{!}{
      \centering
      \begin{tabular}{l|ccr||l|ccr}
        \hline
        Method & Training data & Domain & \# images & Method & Training data & Domain & \#  images \\
        \hline
        BoW \cite{bow} & - & - & - & BLIP \cite{blip} & 129M \cite{blip} & \parbox[t]{3.5cm}{3D objects + web images} & 129M \\
        FIRe \cite{fire} & SfM-120k \cite{sfm-120} & 3D landmark & 120K & BLIP2 \cite{blip2} & 129M \cite{blip} & \parbox[t]{3.5cm}{3D objects + web images} & 129M \\ 
        DELG \cite{delg} & GLD \cite{gld} & 3D landmark & 960K & DINO \cite{dino} & ImageNet \cite{deng2009imagenet} & 3D objects & 1M\\ 
        CLIP \cite{clip} & Proprietary 400M & Web images & 400M & DINOv2 \cite{dinov2} & LVD-142M \cite{dinov2} & 3D objects & 142M\\
        SLIP \cite{slip} & YFCC15M \cite{yfcc100m} & Web images & 15M & DiHT \cite{diht} & LAION-438M \cite{laion} & Web images &  438M\\
        \hline
      \end{tabular}
    }
\end{table}

\begin{table}[t!]
    \caption{Comparison of different image retrieval methods on our FORB benchmark. \textbf{Bolded} numbers indicate the \textbf{best} results. $\dagger$ means the model training data may overlap with FORB and the retrieval accuracy can be interpreted as an ``upper bound'' performance.}
    \label{tab:exp_results}
    \resizebox{1.0\columnwidth}{!}{
        \centering
        \begin{tabular}{l|c|ccc|c|ccr}
        \hline
        & \multicolumn{4}{c|}{mAP@5 (\%)} & \multicolumn{4}{c}{$t$-mAP@5 (\%)} \\
        \hline
        Method & Overall & Easy & Medium & Hard  & Overall & Easy & Medium & Hard \\
        \hline
        BoW \cite{rootsift}            & 78.44 & 90.38 & 79.78 & 52.65 & 62.49 & 78.29 & 63.61 & 35.00\\
        BoW (+ rerank) \cite{rootsift} & 80.38 & 92.69 & 81.77 & 53.70 & 67.83 & 81.95 & 69.04 & 41.03 \\
        FIRe \cite{fire} & 88.08 & \textbf{98.48} & \textbf{90.14} & 56.58 & \textbf{77.50} & \textbf{90.38} & \textbf{79.41} & \textbf{44.97}\\
        \hline
        DELG \cite{delg} & 48.81 & 79.45 & 48.11 & 24.48 & 34.92 & 65.44 & 33.79 & 15.04\\
        DELG (+ rerank) \cite{delg} & 58.74 & 87.96 & 58.43 & 31.91 & 39.47 & 70.64 & 38.45 & 17.74\\
        \hline
        CLIP$^\dagger$ \cite{clip} & \textbf{89.36} & 98.23 & 90.00 & \textbf{73.84} & 67.23 & 87.10 & 67.48 & 44.27\\
        SLIP \cite{slip} & 39.01 & 64.45 & 38.58 & 17.22 & 24.43 & 50.27 & 23.42 & 8.07\\
        BLIP$^\dagger$ \cite{blip} & 74.11 & 94.67 & 74.65 & 47.53 & 49.98 & 81.31 & 49.58 & 21.89 \\
        BLIP2$^\dagger$ \cite{blip2} & 81.73 & 94.28 & 82.72 & 58.85 & 57.11 & 81.59 & 57.43 & 28.77\\
        DINO \cite{dino} & 55.20 & 85.08 & 55.28 & 23.79 & 42.28 & 74.51 & 41.75 & 14.56\\
        DINOv2 \cite{dinov2} & 68.86 & 92.85 & 69.53 & 37.51 & 48.21 & 72.04 & 48.44 & 21.39\\
        DiHT$^\dagger$ \cite{diht} & 84.77 & 96.56 & 85.47 & 65.55 & 60.54 & 83.79 & 61.06 & 31.43\\
        \hline
        \end{tabular}
    }
\end{table}

\subsection{Implementation}
\label{subsec:impl}

In the experiment, we resize the query and database images to standardize the inputs, ensuring that the longest side is no more than 480 while maintaining the original aspect ratio.
For the baseline methods, we implement BoW in Python according to \cite{rootsift}, while for the others we adapt their open source implementations to image retrieval task. Specifically, for both BoW and FIRe, we build the codebook using 10k images randomly sampled from the database images. For DELG, we follow its default protocols and extract multi-scale local and global features for both query and database images. For all top-only methods, we produce multi-scale feature representations as well. To be specific, we firstly build an image pyramid by resizing the input image and then center cropping.
In our implementation, to strike a balance between accuracy and inference speed, we use 3 scales, $\{\frac{1}{\sqrt{2}}, 1, \sqrt{2}\}$, for query images, and 7 scales \cite{gld} for database images. Next, we compute the global image features at each scale and apply $L_2$ normalization to them. Finally, we aggregate all the features by average-pooling, followed by another $L_2$ normalization step.
Such multi-scale features mitigate the issue of lacking scale invariance for top-only methods. In practice, we observe much improved accuracy of multi-scale features compared to the single-scale ones.

The source code for all the implementations is available at https://github.com/pxiangwu/FORB/, and licensed under the MIT license.

\subsection{Evaluation}
\label{subsec:eval}

In Table~\ref{tab:exp_results} we report image retrieval accuracy for different methods in terms of mAP@5 and $t$-mAP@5 (see supplementary material for more results). It can be observed that:

(1) Image embeddings built from handcrafted low-level features can be more universal than many learning-based global image descriptors. In particular, while BoW was introduced decades ago and manually designed, it still outperforms DELG and many top-only methods on our FORB benchmark, demonstrating its strong generalization ability. Moreover, from $t$-mAP it can be observed that BoW is better at separating true positives from irrelevant candidates, giving a larger matching score margin.

(2) Mid-level image features are more discriminative than low-level descriptors, and their induced global image embeddings exhibit a superior generalization ability over OOD domains. In Table~\ref{tab:exp_results} we investigate one baseline method, \ie, FIRe, which builds embeddings from mid-level features.
It can be observed that FIRe overall achieves the best performance among all baselines, with the highest $t$-mAP while giving an mAP on par with CLIP.
To extract mid-level features, FIRe needs a model training procedure. Surprisingly, although FIRe was trained on 3D landmark images, it can still work well on 2D flat object domains.
This could be because in principle the mid-level features of FIRe are similar to the low-level ones, but they typically cover a larger image region and thus incorporate more semantic information, leading to much improved discriminative ability.

(3) For top-only methods, their retrieval accuracies on OOD domains improve with increasing size of model and training data. For example, since the training data of DINO and SLIP are relatively smaller than others, the generalization ability of their image embeddings is inferior to that of CLIP and DiHT, which employ larger model and training set.

\begin{table}[t!]
    \caption{Retrieval accuracies on diverse objects. We report overall mAP and $t$-mAP.
    \textbf{Bolded} numbers indicate the \textbf{best} results.
    $\dagger$ means the model training data may overlap with FORB and the retrieval accuracy can be interpreted as an ``upper bound'' performance.
    }
    \label{tab:breakdown}
    \resizebox{1.0\columnwidth}{!}{
        \centering
        \begin{tabular}{l|cccccccc}
        \hline
        & \multicolumn{8}{c}{mAP@5 (\%) / $t$-mAP@5 (\%)}  \\
        \hline
        Method & \specialcell{Animated\\Card} & \specialcell{Photorealistic\\Card} & \specialcell{Book\\Cover} & \specialcell{Painting}  & Currency & Logo & \specialcell{Packaged\\Goods} & \specialcell{Movie\\Poster} \\
        \hline
        BoW \cite{rootsift}            & 85.93 / 70.42 & 79.82 / 62.98 & 87.92 / 72.68 & 73.33 / 53.64 & 70.79 / 52.64 & 29.98 / 20.20 & 88.57 / 70.93 & 73.40 / 53.19 \\
        BoW (+ rerank) \cite{rootsift} & 89.68 / 76.65 & 84.58 / 70.55 & 89.57 / 77.25 & 77.94 / 62.31 & 73.46 / 59.81 & 20.06 / 16.26 & 82.10 / 70.68 & 76.90 / 61.31 \\
        FIRe \cite{fire} & \textbf{93.92} / \textbf{83.50} & \textbf{95.69} / \textbf{85.17} & 90.55 / \textbf{80.40} & 88.61 / \textbf{78.24} & 81.57 / 69.72 & 42.50 / 33.32 & 92.69 / \textbf{81.21} & 85.35 / \textbf{71.03}\\
        \hline
        DELG \cite{delg} & 53.86 / 43.42 & 43.78 / 24.95 & 58.83 / 39.66 & 29.75 / 16.08 & 65.64 / 47.77 & 13.45 / ~~7.92 & 69.88 / 46.37 & 42.09 / 25.10\\
        DELG (+ rerank) \cite{delg} & 64.95 / 50.42 & 55.63 / 28.38 & 67.91 / 42.50 & 39.83 / 18.24 & 73.94 / 50.93 & 19.17 / ~~9.91 & 76.23 / 48.32 & 49.80 / 27.01\\
        \hline
        CLIP$^\dagger$ \cite{clip} & 91.93 / 72.91 & 74.26 / 54.00 & \textbf{99.17} / 71.48 & 93.12 / 63.43 & \textbf{87.30} / \textbf{73.95} & \textbf{85.29} / \textbf{53.90} & 98.14 / 79.10 & \textbf{86.99} / 53.92 \\
        SLIP \cite{slip} & 34.15 / 24.84 & 47.51 / 34.30 & 45.50 / 22.34 & 55.93 / 26.71 & 29.74 / 24.20 & 14.92 / ~~3.25 & 64.12 / 29.64 & 38.28 / 19.52 \\
        BLIP$^\dagger$ \cite{blip} & 64.87 / 51.64 & 74.22 / 58.61 & 93.43 / 55.50 & 79.60 / 45.10 & 68.93 / 50.17 & 82.86 / 29.25 & 96.37 / 51.04 & 69.51 / 32.64\\
        BLIP2$^\dagger$ \cite{blip2} & 78.21 / 64.47 & 78.23 / 57.77 & 96.44 / 61.86 & 84.43 / 42.36 & 78.32 / 55.11 & 80.59 / 31.74 & 97.70 / 63.32 & 73.53 / 33.82\\
        DINO \cite{dino} & 52.75 / 41.27 & 80.16 / 63.26 & 48.78 / 33.20 & 65.90 / 51.90 & 53.49 / 41.27 & ~~6.15 / ~~3.14 & 77.15 / 56.19 & 55.47 / 41.11\\
        DINOv2 \cite{dinov2} & 70.00 / 45.29 & 87.76 / 76.93 & 68.56 / 37.68 & 79.92 / 57.40 & 65.30 / 55.83 & ~~6.62 / ~~1.44 & 92.26 / 68.55 & 65.04 / 35.88 \\
        DiHT$^\dagger$ \cite{diht} & 83.02 / 67.74 & 78.22 / 62.49 & 95.66 / 65.24 & \textbf{93.25} / 47.97 & 84.41 / 59.14 & 78.09 / 26.77 & \textbf{98.38} / 73.18 & 80.33 / 37.85\\
        \hline
        \end{tabular}
    }
\end{table}

\begin{table}[t!]
    \caption{Architectures and inference speeds (seconds / query) of different methods.}
    \label{tab:speed}
    \resizebox{1.0\columnwidth}{!}{
        \centering
        \begin{tabular}{l|lc||l|lc}
        \hline
        Method & Architecture & Speed & Method & Architecture & Speed\\
        \hline
        BoW \cite{rootsift}            & -               & 0.410 & SLIP \cite{slip}     &  ViT-L/16 & 0.209 \\
        BoW (+ rerank) \cite{rootsift} & -               & 0.418 & BLIP \cite{blip}     &  ViT-L/16 & 0.211 \\
        FIRe \cite{fire}               & ResNet-50       & \textbf{0.124} & BLIP2 \cite{blip2}   &  ViT-g/14 + QFormer & 0.341 \\
        DELG \cite{delg}               & ResNet-50       & 0.376 & DINO \cite{dino}     &  ViT-S/8  & 0.177\\
        DELG (+ rerank) \cite{delg}    & ResNet-50       & 6.015 & DINOv2 \cite{dinov2} &  ViT-L/14 & 0.222\\
        CLIP \cite{clip}               & ViT-L/14@336px  & 0.513 & DiHT \cite{diht}     &  ViT-L/14@336px & 0.357\\
        \hline
        \end{tabular}
    }
\end{table}

(4) Image embeddings based on low- and mid-level features cannot adequately distinguish feature-scarce images. As shown in Table~\ref{tab:breakdown}, both BoW and FIRe fail to accurately recognize logos, which typically consist of simple patterns and contain sparse features. In contrast, the top-only methods are better at handling logos, probably because they describe images based on their high-level semantics and thus suffer less from the lack of lower level features.

In addition to the retrieval accuracies, we also show the inference speeds of different methods in Table~\ref{tab:speed}. We measure the speed on a machine with 120 GB RAM, an NVIDIA T4 GPU and 32 Intel Xeon CPUs (@2.30GHz). Notably, although CLIP achieves the highest mAP among all the methods, it is not efficient since the feature extraction is computationally expensive. In contrast, FIRe runs at a much faster speed, with a similar mAP and even better $t$-mAP. This further demonstrates the advantages of bottom-up strategy and mid-level features.

\subsection{Discussion}

As shown in Table~\ref{tab:exp_results}, while being effective on certain object domains, the embeddings from most of the baseline methods are not universal enough to generalize to diverse open-world objects. This affirms the need for the proposed FORB benchmark to further strengthen the research in the generalization ability of image embeddings. In addition, our benchmark results show that even trained with 3D landmark images, embeddings produced by FIRe can still well distinguish images from OOD domains, indicating the great potential of mid-level features in retrieval tasks. In particular, given the advantages and weaknesses of mid-level features, one future direction would be to develop a retrieval method that jointly leverages the mid- and high-level image features, giving image embeddings that share the benefits of both sides.

\section{Conclusion}
\label{sec:conc}

We present FORB, a benchmark for flat object retrieval and matching. Essentially FORB supplements existing image retrieval benchmarks, and more importantly, it serves as a test bed for evaluating the generalization abilities of image embeddings on OOD domains. Our experiments on FORB shows that embeddings based on low- and mid-level image features overall are more universal than those constructed from high-level semantics. Notably, we observe that the mid-level features introduced by FIRe are surprisingly general and give the best overall retrieval performance, even if the model is trained on 3D landmarks. 
However, despite the overall inferiority, embeddings of high-level semantics are usually more effective for images that contain sparse features. These findings suggest that one potential future direction would be to develop methods that jointly leverage the mid- and high-level image features and combine the strengths of both.

\textbf{Limitations and future work.} In our experiment, we compare baselines which have different model sizes and are trained on various datasets. As a result, the comparisons among these methods and their corresponding embeddings could be unfair to some extent. In addition, our FORB benchmark currently only considers distractors from the same domain as the index images. To improve the diversity and challenges of our benchmark, in the future we plan to collect more distractors from other domains. In addition, to further enrich the OOD queries, we also plan to curate queries beyond the domains of index images, a practice which is similar to GLDv2 \cite{gldv2}. In this way we can better measure the matching score margins of different methods with $t$-mAP. Despite these limitations, our benchmark still serves as a supportive dataset for further research in the task of image retrieval. We hope our work would facilitate the understanding of different image embeddings and promote the design of new methods.

\bibliographystyle{plain}
\bibliography{neurips_data_2023}

\begin{thebibliography}{10}

\bibitem{an2023unicom}
Xiang An, Jiankang Deng, Kaicheng Yang, Jaiwei Li, Ziyong Feng, Jia Guo, Jing Yang, and Tongliang Liu.
\newblock Unicom: Universal and compact representation learning for image retrieval.
\newblock 2023.

\bibitem{rootsift}
Relja Arandjelovi{\'c} and Andrew Zisserman.
\newblock Three things everyone should know to improve object retrieval.
\newblock In {\em Proceedings of the IEEE conference on computer vision and pattern recognition}, pages 2911--2918. IEEE, 2012.

\bibitem{babenko2015aggregating}
Artem Babenko and Victor Lempitsky.
\newblock Aggregating local deep features for image retrieval.
\newblock In {\em Proceedings of the IEEE international conference on computer vision}, pages 1269--1277, 2015.

\bibitem{delg}
Bingyi Cao, Andre Araujo, and Jack Sim.
\newblock Unifying deep local and global features for image search.
\newblock In {\em European conference on computer vision}, pages 726--743. Springer, 2020.

\bibitem{dino}
Mathilde Caron, Hugo Touvron, Ishan Misra, Herv{\'e} J{\'e}gou, Julien Mairal, Piotr Bojanowski, and Armand Joulin.
\newblock Emerging properties in self-supervised vision transformers.
\newblock In {\em Proceedings of the IEEE/CVF international conference on computer vision}, pages 9650--9660, 2021.

\bibitem{simclr}
Ting Chen, Simon Kornblith, Mohammad Norouzi, and Geoffrey Hinton.
\newblock A simple framework for contrastive learning of visual representations.
\newblock In {\em International conference on machine learning}, pages 1597--1607. PMLR, 2020.

\bibitem{chen2021exploring}
Xinlei Chen and Kaiming He.
\newblock Exploring simple siamese representation learning.
\newblock In {\em Proceedings of the IEEE/CVF conference on computer vision and pattern recognition}, pages 15750--15758, 2021.

\bibitem{bow}
Gabriella Csurka, Christopher Dance, Lixin Fan, Jutta Willamowski, and C{\'e}dric Bray.
\newblock Visual categorization with bags of keypoints.
\newblock In {\em Workshop on statistical learning in computer vision, ECCV}, volume~1, pages 1--2. Prague, 2004.

\bibitem{deng2009imagenet}
Jia Deng, Wei Dong, Richard Socher, Li-Jia Li, Kai Li, and Li~Fei-Fei.
\newblock Imagenet: A large-scale hierarchical image database.
\newblock In {\em Proceedings of the IEEE conference on computer vision and pattern recognition}, pages 248--255. Ieee, 2009.

\bibitem{dosovitskiy2020image}
Alexey Dosovitskiy, Lucas Beyer, Alexander Kolesnikov, Dirk Weissenborn, Xiaohua Zhai, Thomas Unterthiner, Mostafa Dehghani, Matthias Minderer, Georg Heigold, Sylvain Gelly, et~al.
\newblock An image is worth 16x16 words: Transformers for image recognition at scale.
\newblock 2020.

\bibitem{gordo2016deep}
Albert Gordo, Jon Almaz{\'a}n, Jerome Revaud, and Diane Larlus.
\newblock Deep image retrieval: Learning global representations for image search.
\newblock In {\em European conference on computer vision}, pages 241--257. Springer, 2016.

\bibitem{byol}
Jean-Bastien Grill, Florian Strub, Florent Altch{\'e}, Corentin Tallec, Pierre Richemond, Elena Buchatskaya, Carl Doersch, Bernardo Avila~Pires, Zhaohan Guo, Mohammad Gheshlaghi~Azar, et~al.
\newblock Bootstrap your own latent-a new approach to self-supervised learning.
\newblock {\em Advances in neural information processing systems}, 33:21271--21284, 2020.

\bibitem{he2022masked}
Kaiming He, Xinlei Chen, Saining Xie, Yanghao Li, Piotr Doll{\'a}r, and Ross Girshick.
\newblock Masked autoencoders are scalable vision learners.
\newblock In {\em Proceedings of the IEEE/CVF conference on computer vision and pattern recognition}, pages 16000--16009, 2022.

\bibitem{he2020momentum}
Kaiming He, Haoqi Fan, Yuxin Wu, Saining Xie, and Ross Girshick.
\newblock Momentum contrast for unsupervised visual representation learning.
\newblock In {\em Proceedings of the IEEE/CVF conference on computer vision and pattern recognition}, pages 9729--9738, 2020.

\bibitem{he2016deep}
Kaiming He, Xiangyu Zhang, Shaoqing Ren, and Jian Sun.
\newblock Deep residual learning for image recognition.
\newblock In {\em Proceedings of the IEEE conference on computer vision and pattern recognition}, pages 770--778, 2016.

\bibitem{hu2018squeeze}
Jie Hu, Li~Shen, and Gang Sun.
\newblock Squeeze-and-excitation networks.
\newblock In {\em Proceedings of the IEEE conference on computer vision and pattern recognition}, pages 7132--7141, 2018.

\bibitem{huang2017densely}
Gao Huang, Zhuang Liu, Laurens Van Der~Maaten, and Kilian~Q Weinberger.
\newblock Densely connected convolutional networks.
\newblock In {\em Proceedings of the IEEE conference on computer vision and pattern recognition}, pages 4700--4708, 2017.

\bibitem{vlad}
Herv{\'e} J{\'e}gou, Matthijs Douze, Cordelia Schmid, and Patrick P{\'e}rez.
\newblock Aggregating local descriptors into a compact image representation.
\newblock In {\em Proceedings of the IEEE conference on computer vision and pattern recognition}, pages 3304--3311. IEEE, 2010.

\bibitem{align}
Chao Jia, Yinfei Yang, Ye~Xia, Yi-Ting Chen, Zarana Parekh, Hieu Pham, Quoc Le, Yun-Hsuan Sung, Zhen Li, and Tom Duerig.
\newblock Scaling up visual and vision-language representation learning with noisy text supervision.
\newblock In {\em International conference on machine learning}, pages 4904--4916. PMLR, 2021.

\bibitem{krause20133d}
Jonathan Krause, Michael Stark, Jia Deng, and Li~Fei-Fei.
\newblock 3d object representations for fine-grained categorization.
\newblock In {\em Proceedings of the IEEE international conference on computer vision workshops}, pages 554--561, 2013.

\bibitem{OpenImages}
Alina Kuznetsova, Hassan Rom, Neil Alldrin, Jasper Uijlings, Ivan Krasin, Jordi Pont-Tuset, Shahab Kamali, Stefan Popov, Matteo Malloci, Alexander Kolesnikov, Tom Duerig, and Vittorio Ferrari.
\newblock The open images dataset v4: Unified image classification, object detection, and visual relationship detection at scale.
\newblock {\em International journal of computer vision}, 2020.

\bibitem{blip2}
Junnan Li, Dongxu Li, Silvio Savarese, and Steven Hoi.
\newblock Blip-2: Bootstrapping language-image pre-training with frozen image encoders and large language models.
\newblock {\em International conference on machine learning}, 2023.

\bibitem{blip}
Junnan Li, Dongxu Li, Caiming Xiong, and Steven Hoi.
\newblock Blip: Bootstrapping language-image pre-training for unified vision-language understanding and generation.
\newblock In {\em International conference on machine learning}, pages 12888--12900. PMLR, 2022.

\bibitem{liu2016deepvehicle}
Hongye Liu, Yonghong Tian, Yaowei Yang, Lu~Pang, and Tiejun Huang.
\newblock Deep relative distance learning: Tell the difference between similar vehicles.
\newblock In {\em Proceedings of the IEEE conference on computer vision and pattern recognition}, pages 2167--2175, 2016.

\bibitem{liu2016deepfashion}
Ziwei Liu, Ping Luo, Shi Qiu, Xiaogang Wang, and Xiaoou Tang.
\newblock Deepfashion: Powering robust clothes recognition and retrieval with rich annotations.
\newblock In {\em Proceedings of the IEEE conference on computer vision and pattern recognition}, pages 1096--1104, 2016.

\bibitem{lowe2004distinctive}
David~G Lowe.
\newblock Distinctive image features from scale-invariant keypoints.
\newblock {\em International journal of computer vision}, 60:91--110, 2004.

\bibitem{mikulik2010learning}
Andrej Mikul{\'\i}k, Michal Perdoch, Ond{\v{r}}ej Chum, and Ji{\v{r}}{\'\i} Matas.
\newblock Learning a fine vocabulary.
\newblock In {\em European conference on computer vision}, pages 1--14. Springer, 2010.

\bibitem{slip}
Norman Mu, Alexander Kirillov, David Wagner, and Saining Xie.
\newblock Slip: Self-supervision meets language-image pre-training.
\newblock In {\em European conference on computer vision}, pages 529--544. Springer, 2022.

\bibitem{delf}
Hyeonwoo Noh, Andre Araujo, Jack Sim, Tobias Weyand, and Bohyung Han.
\newblock Large-scale image retrieval with attentive deep local features.
\newblock In {\em Proceedings of the IEEE international conference on computer vision}, pages 3456--3465, 2017.

\bibitem{gld}
Hyeonwoo Noh, Andre Araujo, Jack Sim, Tobias Weyand, and Bohyung Han.
\newblock Large-scale image retrieval with attentive deep local features.
\newblock In {\em Proceedings of the IEEE international conference on computer vision}, pages 3456--3465, 2017.

\bibitem{sop}
Hyun Oh~Song, Yu~Xiang, Stefanie Jegelka, and Silvio Savarese.
\newblock Deep metric learning via lifted structured feature embedding.
\newblock In {\em Proceedings of the IEEE conference on computer vision and pattern recognition}, pages 4004--4012, 2016.

\bibitem{dinov2}
Maxime Oquab, Timoth{\'e}e Darcet, Th{\'e}o Moutakanni, Huy Vo, Marc Szafraniec, Vasil Khalidov, Pierre Fernandez, Daniel Haziza, Francisco Massa, Alaaeldin El-Nouby, et~al.
\newblock Dinov2: Learning robust visual features without supervision.
\newblock {\em arXiv preprint arXiv:2304.07193}, 2023.

\bibitem{oxford}
James Philbin, Ondrej Chum, Michael Isard, Josef Sivic, and Andrew Zisserman.
\newblock Object retrieval with large vocabularies and fast spatial matching.
\newblock In {\em Proceedings of the IEEE conference on computer vision and pattern recognition}, pages 1--8. IEEE, 2007.

\bibitem{philbin2007object}
James Philbin, Ondrej Chum, Michael Isard, Josef Sivic, and Andrew Zisserman.
\newblock Object retrieval with large vocabularies and fast spatial matching.
\newblock In {\em Proceedings of the IEEE conference on computer vision and pattern recognition}, pages 1--8. IEEE, 2007.

\bibitem{paris}
James Philbin, Ondrej Chum, Michael Isard, Josef Sivic, and Andrew Zisserman.
\newblock Lost in quantization: Improving particular object retrieval in large scale image databases.
\newblock In {\em Proceedings of the IEEE conference on computer vision and pattern recognition}, pages 1--8. IEEE, 2008.

\bibitem{diht}
Filip Radenovic, Abhimanyu Dubey, Abhishek Kadian, Todor Mihaylov, Simon Vandenhende, Yash Patel, Yi~Wen, Vignesh Ramanathan, and Dhruv Mahajan.
\newblock Filtering, distillation, and hard negatives for vision-language pre-training.
\newblock {\em Proceedings of the IEEE/CVF conference on computer vision and pattern recognition}, 2023.

\bibitem{revisited_oxford_paris}
Filip Radenovi{\'c}, Ahmet Iscen, Giorgos Tolias, Yannis Avrithis, and Ond{\v{r}}ej Chum.
\newblock Revisiting oxford and paris: Large-scale image retrieval benchmarking.
\newblock In {\em Proceedings of the IEEE conference on computer vision and pattern recognition}, pages 5706--5715, 2018.

\bibitem{radenovic2018fine}
Filip Radenovi{\'c}, Giorgos Tolias, and Ond{\v{r}}ej Chum.
\newblock Fine-tuning cnn image retrieval with no human annotation.
\newblock {\em IEEE transactions on pattern analysis and machine intelligence}, 41(7):1655--1668, 2018.

\bibitem{sfm-120}
Filip Radenovi{\'c}, Giorgos Tolias, and Ond{\v{r}}ej Chum.
\newblock Fine-tuning cnn image retrieval with no human annotation.
\newblock {\em IEEE transactions on pattern analysis and machine intelligence}, 41(7):1655--1668, 2018.

\bibitem{clip}
Alec Radford, Jong~Wook Kim, Chris Hallacy, Aditya Ramesh, Gabriel Goh, Sandhini Agarwal, Girish Sastry, Amanda Askell, Pamela Mishkin, Jack Clark, et~al.
\newblock Learning transferable visual models from natural language supervision.
\newblock In {\em International conference on machine learning}, pages 8748--8763. PMLR, 2021.

\bibitem{FlickrLogos}
Stefan Romberg, Lluis~Garcia Pueyo, Rainer Lienhart, and Roelof Van~Zwol.
\newblock Scalable logo recognition in real-world images.
\newblock In {\em Proceedings of the 1st ACM international conference on multimedia retrieval}, pages 1--8, 2011.

\bibitem{laion}
Christoph Schuhmann, Romain Beaumont, Richard Vencu, Cade Gordon, Ross Wightman, Mehdi Cherti, Theo Coombes, Aarush Katta, Clayton Mullis, Mitchell Wortsman, et~al.
\newblock Laion-5b: An open large-scale dataset for training next generation image-text models.
\newblock {\em arXiv preprint arXiv:2210.08402}, 2022.

\bibitem{yfcc100m}
Bart Thomee, David~A Shamma, Gerald Friedland, Benjamin Elizalde, Karl Ni, Douglas Poland, Damian Borth, and Li-Jia Li.
\newblock Yfcc100m: The new data in multimedia research.
\newblock {\em Communications of the ACM}, 59(2):64--73, 2016.

\bibitem{asmk}
Giorgos Tolias, Yannis Avrithis, and Herv{\'e} J{\'e}gou.
\newblock To aggregate or not to aggregate: Selective match kernels for image search.
\newblock In {\em Proceedings of the IEEE international conference on computer vision}, pages 1401--1408, 2013.

\bibitem{opensetlogo}
Andras T{\"u}zk{\"o}, Christian Herrmann, Daniel Manger, and J{\"u}rgen Beyerer.
\newblock Open set logo detection and retrieval.
\newblock {\em arXiv preprint arXiv:1710.10891}, 2017.

\bibitem{van2018inaturalist}
Grant Van~Horn, Oisin Mac~Aodha, Yang Song, Yin Cui, Chen Sun, Alex Shepard, Hartwig Adam, Pietro Perona, and Serge Belongie.
\newblock The inaturalist species classification and detection dataset.
\newblock In {\em Proceedings of the IEEE conference on computer vision and pattern recognition}, pages 8769--8778, 2018.

\bibitem{fire}
Philippe Weinzaepfel, Thomas Lucas, Diane Larlus, and Yannis Kalantidis.
\newblock Learning super-features for image retrieval.
\newblock {\em International conference on learning representations}, 2022.

\bibitem{welinder2010caltech}
Peter Welinder, Steve Branson, Takeshi Mita, Catherine Wah, Florian Schroff, Serge Belongie, and Pietro Perona.
\newblock Caltech-ucsd birds 200.
\newblock 2010.

\bibitem{gldv2}
Tobias Weyand, Andre Araujo, Bingyi Cao, and Jack Sim.
\newblock Google landmarks dataset v2-a large-scale benchmark for instance-level recognition and retrieval.
\newblock In {\em Proceedings of the IEEE/CVF conference on computer vision and pattern recognition}, pages 2575--2584, 2020.

\end{thebibliography}


\begin{thebibliography}{10}

\bibitem{rootsift}
Relja Arandjelovi{\'c} and Andrew Zisserman.
\newblock Three things everyone should know to improve object retrieval.
\newblock In {\em Proceedings of the IEEE conference on computer vision and pattern recognition}, pages 2911--2918. IEEE, 2012.

\bibitem{delg}
Bingyi Cao, Andre Araujo, and Jack Sim.
\newblock Unifying deep local and global features for image search.
\newblock In {\em European conference on computer vision}, pages 726--743. Springer, 2020.

\bibitem{dino}
Mathilde Caron, Hugo Touvron, Ishan Misra, Herv{\'e} J{\'e}gou, Julien Mairal, Piotr Bojanowski, and Armand Joulin.
\newblock Emerging properties in self-supervised vision transformers.
\newblock In {\em Proceedings of the IEEE/CVF international conference on computer vision}, pages 9650--9660, 2021.

\bibitem{bow}
Gabriella Csurka, Christopher Dance, Lixin Fan, Jutta Willamowski, and C{\'e}dric Bray.
\newblock Visual categorization with bags of keypoints.
\newblock In {\em Workshop on statistical learning in computer vision, ECCV}, volume~1, pages 1--2. Prague, 2004.

\bibitem{blip2}
Junnan Li, Dongxu Li, Silvio Savarese, and Steven Hoi.
\newblock Blip-2: Bootstrapping language-image pre-training with frozen image encoders and large language models.
\newblock {\em International conference on machine learning}, 2023.

\bibitem{blip}
Junnan Li, Dongxu Li, Caiming Xiong, and Steven Hoi.
\newblock Blip: Bootstrapping language-image pre-training for unified vision-language understanding and generation.
\newblock In {\em International conference on machine learning}, pages 12888--12900. PMLR, 2022.

\bibitem{slip}
Norman Mu, Alexander Kirillov, David Wagner, and Saining Xie.
\newblock Slip: Self-supervision meets language-image pre-training.
\newblock In {\em European conference on computer vision}, pages 529--544. Springer, 2022.

\bibitem{dinov2}
Maxime Oquab, Timoth{\'e}e Darcet, Th{\'e}o Moutakanni, Huy Vo, Marc Szafraniec, Vasil Khalidov, Pierre Fernandez, Daniel Haziza, Francisco Massa, Alaaeldin El-Nouby, et~al.
\newblock Dinov2: Learning robust visual features without supervision.
\newblock {\em arXiv preprint arXiv:2304.07193}, 2023.

\bibitem{diht}
Filip Radenovic, Abhimanyu Dubey, Abhishek Kadian, Todor Mihaylov, Simon Vandenhende, Yash Patel, Yi~Wen, Vignesh Ramanathan, and Dhruv Mahajan.
\newblock Filtering, distillation, and hard negatives for vision-language pre-training.
\newblock {\em Proceedings of the IEEE/CVF conference on computer vision and pattern recognition}, 2023.

\bibitem{clip}
Alec Radford, Jong~Wook Kim, Chris Hallacy, Aditya Ramesh, Gabriel Goh, Sandhini Agarwal, Girish Sastry, Amanda Askell, Pamela Mishkin, Jack Clark, et~al.
\newblock Learning transferable visual models from natural language supervision.
\newblock In {\em International conference on machine learning}, pages 8748--8763. PMLR, 2021.

\bibitem{laion}
Christoph Schuhmann, Romain Beaumont, Richard Vencu, Cade Gordon, Ross Wightman, Mehdi Cherti, Theo Coombes, Aarush Katta, Clayton Mullis, Mitchell Wortsman, et~al.
\newblock Laion-5b: An open large-scale dataset for training next generation image-text models.
\newblock {\em arXiv preprint arXiv:2210.08402}, 2022.

\bibitem{fire}
Philippe Weinzaepfel, Thomas Lucas, Diane Larlus, and Yannis Kalantidis.
\newblock Learning super-features for image retrieval.
\newblock {\em International conference on learning representations}, 2022.

\end{thebibliography}

\section*{Checklist}


\begin{enumerate}

\item For all authors...
\begin{enumerate}
  \item Do the main claims made in the abstract and introduction accurately reflect the paper's contributions and scope?
    \answerYes{}
  \item Did you describe the limitations of your work?
    \answerYes{See Section~\ref{sec:conc}.}
  \item Did you discuss any potential negative societal impacts of your work?
    \answerNA{}
  \item Have you read the ethics review guidelines and ensured that your paper conforms to them?
    \answerYes{}
\end{enumerate}

\item If you are including theoretical results...
\begin{enumerate}
  \item Did you state the full set of assumptions of all theoretical results?
    \answerNA{}
	\item Did you include complete proofs of all theoretical results?
    \answerNA{}
\end{enumerate}

\item If you ran experiments (e.g. for benchmarks)...
\begin{enumerate}
  \item Did you include the code, data, and instructions needed to reproduce the main experimental results (either in the supplemental material or as a URL)?
    \answerYes{Datasets and supportive code to reproduce the results in this paper are available at https://github.com/pxiangwu/FORB/.}
  \item Did you specify all the training details (e.g., data splits, hyperparameters, how they were chosen)?
    \answerYes{See Subsection~\ref{subsec:impl} and Table~\ref{tab:train_data}.}
	\item Did you report error bars (e.g., with respect to the random seed after running experiments multiple times)?
    \answerNA{For top-down and top-only methods, we use the fixed pretrained models from existing works. For bottom-up methods, they are quite robust to the randomness in the construction of codebook, which is performed via k-means.}
	\item Did you include the total amount of compute and the type of resources used (e.g., type of GPUs, internal cluster, or cloud provider)?
    \answerYes{See Subsection~\ref{subsec:eval}. Specifically, we use a machine with 120 GB RAM, an NVIDIA T4 GPU and 32 Intel Xeon CPUs (@2.30GHz).}
\end{enumerate}

\item If you are using existing assets (e.g., code, data, models) or curating/releasing new assets...
\begin{enumerate}
  \item If your work uses existing assets, did you cite the creators?
    \answerYes{}
  \item Did you mention the license of the assets?
    \answerYes{See supplementary material and our GitHub repository: https://github.com/pxiangwu/FORB/.}
  \item Did you include any new assets either in the supplemental material or as a URL?
    \answerYes{See Abstract and Section\ref{sec:intro}.}
  \item Did you discuss whether and how consent was obtained from people whose data you're using/curating?
    \answerNo{We sourced the benchmark images from Internet.}
  \item Did you discuss whether the data you are using/curating contains personally identifiable information or offensive content?
    \answerNA{}
\end{enumerate}

\item If you used crowdsourcing or conducted research with human subjects...
\begin{enumerate}
  \item Did you include the full text of instructions given to participants and screenshots, if applicable?
    \answerNA{}
  \item Did you describe any potential participant risks, with links to Institutional Review Board (IRB) approvals, if applicable?
    \answerNA{}
  \item Did you include the estimated hourly wage paid to participants and the total amount spent on participant compensation?
    \answerNA{}
\end{enumerate}

\end{enumerate}

\end{document}


\maketitle

\appendix

\section{Appendix}

\subsection{More Query Images with Different Difficulties}

\begin{figure}[h!]
    \begin{center}
        \begin{minipage}[b]{.154\linewidth}
            \centering
            \includegraphics[width=\textwidth]{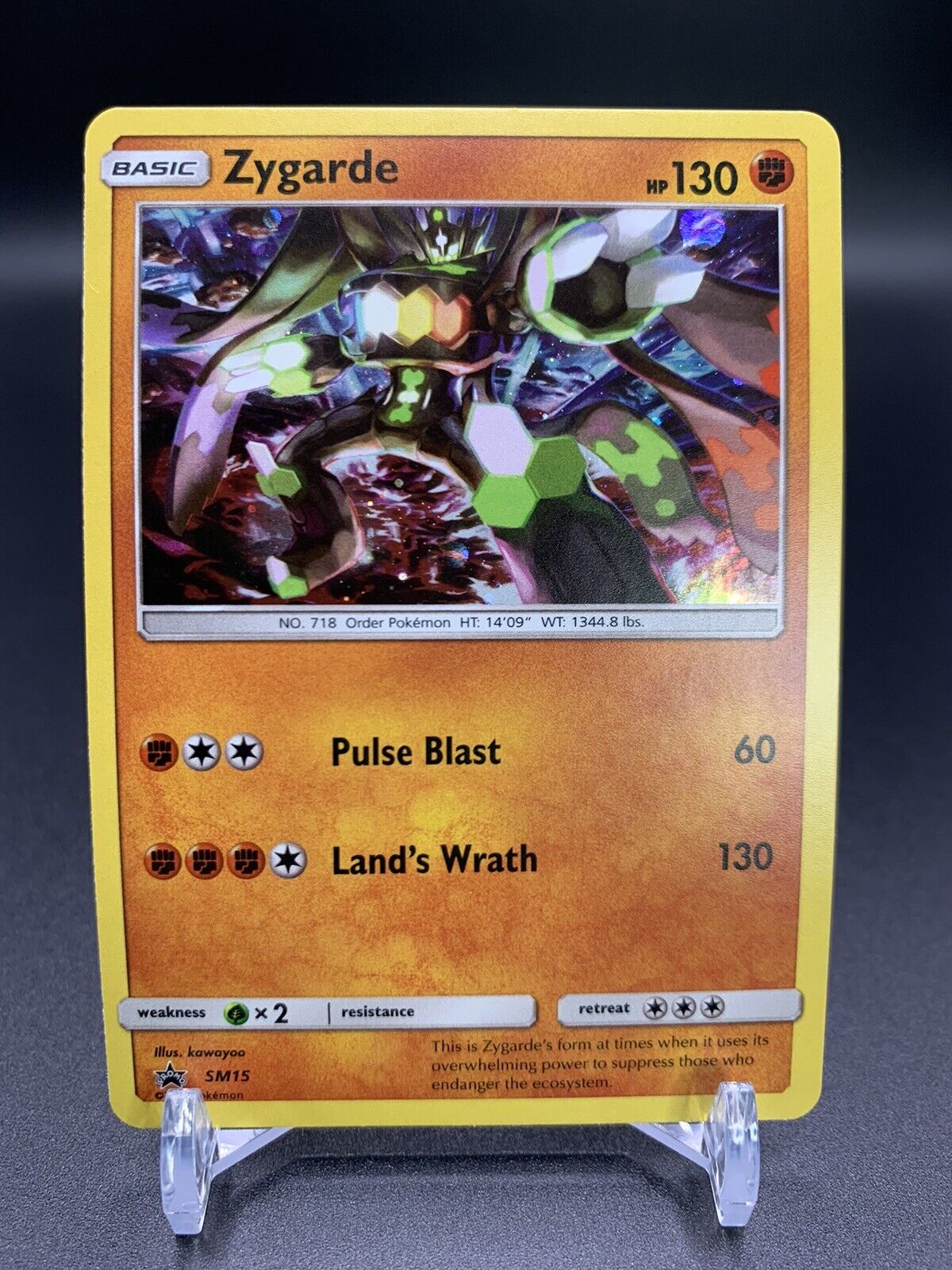}
            \scriptsize
            Query (easy)
        \end{minipage}
        \begin{minipage}[b]{.154\linewidth}
            \centering
            \includegraphics[width=\textwidth]{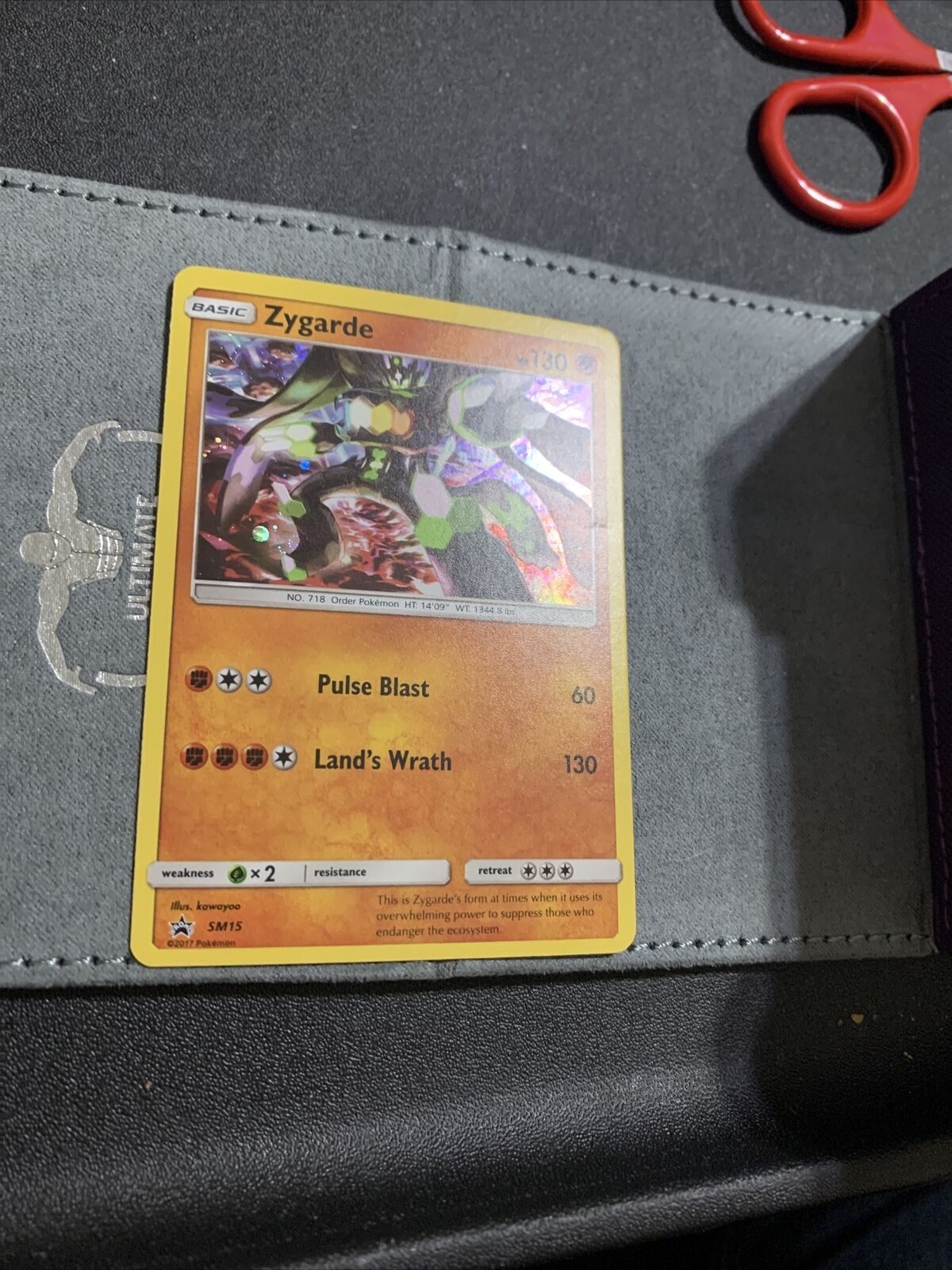}
            \scriptsize
            Query (medium)
        \end{minipage}
        \begin{minipage}[b]{.368\linewidth}
            \centering
            \includegraphics[width=\textwidth]{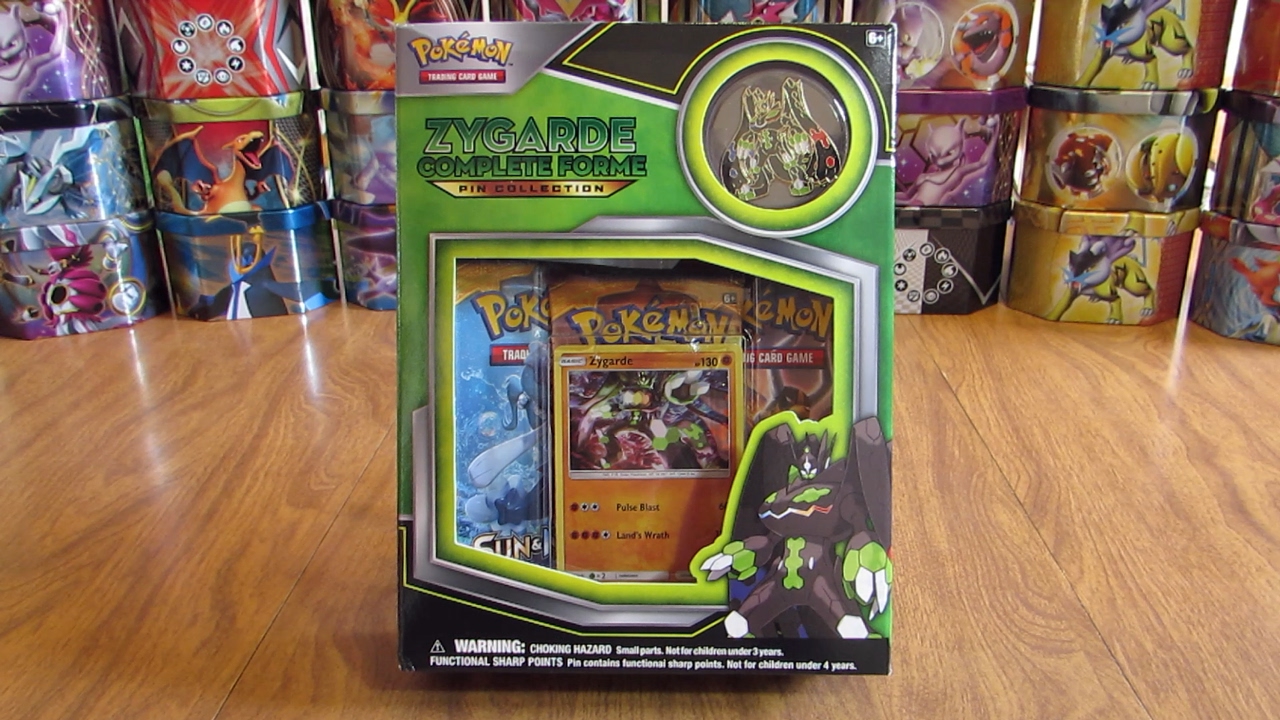}
            \scriptsize
            Query (hard)
        \end{minipage}
        \begin{minipage}[b]{.148\linewidth}
            \centering
            \includegraphics[width=\textwidth]{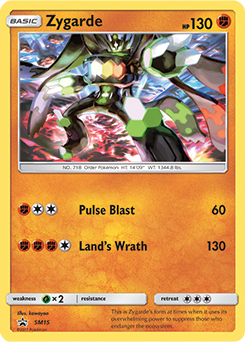}
            \scriptsize
            Index
        \end{minipage}
    \end{center}
    \vspace{-0.22cm}
    \caption{Example images of \textbf{animated trading card}. In the query image with ``hard'' difficulty, the target object only occupies a small area, making it non-trivial to recognize due to the distraction of background. In contrast, in ``easy'' and ``medium'' queries, the target object occupies a larger area.}
    \label{fig:pokemon}
\end{figure}

\begin{figure}[h!]
    \begin{center}
        \begin{minipage}[b]{.193\linewidth}
            \centering
            \includegraphics[width=\textwidth]{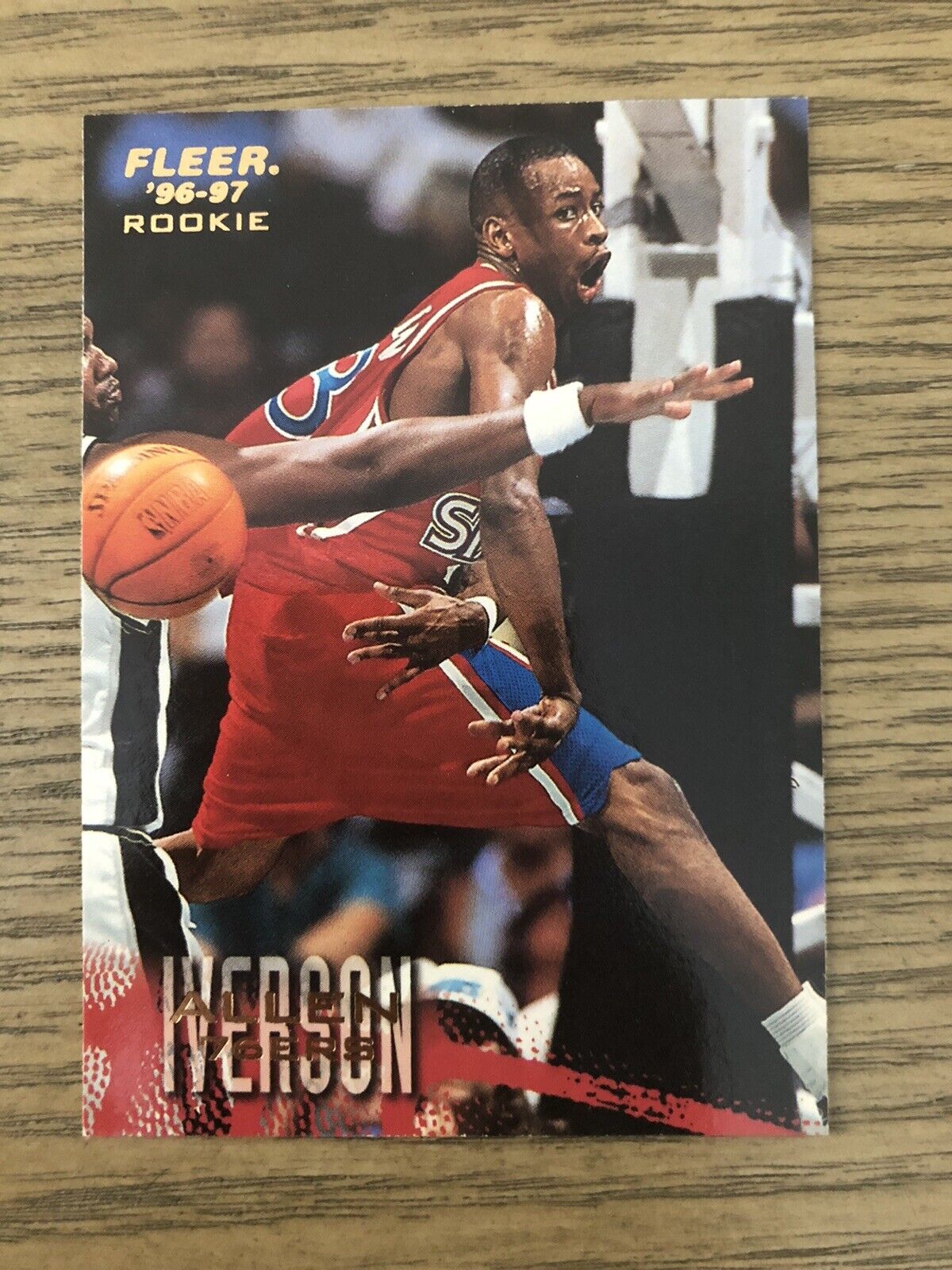}
            \scriptsize
            Query (easy)
        \end{minipage}
        \begin{minipage}[b]{.193\linewidth}
            \centering
            \includegraphics[width=\textwidth]{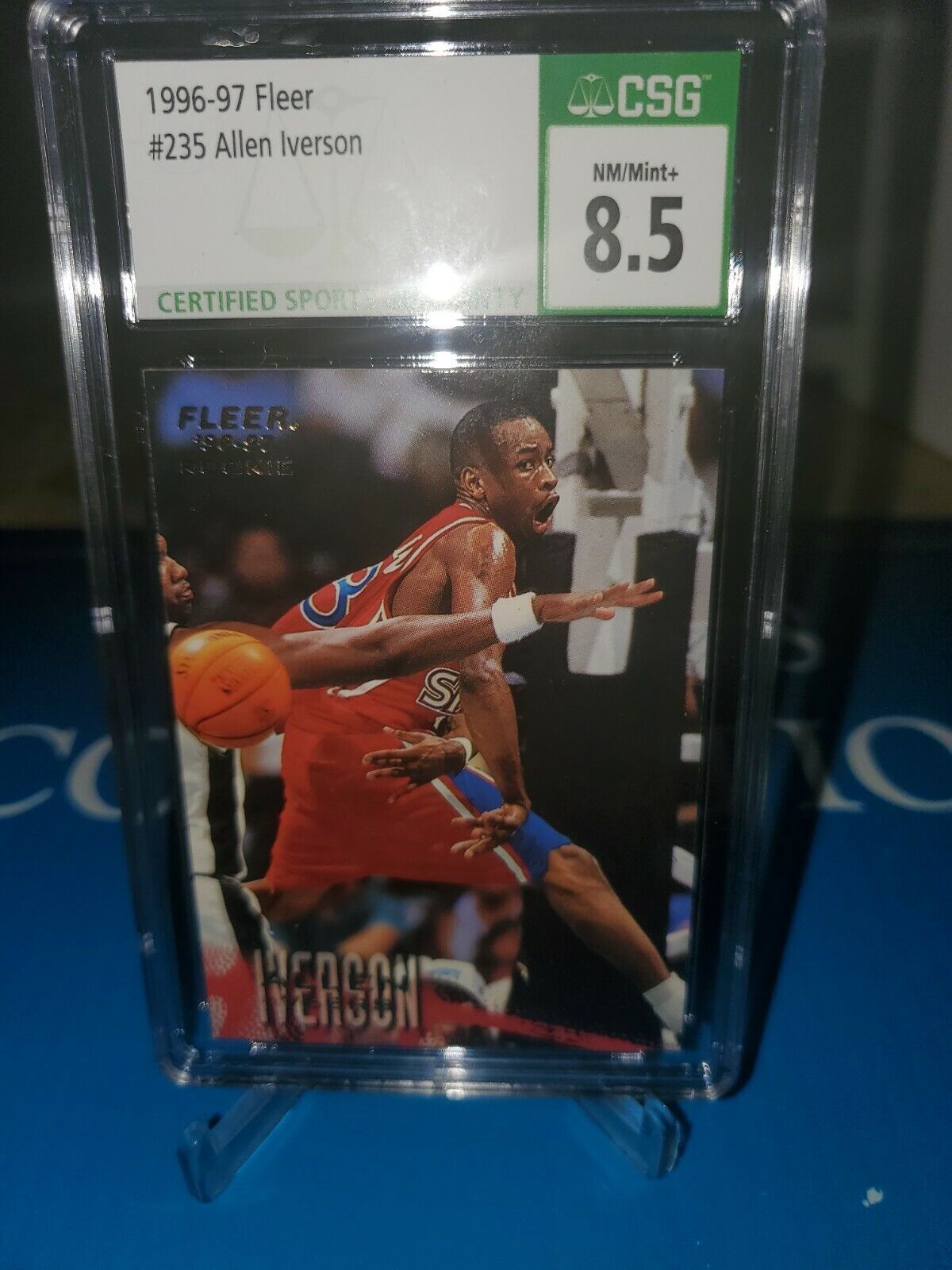}
            \scriptsize
            Query (medium)
        \end{minipage}
        \begin{minipage}[b]{.258\linewidth}
            \centering
            \includegraphics[width=\textwidth]{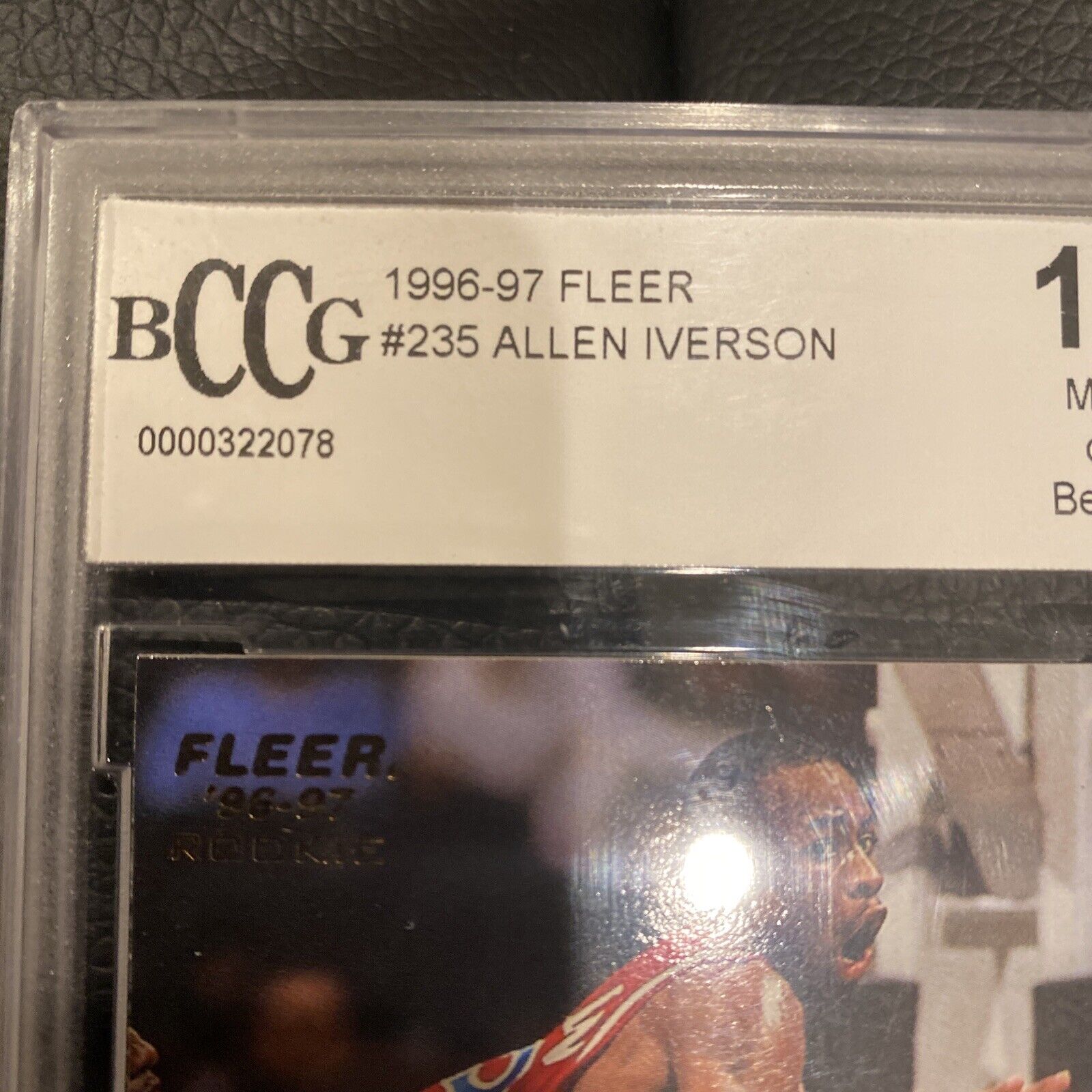}
            \scriptsize
            Query (hard)
        \end{minipage}
        \begin{minipage}[b]{.183\linewidth}
            \centering
            \includegraphics[width=\textwidth]{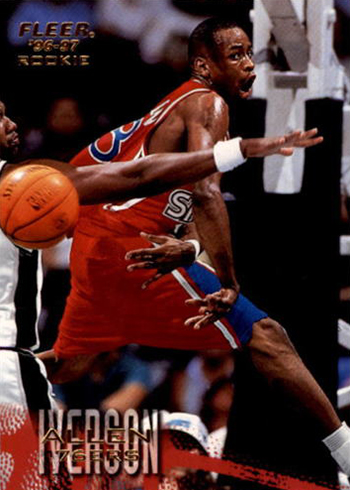}
            \scriptsize
            Index
        \end{minipage}
    \end{center}
    \vspace{-0.22cm}
    \caption{Example images of \textbf{photorealistic trading card}. In the query image with ``hard'' difficulty, the target trading card is severely truncated and thus difficult to recognize.}
    \label{fig:sports}
\end{figure}

\begin{figure}[h!]
    \begin{center}
        \begin{minipage}[b]{.1735\linewidth}
            \centering
            \includegraphics[width=\textwidth]{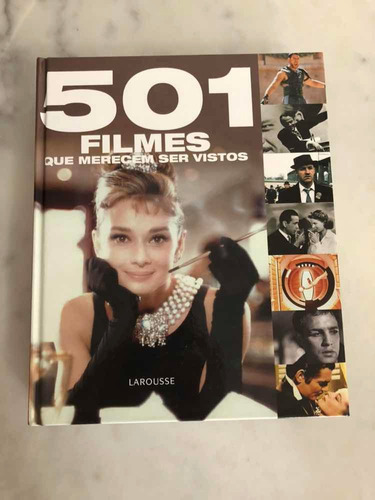}
            \scriptsize
            Query (easy)
        \end{minipage}
        \begin{minipage}[b]{.231\linewidth}
            \centering
            \includegraphics[width=\textwidth]{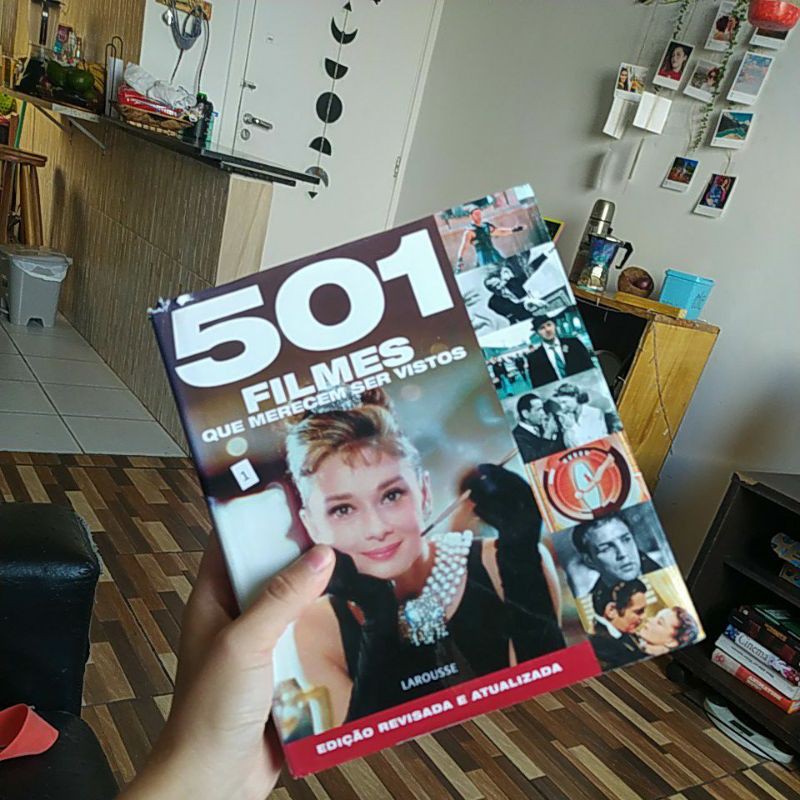}
            \scriptsize
            Query (medium)
        \end{minipage}
        \begin{minipage}[b]{.231\linewidth}
            \centering
            \includegraphics[width=\textwidth]{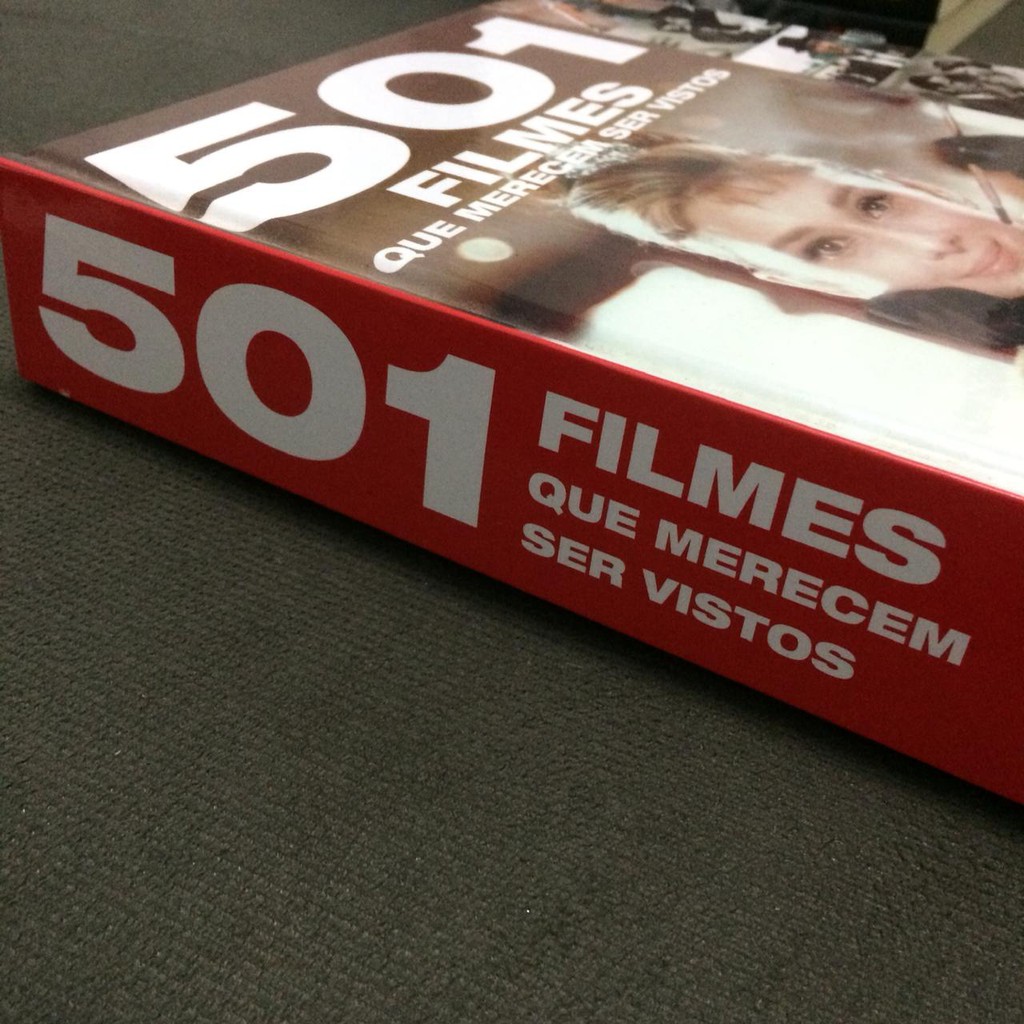}
            \scriptsize
            Query (hard)
        \end{minipage}
        \begin{minipage}[b]{.191\linewidth}
            \centering
            \includegraphics[width=\textwidth]{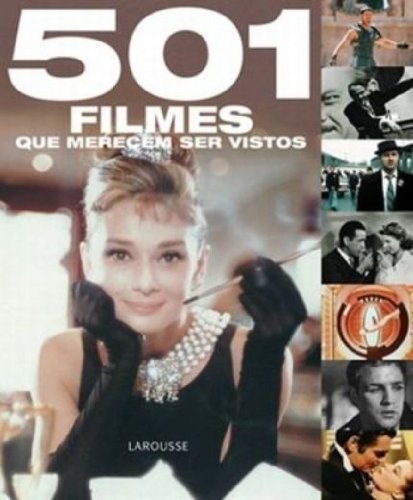}
            \scriptsize
            Index
        \end{minipage}
    \end{center}
    \vspace{-0.22cm}
    \caption{Example images of \textbf{book cover}. In the query image with ``hard'' difficulty, the target book cover is truncated and under a large perspective transformation.}
    \label{fig:bookcover}
\end{figure}

\begin{figure}[h!]
    \begin{center}
        \begin{minipage}[b]{.153\linewidth}
            \centering
            \includegraphics[width=\textwidth]{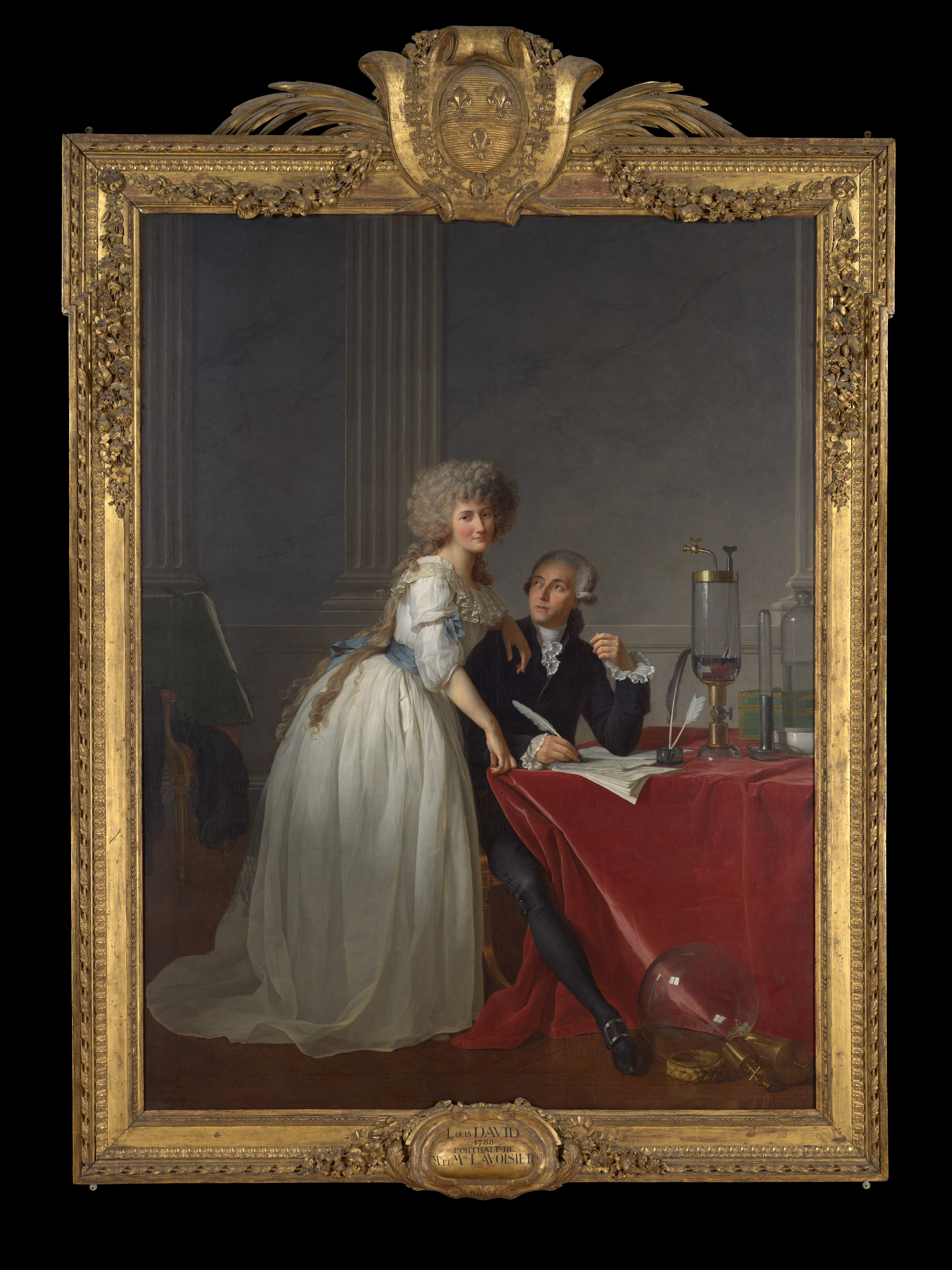}
            \scriptsize
            Query (easy)
        \end{minipage}
        \begin{minipage}[b]{.153\linewidth}
            \centering
            \includegraphics[width=\textwidth]{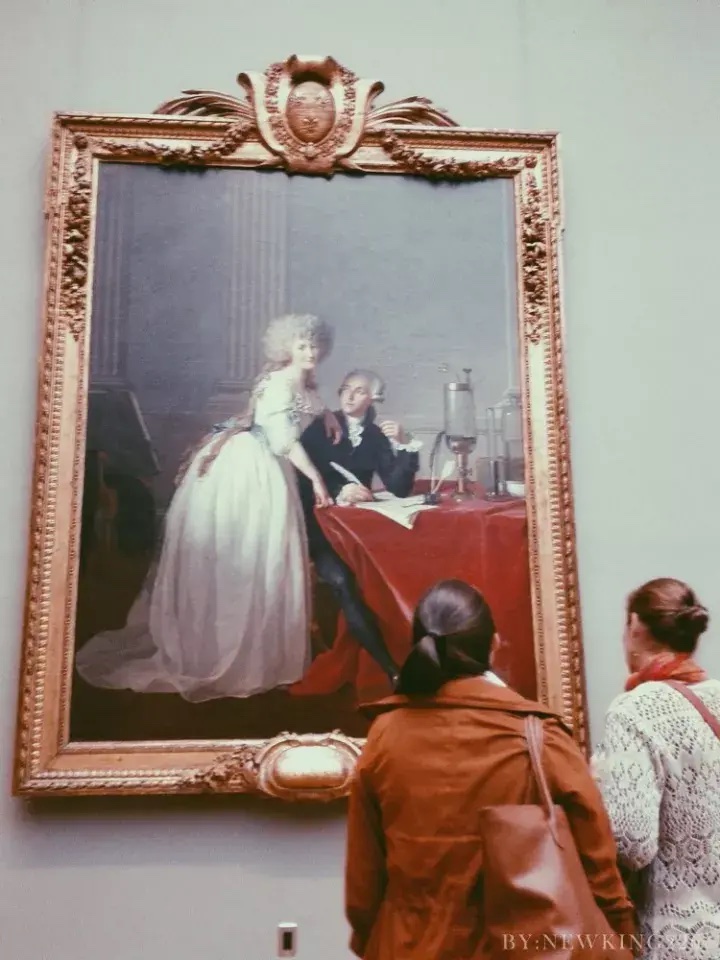}
            \scriptsize
            Query (medium)
        \end{minipage}
        \begin{minipage}[b]{.366\linewidth}
            \centering
            \includegraphics[width=\textwidth]{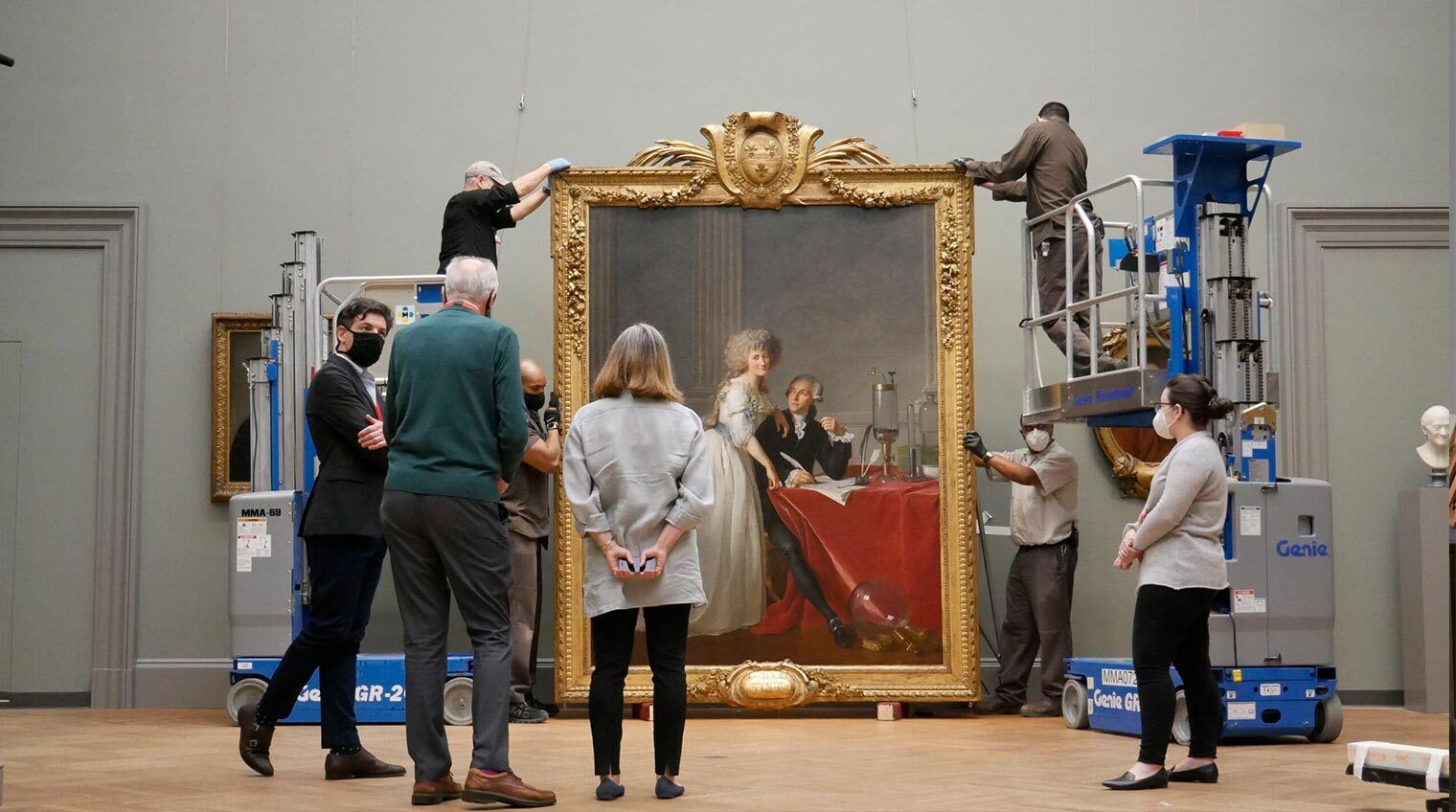}
            \scriptsize
            Query (hard)
        \end{minipage}
        \begin{minipage}[b]{.1535\linewidth}
            \centering
            \includegraphics[width=\textwidth]{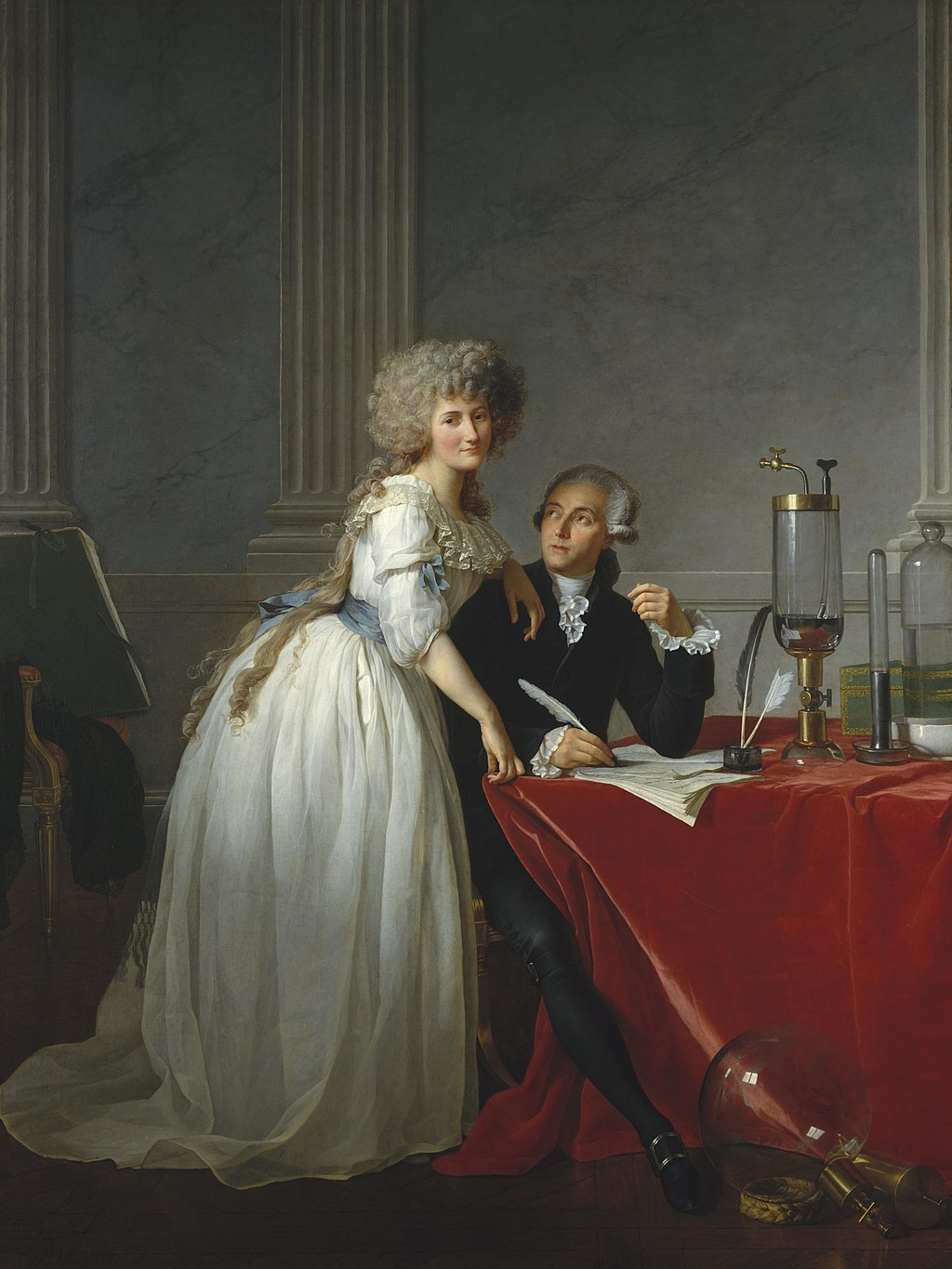}
            \scriptsize
            Index
        \end{minipage}
    \end{center}
    \vspace{-0.22cm}
    \caption{Example images of \textbf{painting}. In the query image with ``hard'' difficulty, the target painting is occluded and occupies a small area.}
    \label{fig:painting}
\end{figure}

\begin{figure}[h!]
    \begin{center}
        \begin{minipage}[b]{.31\linewidth}
            \centering
            \includegraphics[width=\textwidth]{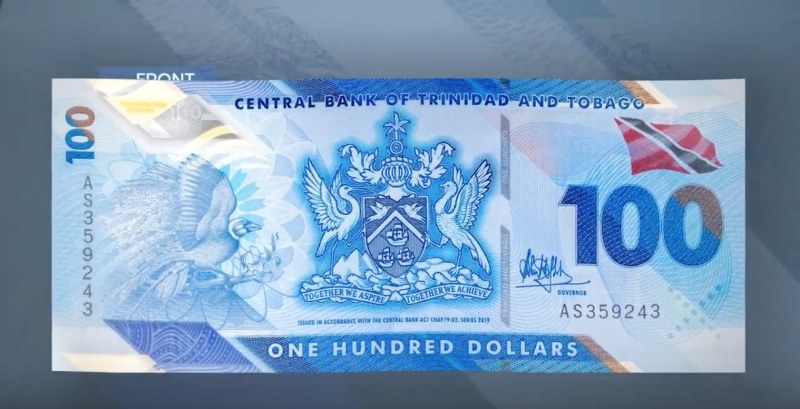}
            \scriptsize
            Query (easy)
        \end{minipage}
            \begin{minipage}[b]{.2105\linewidth}
            \centering
            \includegraphics[width=\textwidth]{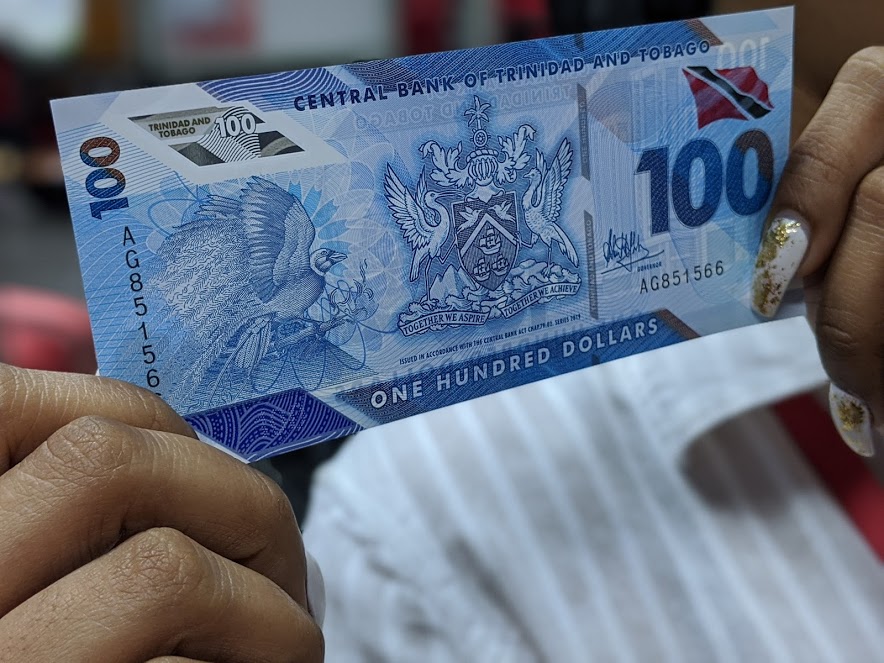}
            \scriptsize
            Query (medium)
        \end{minipage}
            \begin{minipage}[b]{.24\linewidth}
            \centering
            \includegraphics[width=\textwidth]{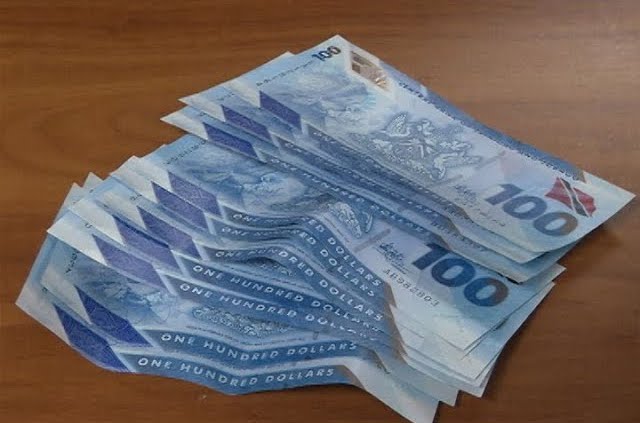}
            \scriptsize
            Query (hard)
        \end{minipage}
            \begin{minipage}[b]{.0678\linewidth}
            \centering
            \includegraphics[width=\textwidth]{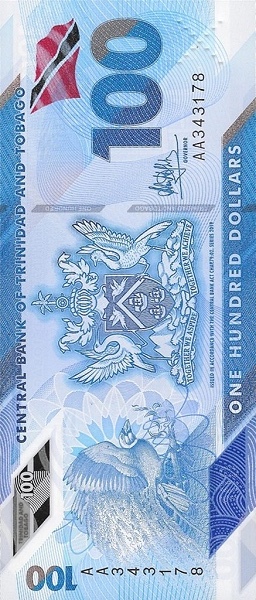}
            \scriptsize
            Index
        \end{minipage}
    \end{center}
    \vspace{-0.22cm}
    \caption{Example images of \textbf{currency}. In the query image with ``hard'' difficulty, the target currency is blurry and only occupies a small area.}
    \label{fig:currency}
\end{figure}

\begin{figure}[h!]
    \begin{center}
        \begin{minipage}[b]{.20\linewidth}
            \centering
            \includegraphics[width=\textwidth]{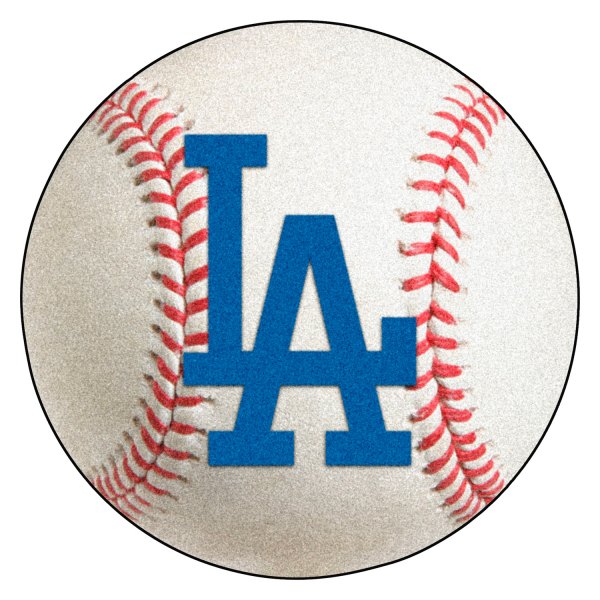}
            \scriptsize
            Query (easy)
        \end{minipage}
        \begin{minipage}[b]{.20\linewidth}
            \centering
            \includegraphics[width=\textwidth]{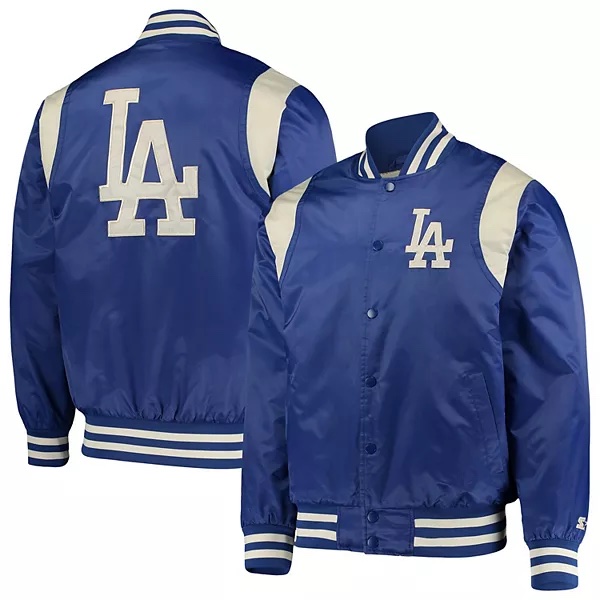}
            \scriptsize
            Query (medium)
        \end{minipage}
        \begin{minipage}[b]{.285\linewidth}
            \centering
            \includegraphics[width=\textwidth]{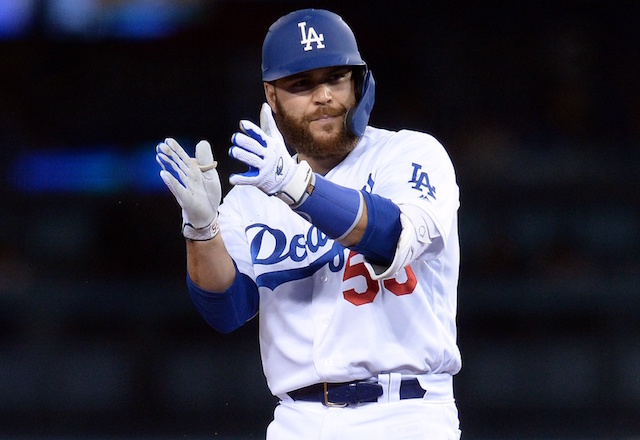}
            \scriptsize
            Query (hard)
        \end{minipage}
        \begin{minipage}[b]{.155\linewidth}
            \centering
            \includegraphics[width=\textwidth]{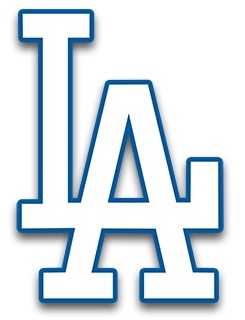}
            \scriptsize
            Index
        \end{minipage}
    \end{center}
    \vspace{-0.22cm}
    \caption{Example images of \textbf{logo}. In the query image with ``hard'' difficulty, the target logo only occupies a small area, with much distraction from the background.}
    \label{fig:logo}
\end{figure}

\begin{figure}[h!]
    \begin{center}
        \begin{minipage}[b]{.217\linewidth}
            \centering
            \includegraphics[width=\textwidth]{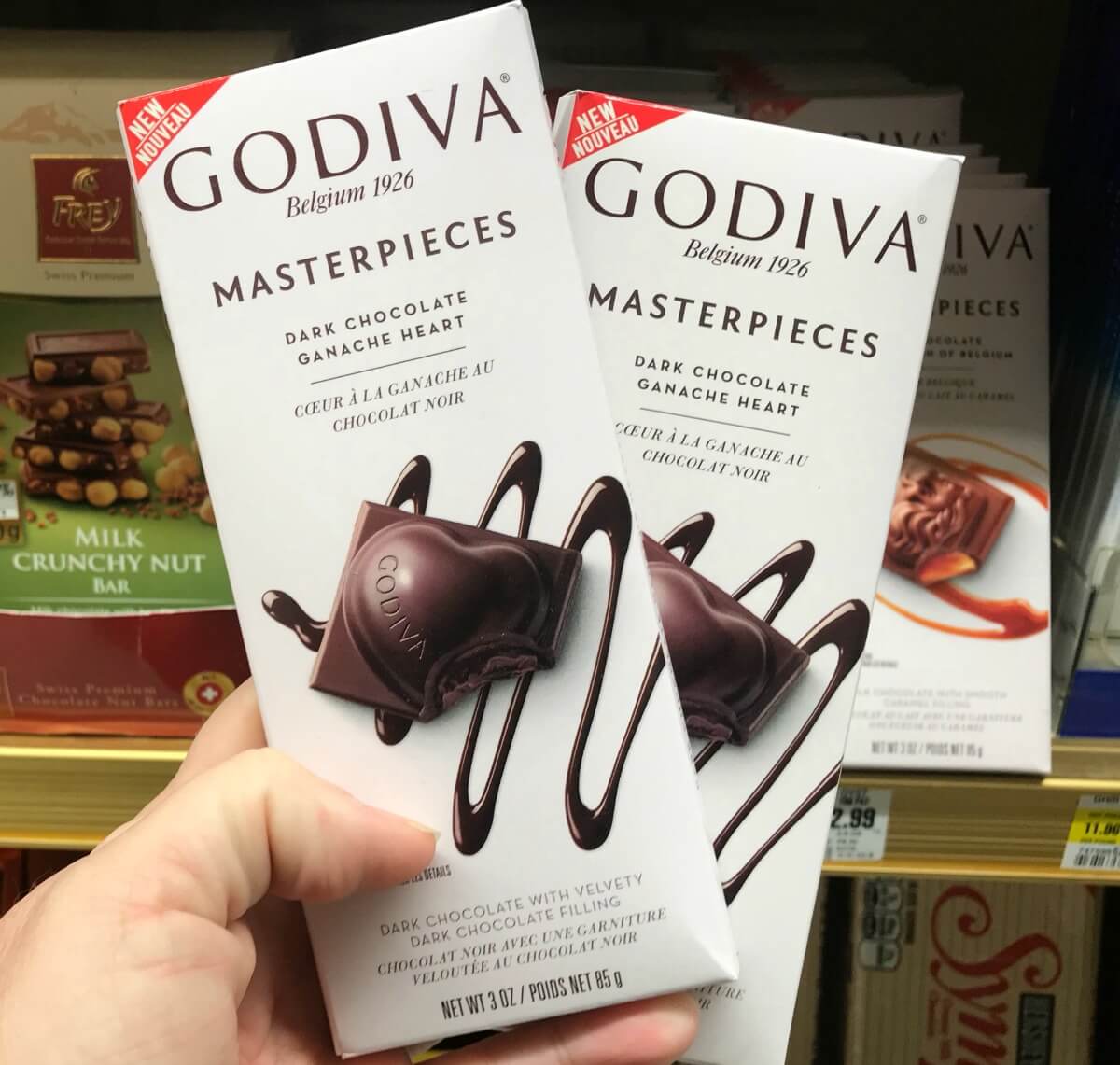}
            \scriptsize
            Query (easy)
        \end{minipage}
        \begin{minipage}[b]{.297\linewidth}
            \centering
            \includegraphics[width=\textwidth]{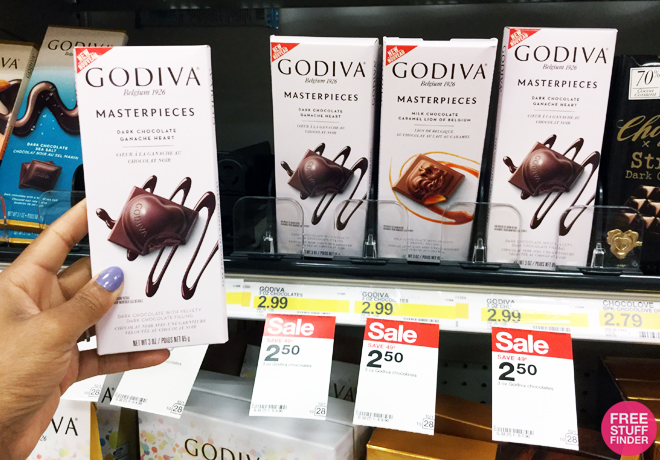}
            \scriptsize
            Query (medium)
        \end{minipage}
        \begin{minipage}[b]{.207\linewidth}
            \centering
            \includegraphics[width=\textwidth]{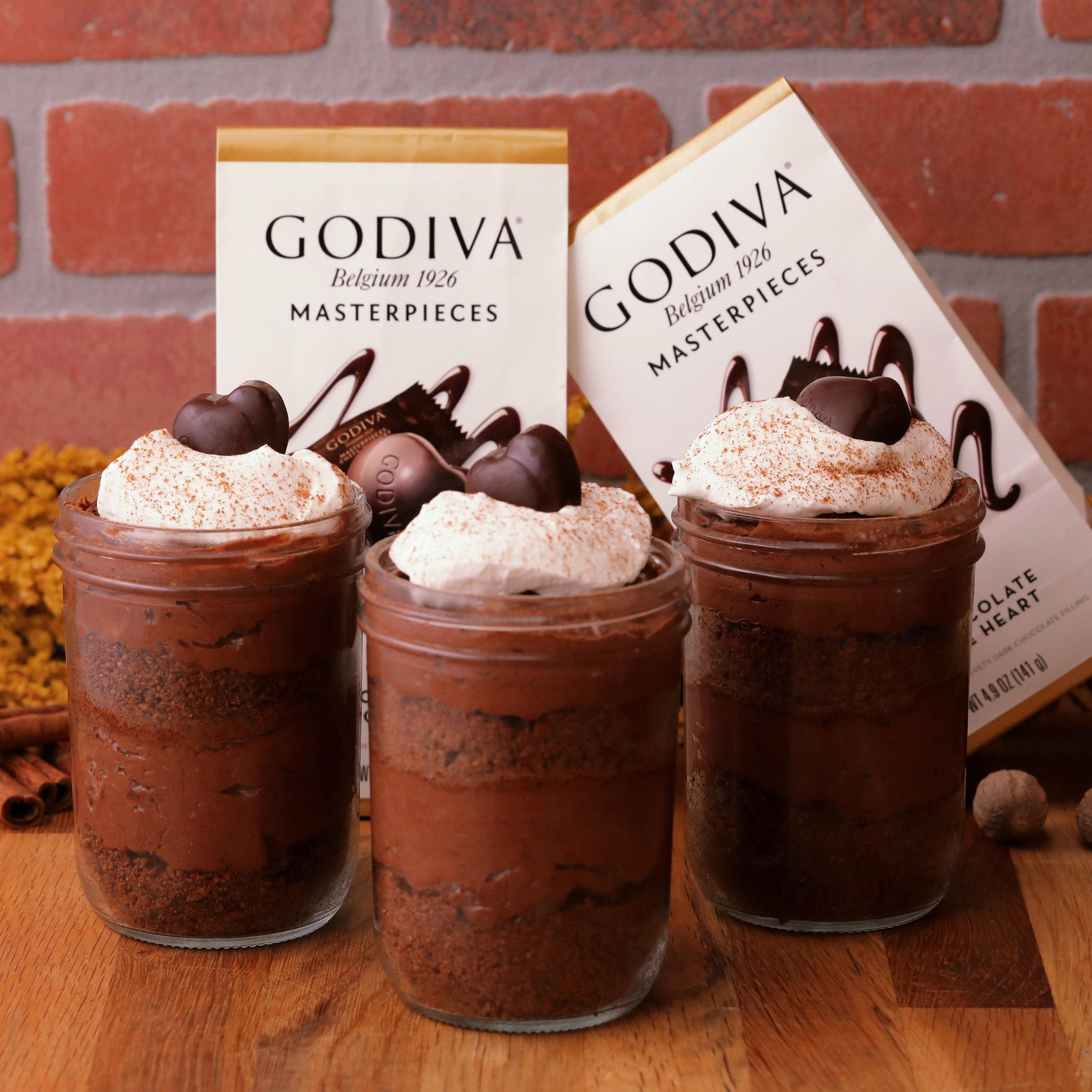}
            \scriptsize
            Query (hard)
        \end{minipage}
        \begin{minipage}[b]{.0896\linewidth}
            \centering
            \includegraphics[width=\textwidth]{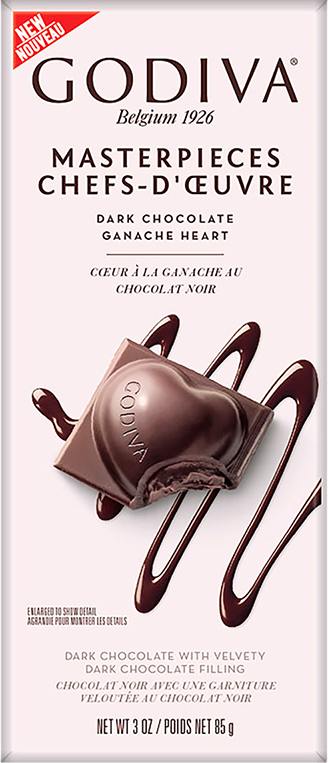}
            \scriptsize
            Index
        \end{minipage}
    \end{center}
    \vspace{-0.22cm}
    \caption{Example images of \textbf{packaged goods}. In the query image with ``hard'' difficulty, the target packaged goods is heavily occluded.}
    \label{fig:packaged_goods}
\end{figure}

\begin{figure}[h!]
    \begin{center}
        \begin{minipage}[b]{.103\linewidth}
            \centering
            \includegraphics[width=\textwidth]{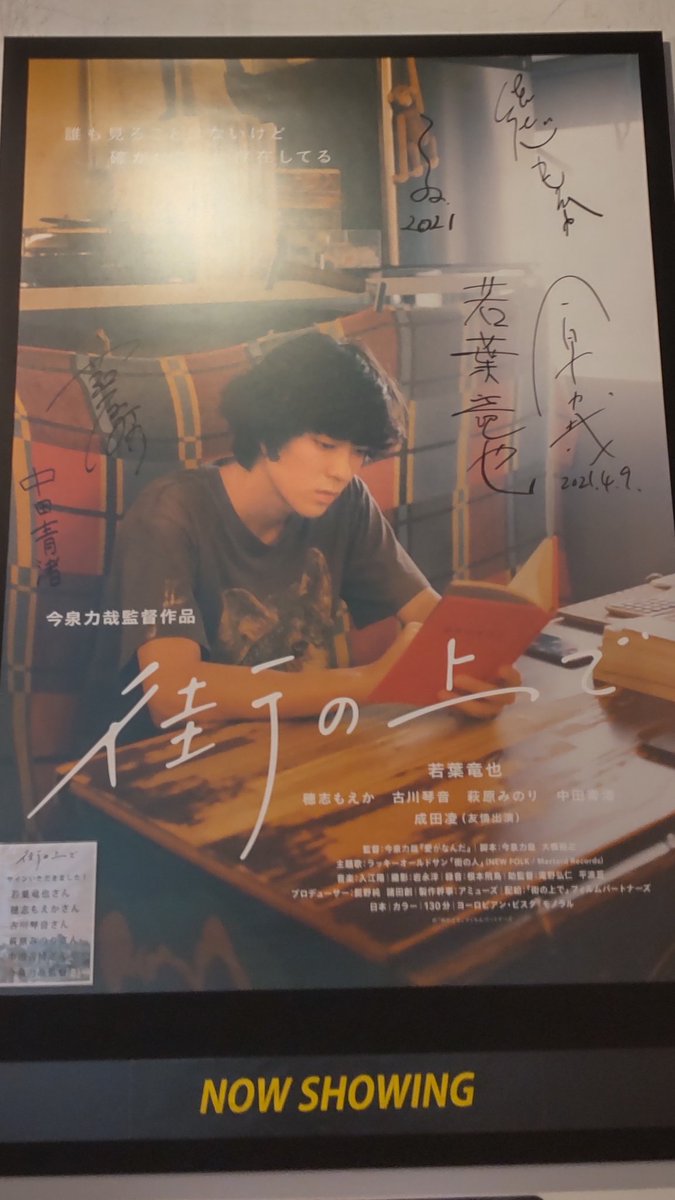}
            \scriptsize
            Query (easy)
        \end{minipage}
        \begin{minipage}[b]{.244\linewidth}
            \centering
            \includegraphics[width=\textwidth]{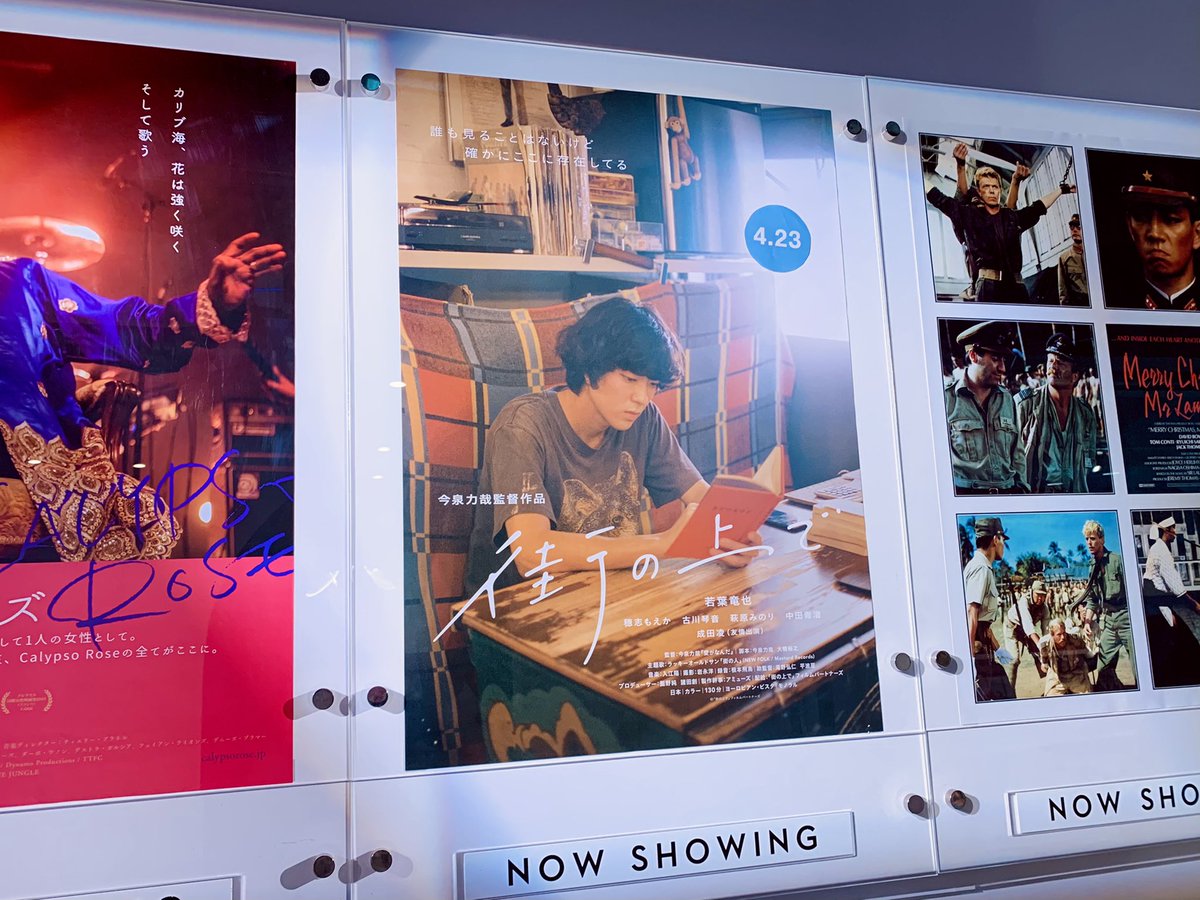}
            \scriptsize
            Query (medium)
        \end{minipage}
        \begin{minipage}[b]{.3290\linewidth}
            \centering
            \includegraphics[width=\textwidth]{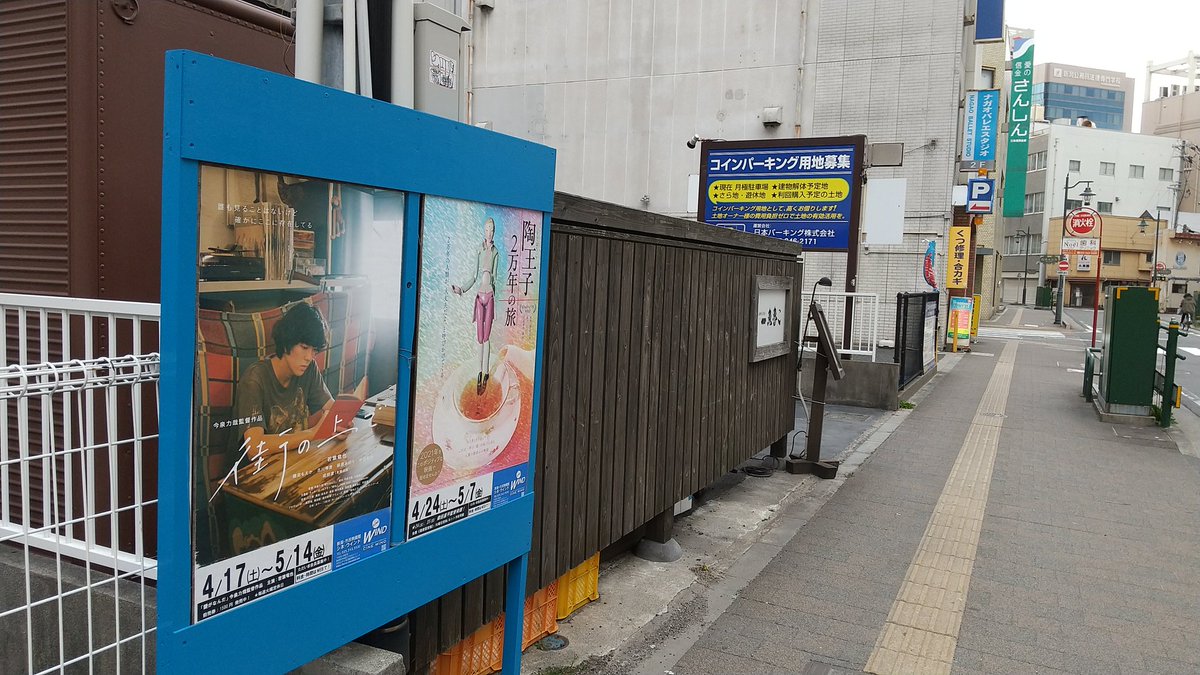}
            \scriptsize
            Query (hard)
        \end{minipage}
        \begin{minipage}[b]{.1345\linewidth}
            \centering
            \includegraphics[width=\textwidth]{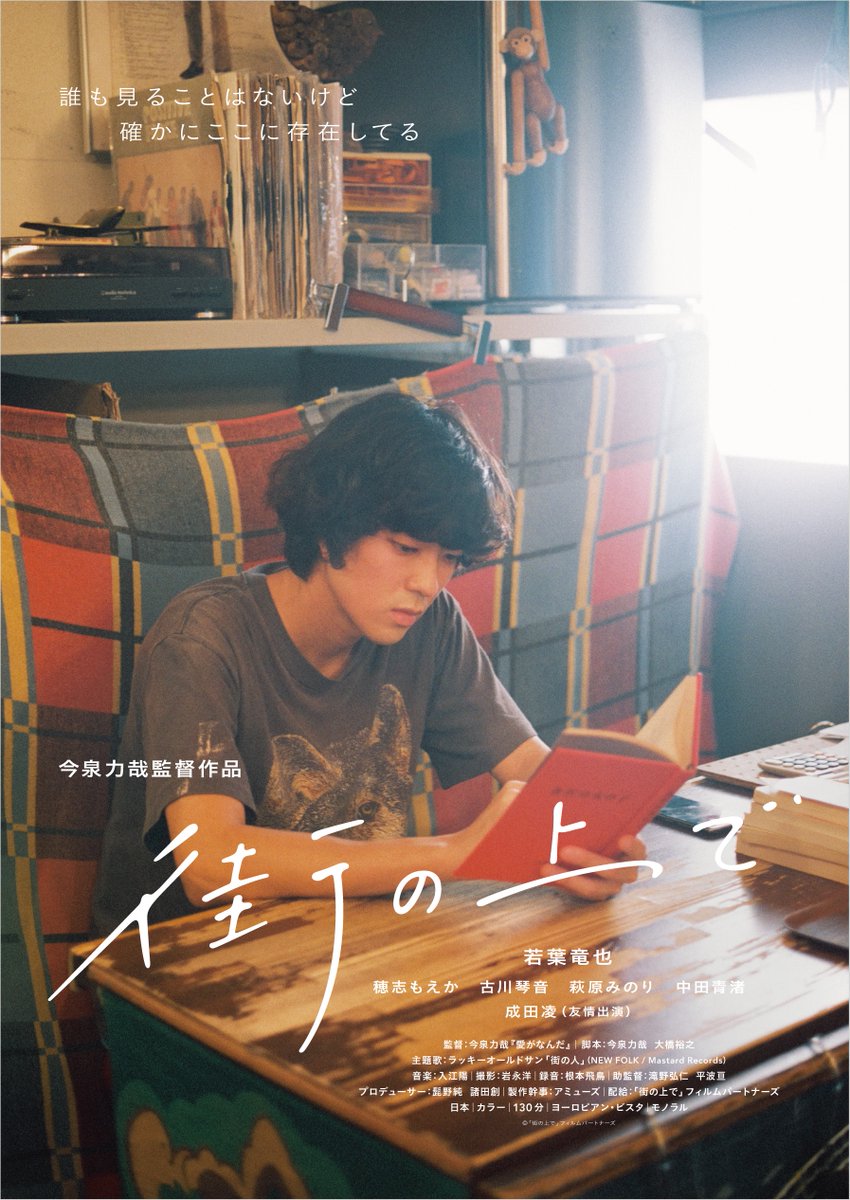}
            \scriptsize
            Index
        \end{minipage}
    \end{center}
    \caption{Example images of \textbf{movie poster}. In the query image with ``hard'' difficulty, the target movie poster only occupies a small area and is under perspective transformation.}
    \label{fig:movie}
\end{figure}

In Figures~\ref{fig:pokemon} - \ref{fig:movie} we provide additional sample images from our FORB benchmark. 
For each flat object type, we showcase query images of different difficulties and their corresponding index image.
Query images with difficulty ``hard'' overall present the greatest retrieval challenge, due to truncation, occlusion, perspective transformation, and the distraction of background. Note that in our benchmark, for objects with the same pattern but in different colors, we consider them equivalent and matched; see the easy query and index images in Figure~\ref{fig:logo}.

\subsection{Distractor Images}

\begin{figure}[t!]
	\begin{center}
            \begin{minipage}[b]{.355\linewidth}
                \centering
                \includegraphics[width=\textwidth]{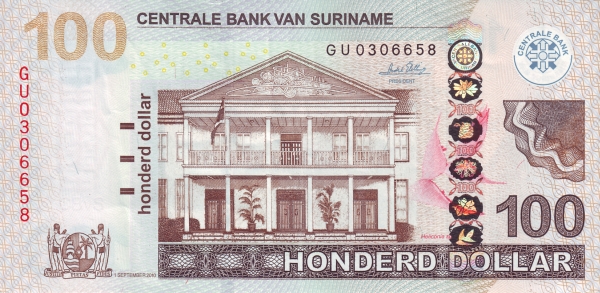}
                \scriptsize
                Index
            \end{minipage}
            \begin{minipage}[b]{.350\linewidth}
                \centering
                \includegraphics[width=\textwidth]{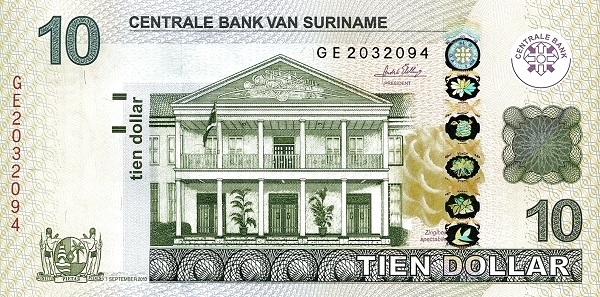}
                \scriptsize
                Distractor
            \end{minipage}
            \begin{minipage}[b]{.275\linewidth}
                \centering
                \includegraphics[width=\textwidth]{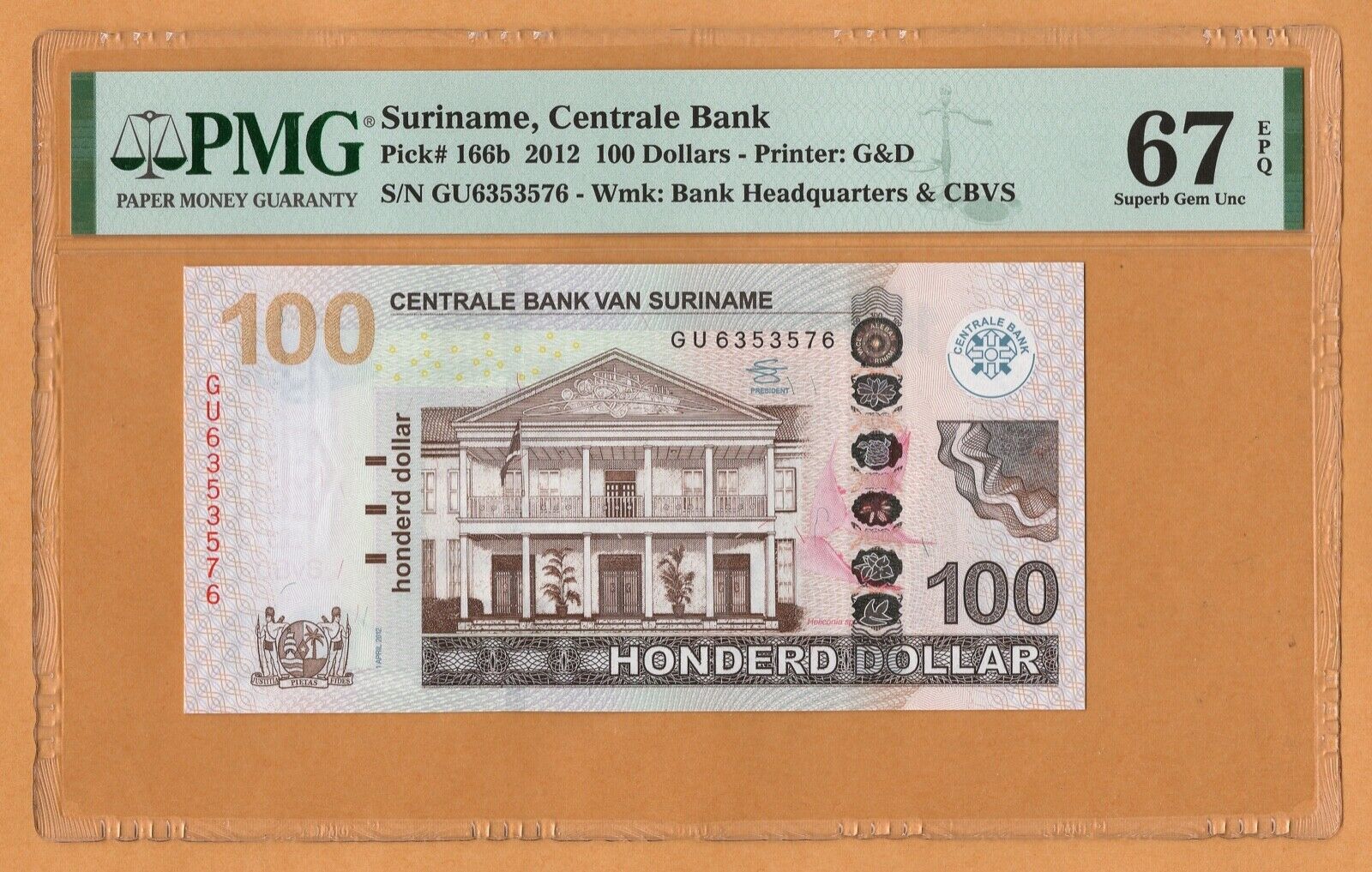}
                \scriptsize
                Query
            \end{minipage}
            \\
            \begin{minipage}[b]{\linewidth}
                \centering
                \small
                (a) Index and distractor images share similar content.
            \end{minipage}
            \vspace{0.01cm}
            \\
            \begin{minipage}[b]{.145\linewidth}
                \centering
                \includegraphics[width=\textwidth]{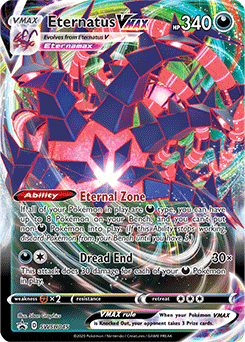}
                \scriptsize
                Index
            \end{minipage}
            \begin{minipage}[b]{.145\linewidth}
                \centering
                \includegraphics[width=\textwidth]{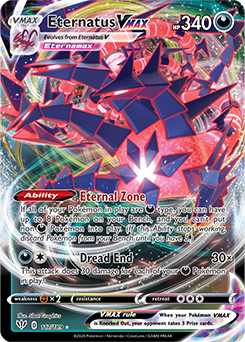}
                \scriptsize
                Distractor
            \end{minipage}
            \begin{minipage}[b]{.203\linewidth}
                \centering
                \includegraphics[width=\textwidth]{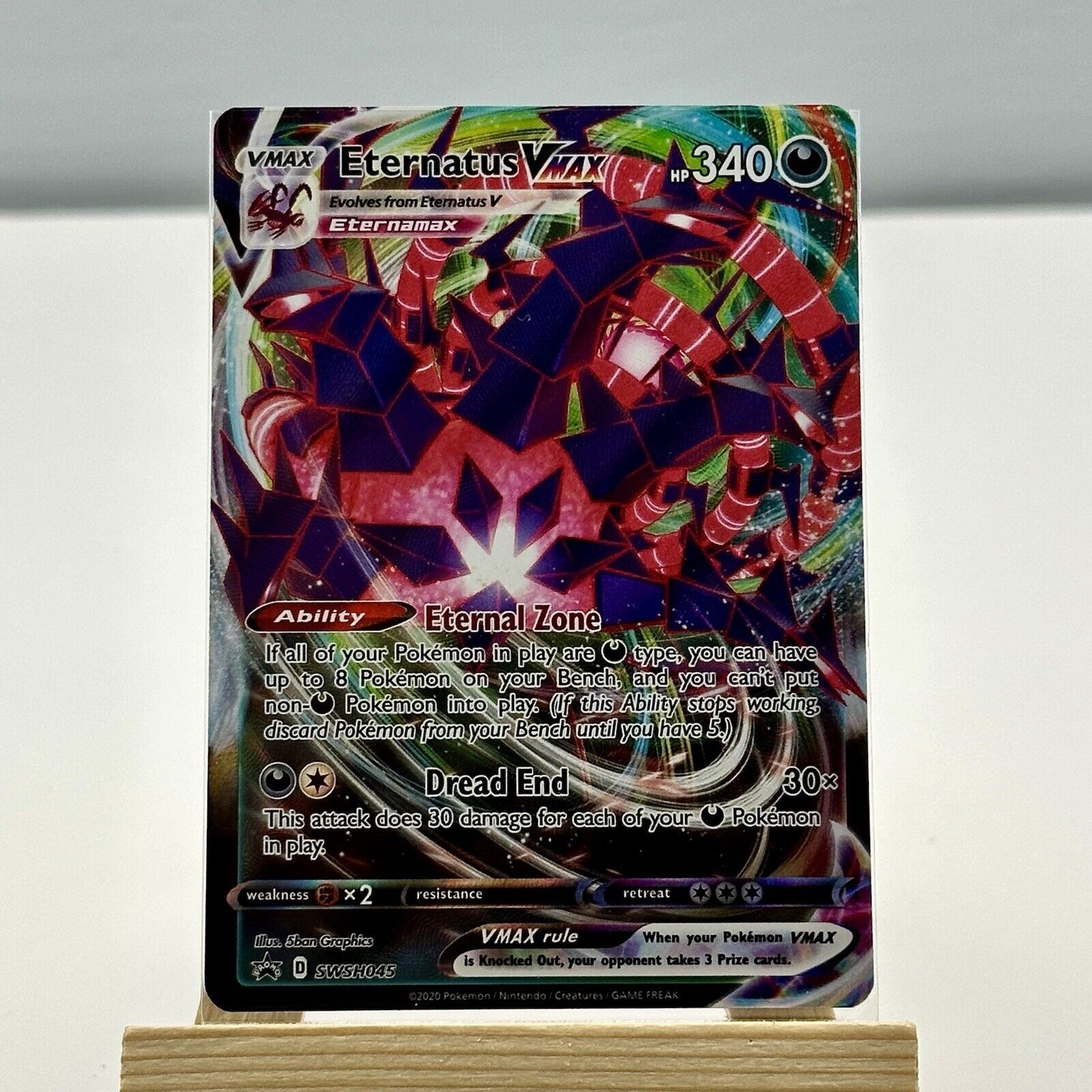}
                \scriptsize
                Query
            \end{minipage}
            ~
            \begin{minipage}[b]{.1515\linewidth}
			\centering
			\includegraphics[width=\textwidth]{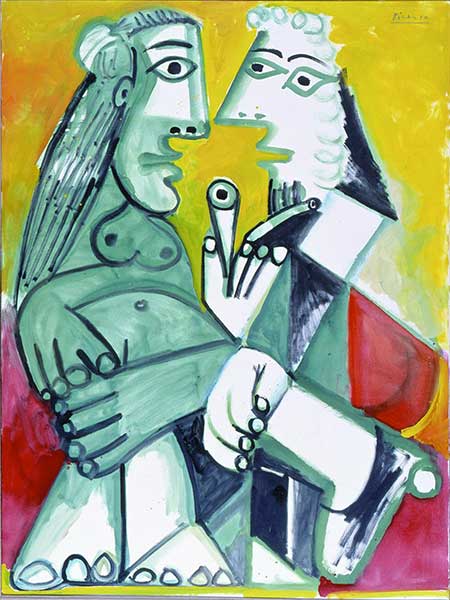}
			\scriptsize
			Index
		\end{minipage}
            \begin{minipage}[b]{.136\linewidth}
			\centering
			\includegraphics[width=\textwidth]{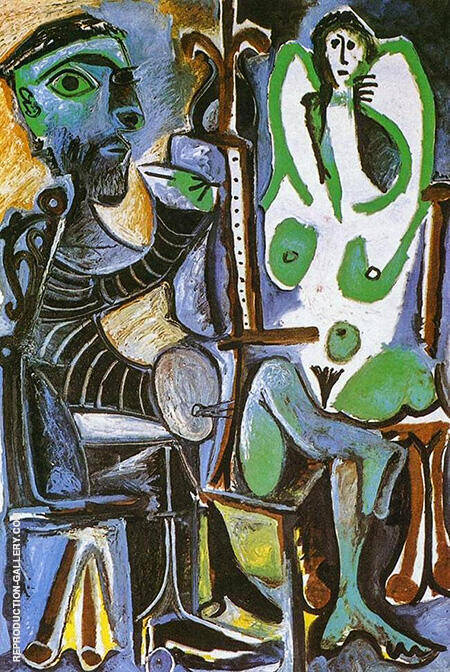}
			\scriptsize
			Distractor
		\end{minipage}
            \begin{minipage}[b]{.1635\linewidth}
			\centering
			\includegraphics[width=\textwidth]{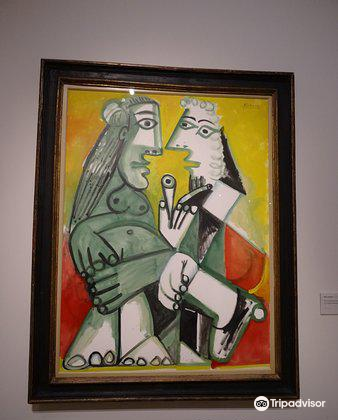}
			\scriptsize
			Query
		\end{minipage}
            \\
            \begin{minipage}[b]{.493\linewidth}
                \centering
                \small
                (b) Index and distractor images share similar content.
            \end{minipage}~
            \begin{minipage}[b]{.454\linewidth}
                \centering
                \small
                (c) Index and distractor images share similar style.
            \end{minipage}
            \vspace{0.01cm}
            \\
            \begin{minipage}[b]{.150\linewidth}
			\centering
			\includegraphics[width=\textwidth]{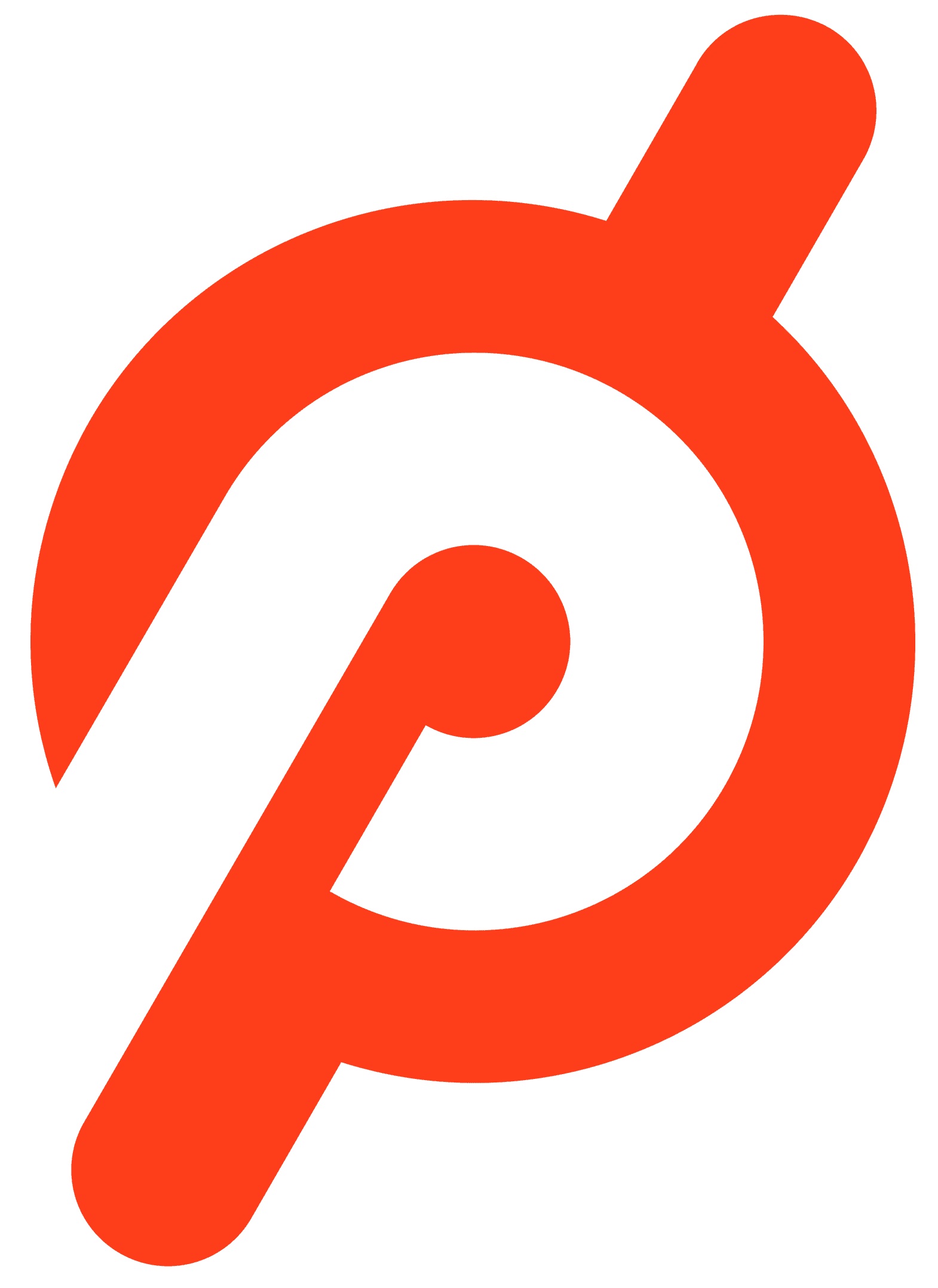}
			\scriptsize
			Index
		\end{minipage}
            \begin{minipage}[b]{.188\linewidth}
			\centering
			\includegraphics[width=\textwidth]{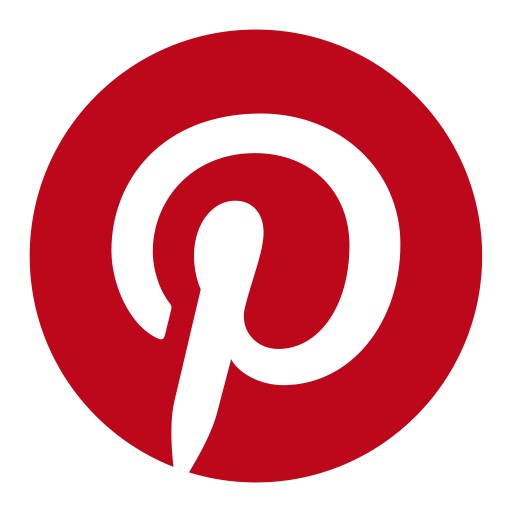}
			\scriptsize
			Distractor
		\end{minipage}
            \begin{minipage}[b]{.150\linewidth}
			\centering
			\includegraphics[width=\textwidth]{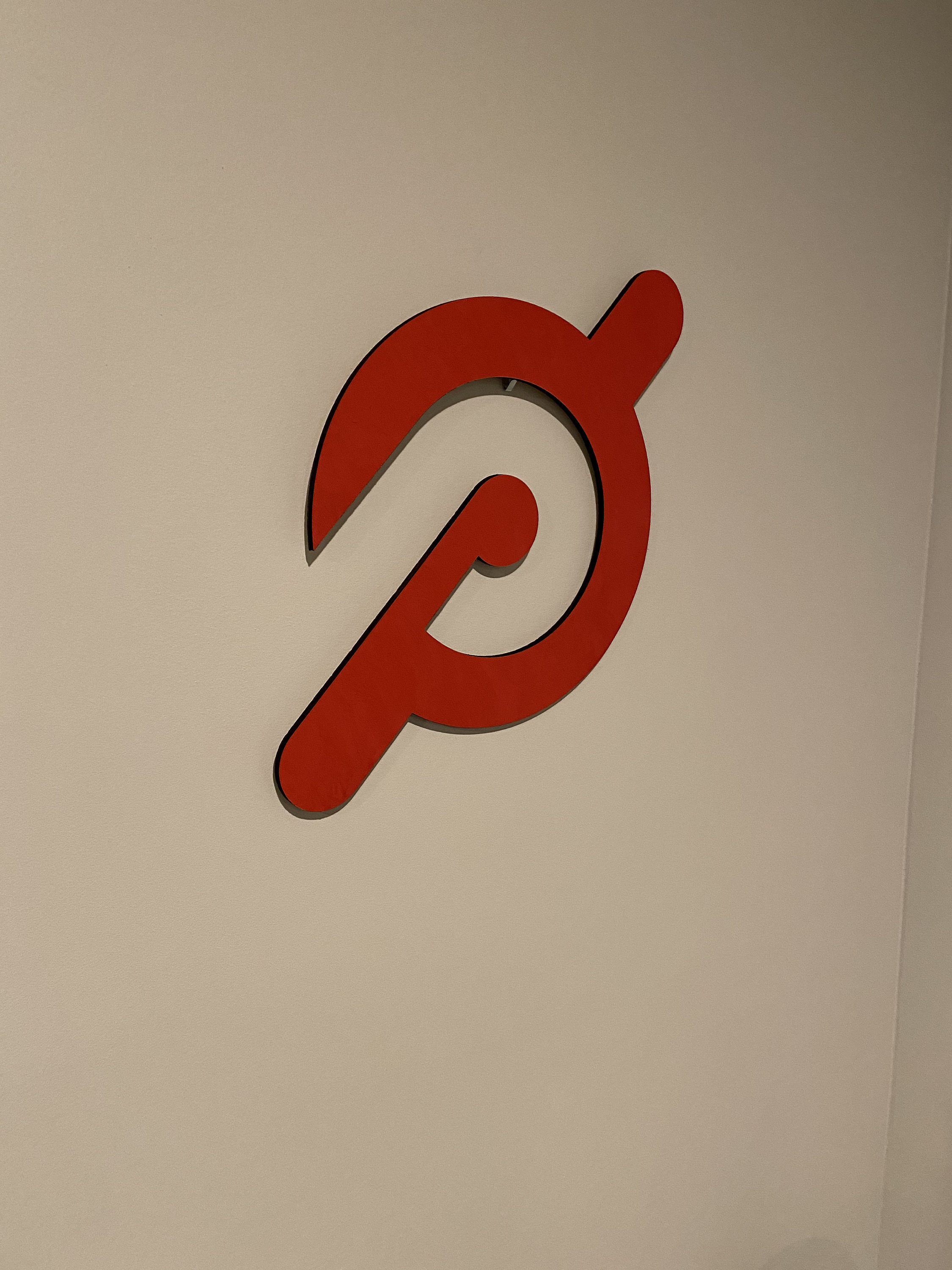}
			\scriptsize
			Query
		\end{minipage}
            ~
            \begin{minipage}[b]{.133\linewidth}
			\centering
			\includegraphics[width=\textwidth]{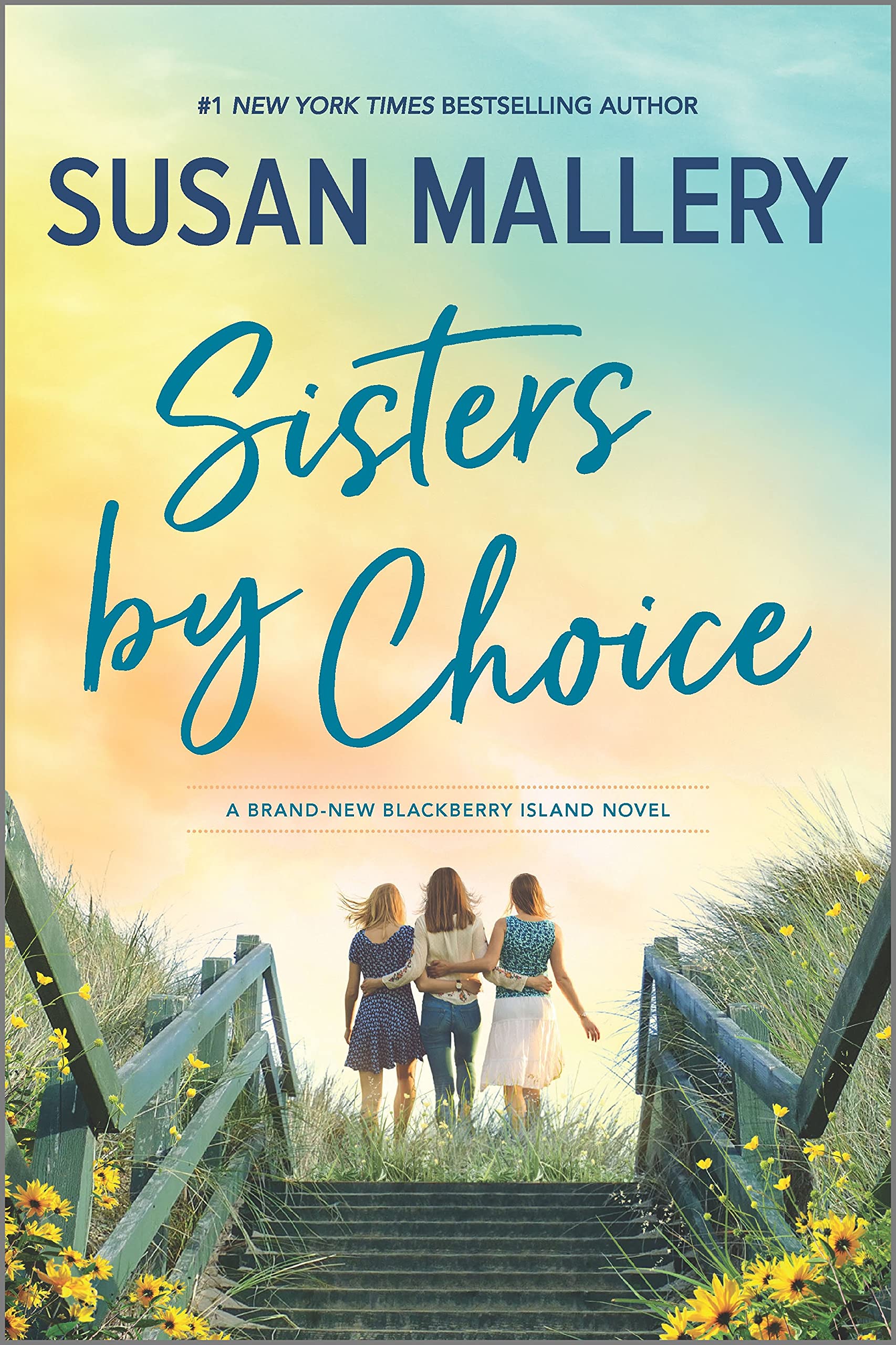}
			\scriptsize
			Index
		\end{minipage}
            \begin{minipage}[b]{.133\linewidth}
			\centering
			\includegraphics[width=\textwidth]{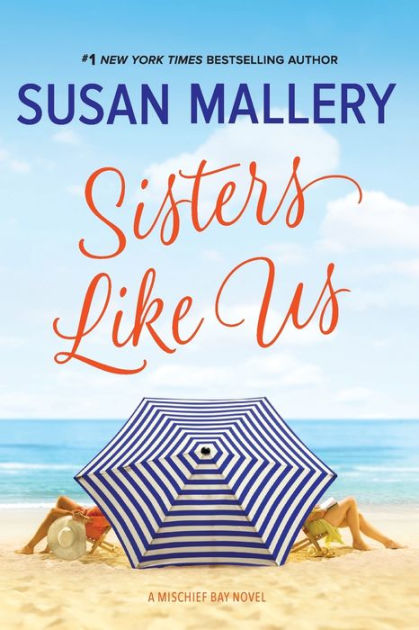}
			\scriptsize
			Distractor
		\end{minipage}
            \begin{minipage}[b]{.1775\linewidth}
			\centering
			\includegraphics[width=\textwidth]{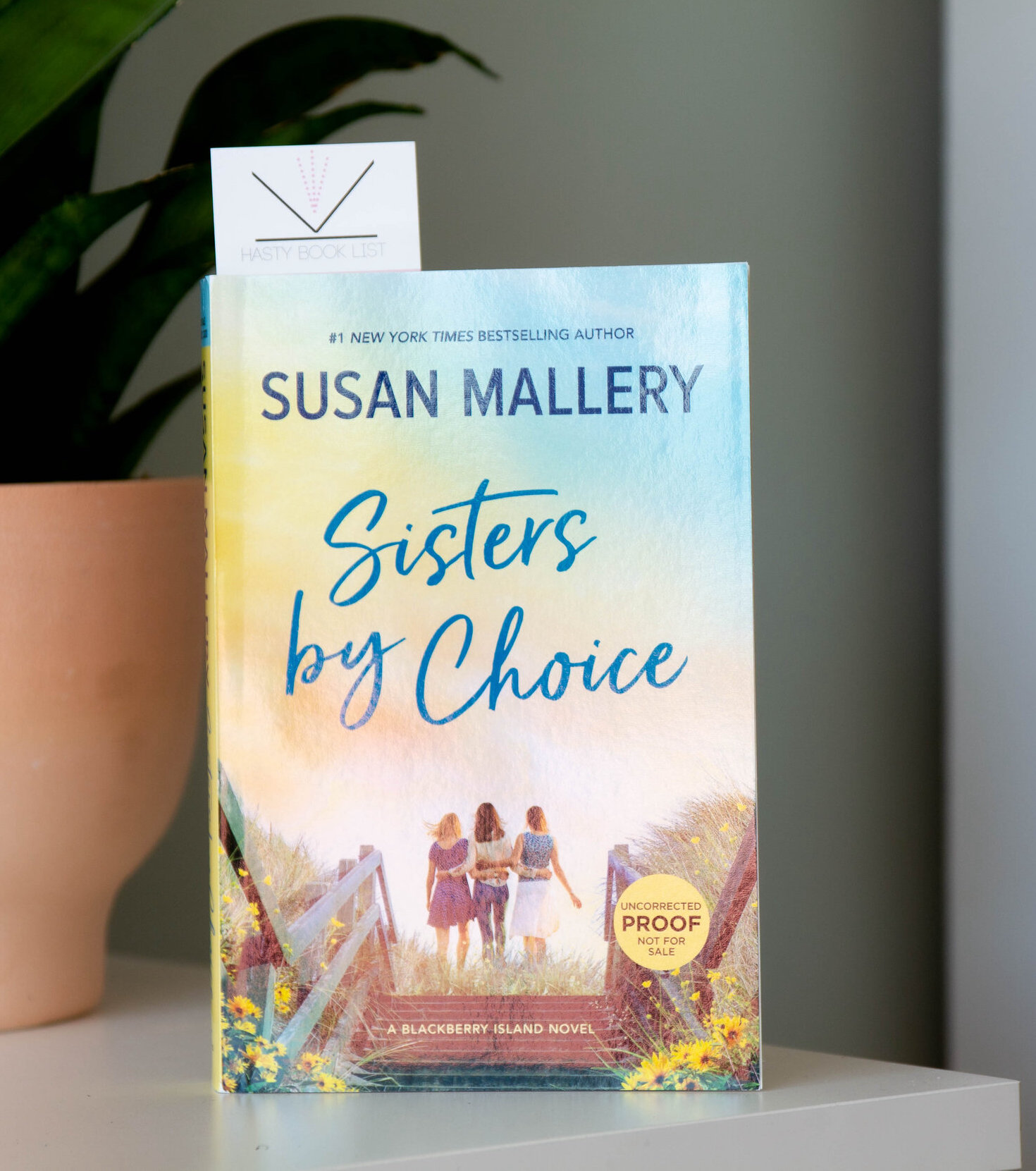}
			\scriptsize
			Query
		\end{minipage}
            \\
            \begin{minipage}[b]{.488\linewidth}
                \centering
                \small
                (d) Index and distractor images share similar shape.
            \end{minipage}
            ~
            \begin{minipage}[b]{.45\linewidth}
                \centering
                \small
                (e) Index and distractor images share similar texts.
            \end{minipage}
            \vspace{0.01cm}
            \\
            \begin{minipage}[b]{.160\linewidth}
			\centering
			\includegraphics[width=\textwidth]{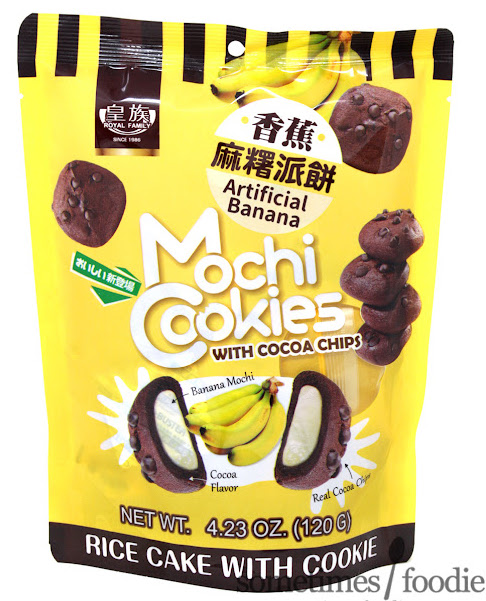}
			\scriptsize
			Index
		\end{minipage}
            \begin{minipage}[b]{.148\linewidth}
			\centering
			\includegraphics[width=\textwidth]{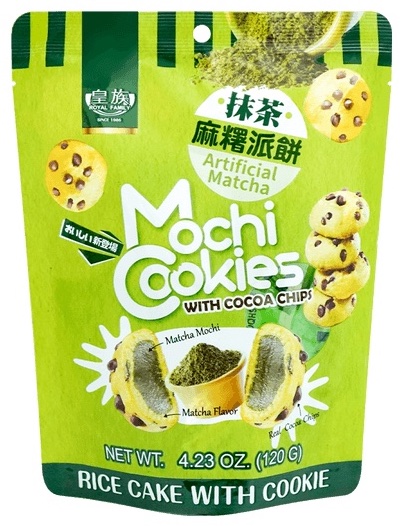}
			\scriptsize
			Distractor
		\end{minipage}
            \begin{minipage}[b]{.190\linewidth}
			\centering
			\includegraphics[width=\textwidth]{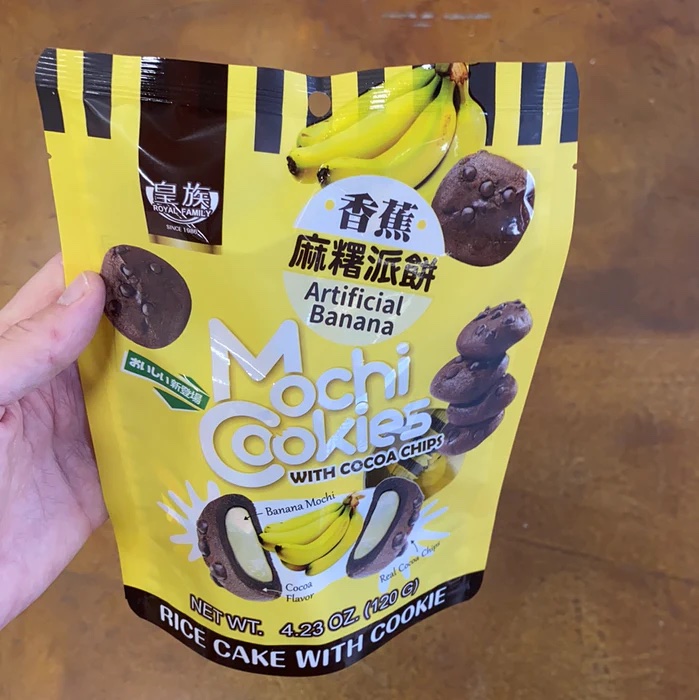}
			\scriptsize
			Query
		\end{minipage}
            ~
            \begin{minipage}[b]{.128\linewidth}
			\centering
			\includegraphics[width=\textwidth]{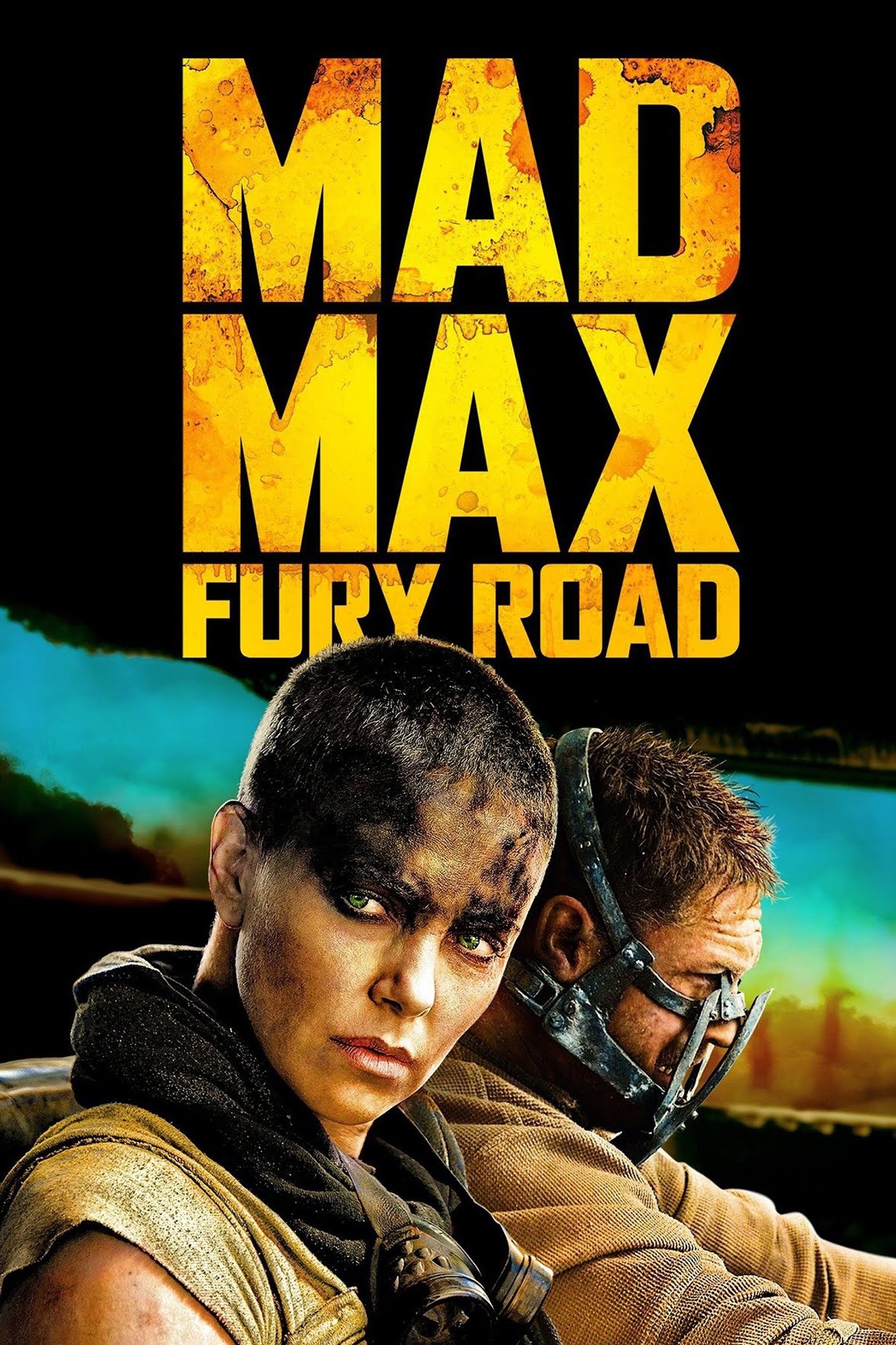}
			\scriptsize
			Index
		\end{minipage}
            \begin{minipage}[b]{.130\linewidth}
			\centering
			\includegraphics[width=\textwidth]{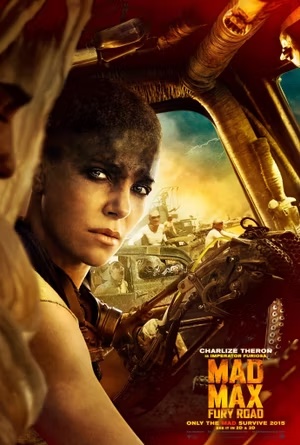}
			\scriptsize
			Distractor
		\end{minipage}
            \begin{minipage}[b]{.146\linewidth}
			\centering
			\includegraphics[width=\textwidth]{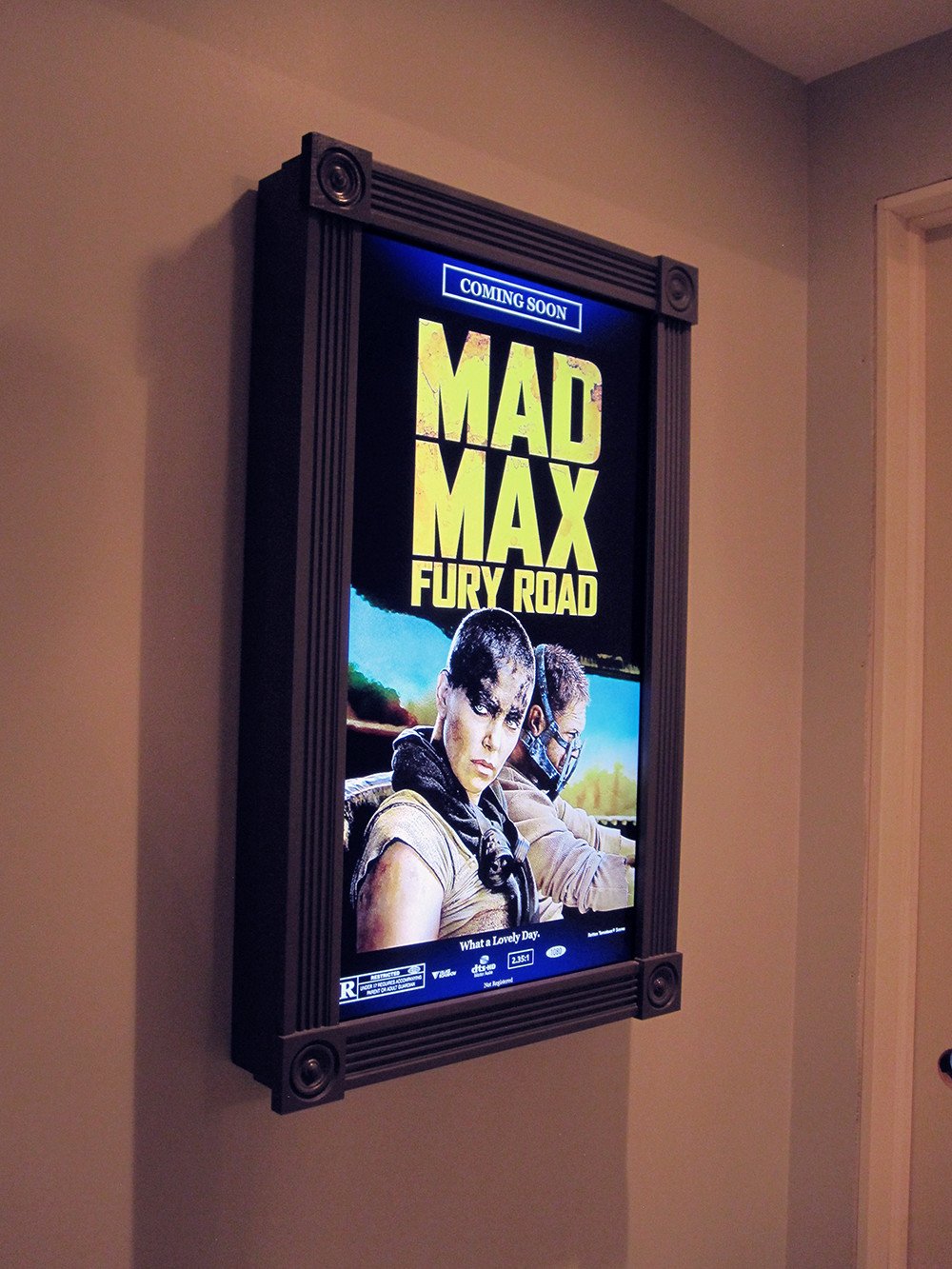}
			\scriptsize
			Query
		\end{minipage}
            \\
            \begin{minipage}[b]{.498\linewidth}
                \centering
                \small
                (f) Index and distractor images involve similar product.
            \end{minipage}
            ~
            \begin{minipage}[b]{.46\linewidth}
                \centering
                \small
                (g) Index and distractor images have similar content.
            \end{minipage}
            \vspace{0.01cm}
            \\
            \begin{minipage}[b]{.140\linewidth}
			\centering
			\includegraphics[width=\textwidth]{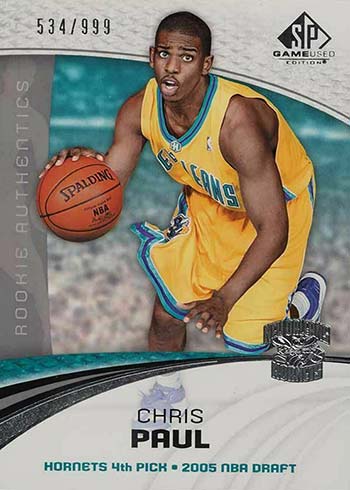}
			\scriptsize
			Index
		\end{minipage}
            \begin{minipage}[b]{.140\linewidth}
			\centering
			\includegraphics[width=\textwidth]{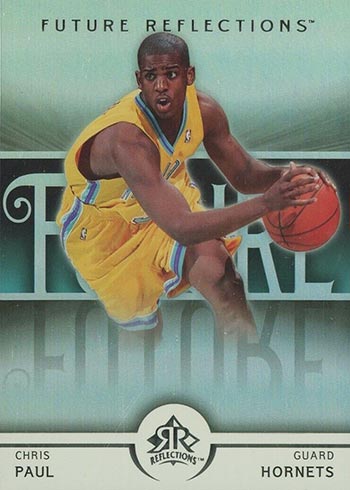}
			\scriptsize
			Distractor
		\end{minipage}
            \begin{minipage}[b]{.150\linewidth}
			\centering
			\includegraphics[width=\textwidth]{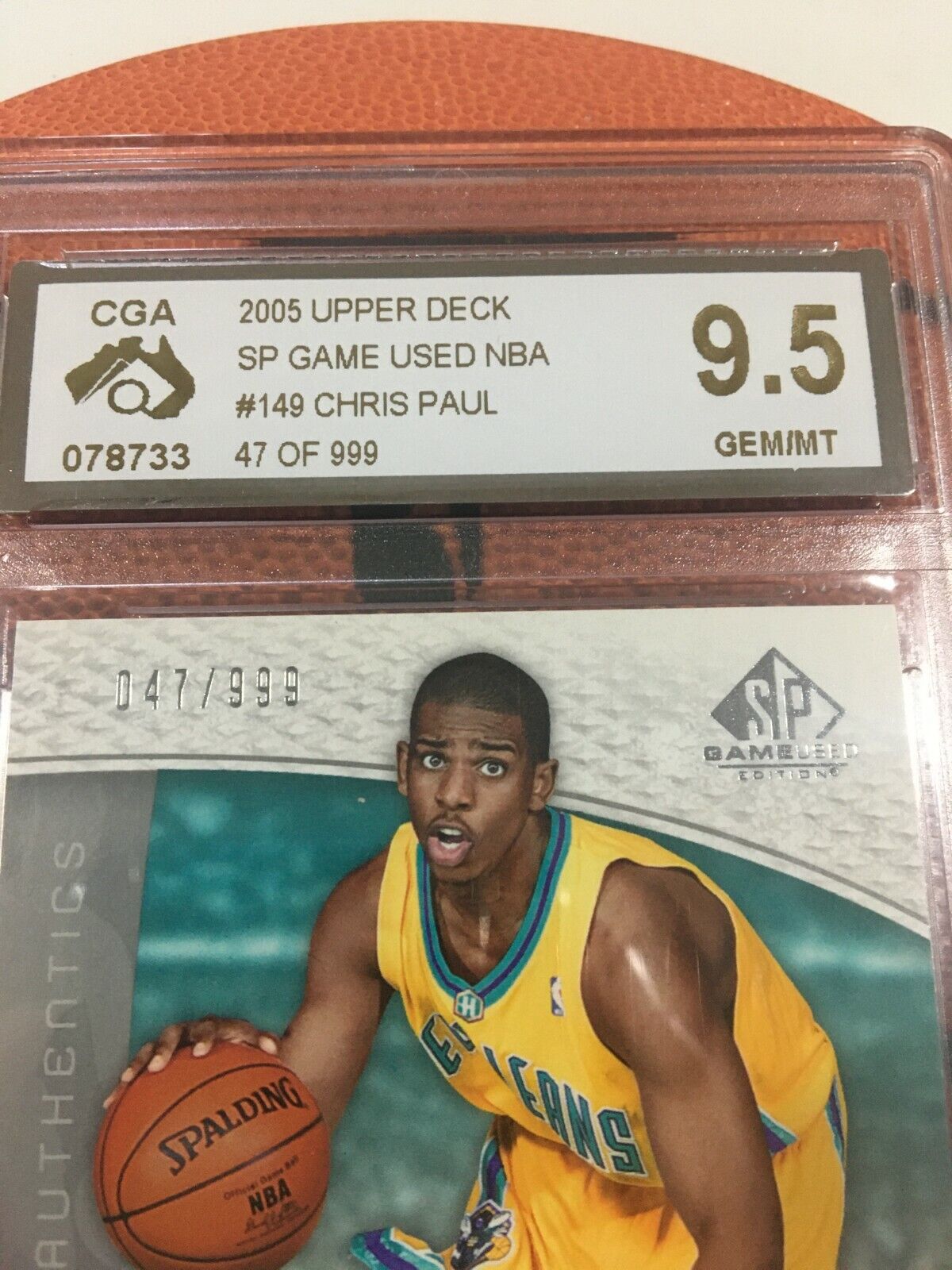}
			\scriptsize
			Query
		\end{minipage}
            ~
            \begin{minipage}[b]{.142\linewidth}
			\centering
			\includegraphics[width=\textwidth]{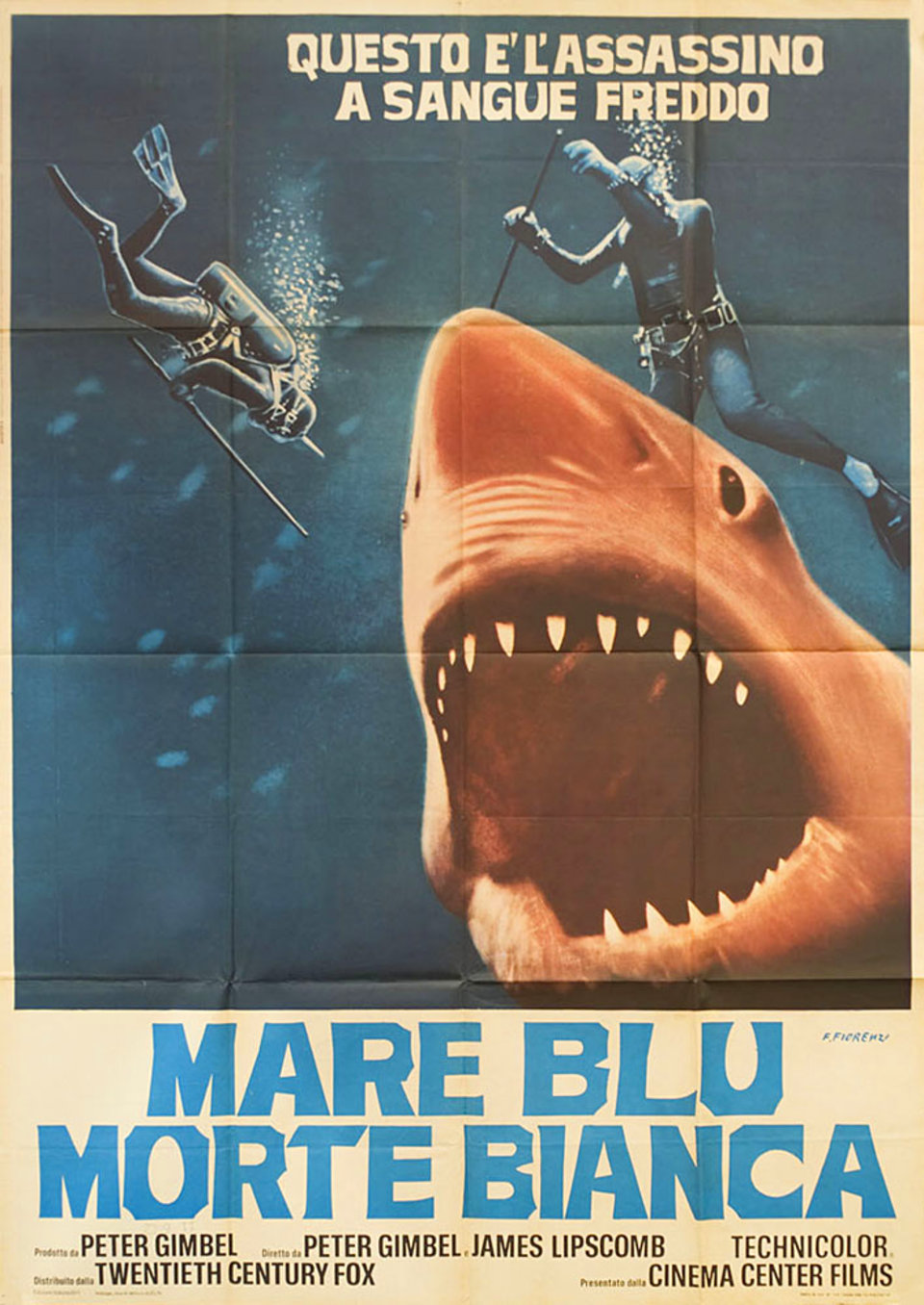}
			\scriptsize
			Index
		\end{minipage}
            \begin{minipage}[b]{.142\linewidth}
			\centering
			\includegraphics[width=\textwidth]{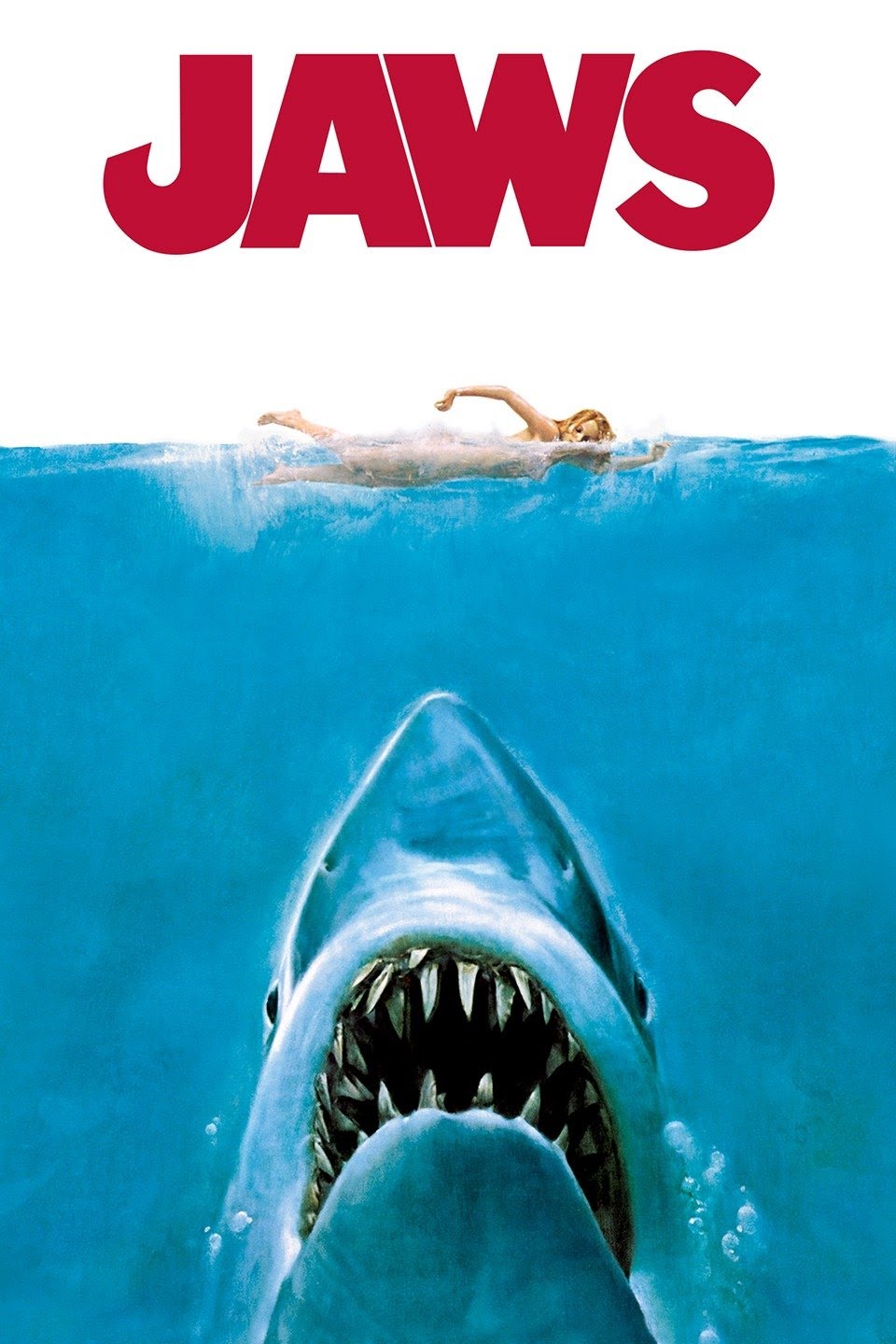}
			\scriptsize
			Distractor
		\end{minipage}
            \begin{minipage}[b]{.2\linewidth}
			\centering
			\includegraphics[width=\textwidth]{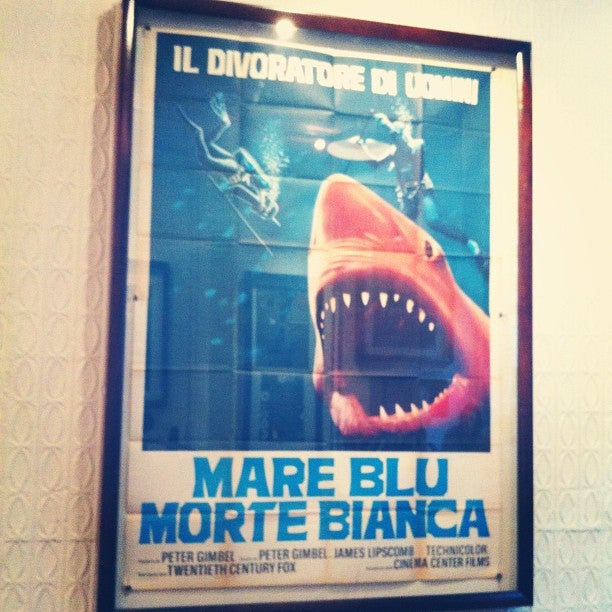}
			\scriptsize
			Query
		\end{minipage}
            \\
             \begin{minipage}[b]{.498\linewidth}
                \centering
                \small
                (h) Index and distractor images refer to the same person.
            \end{minipage}
            ~
            \begin{minipage}[b]{.46\linewidth}
                \centering
                \small
                (i) Index and distractor images contain same object.
            \end{minipage}
	\end{center}
	\caption{Examples of index and distractor images. We show distractor images that are similar to the index images in various aspects.
    }
	\label{fig:eg_distractor}
\end{figure}

In Figure~\ref{fig:eg_distractor}, we illustrate different distractor images that bear similarities to the index images in various aspects. For example, in Figures~\ref{fig:eg_distractor}(a)(b) both index and distractor images share very similar contents and textures. In Figure~\ref{fig:eg_distractor}(c), although the index and distractor images have different contents, they share similar styles. In Figure~\ref{fig:eg_distractor}(d), both logos are similar in shapes. In Figure~\ref{fig:eg_distractor}(e), the index and distractor images contain the same texts, which would pose challenges for text-sensitive methods such as CLIP \cite{clip}. In Figure~\ref{fig:eg_distractor}(f), the index and distractor images refer to similar products of the same brand. In Figure~\ref{fig:eg_distractor}(g), both index and distractor images are about the same movie and thus share the same semantics. This would pose challenges for top-only methods since their embeddings capture more about high-level semantics. Similarly, in Figures~\ref{fig:eg_distractor}(h)(i) both index and distractor images refer to the same person or object.

\subsection{Additional Evaluation Results on FORB}

In Tables~\ref{tab:exp_results_extra_1} and \ref{tab:breakdown_extra_tmp} we show additional benchmark results on FORB in terms of mAP@1 and $t$-mAP@1. Note that mAP@1 is essentially equivalent to Recall@1. We observe that FIRe overall achieves the best performance. This demonstrates its superior generalization ability on OOD domains.

\subsection{Evaluation Results on FORB Subset}

The baselines considered in our experiments are trained on various datasets, some of which overlap with our FORB dataset. In particular, we find a few images from FORB are also included in LAION-5B \cite{laion}, and as a result, training sets based on the subset of LAION-5B (\eg, LAION-438M \cite{laion} and 129M \cite{blip}) may also share duplicate images with FORB. To make FORB a real OOD query set and evaluate the generalization abilities of baselines trained on LAION-based datasets, we remove duplicate images shared with LAION-5B from FORB and compare baselines on the reduced FORB dataset. In total the number of duplicate images is small: only 5.91\% query images and 19.93\% database images are included in LAION-5B. 
In Tables~\ref{tab:exp_results_reduced_overall_5}-\ref{tab:breakdown_reduced_detail_1} we report the evaluation results on the reduced FORB benchmark.
Note that since the training data used by CLIP \cite{clip} are not publicly available, it is difficult to determine if FORB overlaps with it. Therefore, the performance of CLIP in Tables~\ref{tab:exp_results_reduced_overall_5}-\ref{tab:breakdown_reduced_detail_1} can be considered as an ``upper bound''.

In Tables~\ref{tab:exp_results_reduced_overall_5}-\ref{tab:breakdown_reduced_detail_1}, we observe that the results on reduced FORB are very similar to those from the original FORB dataset. This shows that duplicate images only have little impact on the evaluations.

\begin{table}[h]
    \caption{Comparison of different image retrieval methods on our FORB benchmark. We measure the performance in terms of mAP@1 and $t$-mAP@1. \textbf{Bolded} numbers indicate the \textbf{best} results. $\dagger$ means the model training data may overlap with FORB and the retrieval accuracy can be interpreted as an ``upper bound'' performance.}
    \label{tab:exp_results_extra_1}
    \resizebox{1.0\columnwidth}{!}{
        \centering
        \begin{tabular}{l|c|ccc|c|ccr}
        \hline
        & \multicolumn{4}{c|}{mAP@1 (\%)} & \multicolumn{4}{c}{$t$-mAP@1 (\%)} \\
        \hline
        Method & Overall & Easy & Medium & Hard  & Overall & Easy & Medium & Hard \\
        \hline
        BoW \cite{rootsift}            & 74.69 & 87.35 & 76.07 & 47.79 & 60.29 & 76.12 & 61.43 & 32.60\\
        BoW (+ rerank) \cite{rootsift} & 78.88 & 91.05 & 80.31 & 51.95 & 66.65 & 80.52 & 67.90 & 39.79\\
        FIRe \cite{fire} & \textbf{86.57} & \textbf{98.11} & \textbf{88.67} & 53.52 & \textbf{76.70} & \textbf{90.14} & \textbf{78.59} & \textbf{43.76}\\
        \hline
        DELG \cite{delg} & 44.88 & 75.73 & 44.07 & 21.42 & 32.91 & 63.02 & 31.74 & 13.84\\
        DELG (+ rerank) \cite{delg} & 58.33 & 87.82 & 57.99 & 31.46 & 39.26 & 70.52 & 38.24 & 17.55\\
        \hline
        CLIP$^\dagger$ \cite{clip} & 84.35 & 96.58 & 84.97 & \textbf{65.64} & 64.24 & 85.72 & 64.42 & 40.43\\
        SLIP \cite{slip} & 32.61 & 58.11 & 31.98 & 12.81 & 21.09 & 45.88 & 20.02 & ~~6.47 \\
        BLIP$^\dagger$ \cite{blip} & 68.04 & 91.56 & 68.54 & 38.85 & 46.74 & 79.02 & 46.23 & 18.84 \\
        BLIP2$^\dagger$ \cite{blip2} & 75.28 & 90.37 & 76.30 & 49.50 & 53.57 & 78.56 & 53.84 & 25.25\\
        DINO \cite{dino} & 51.79 & 83.05 & 51.64 & 21.29 & 40.62 & 73.11 & 39.99 & 13.60\\
        DINOv2 \cite{dinov2} & 65.19 & 90.98 & 65.74 & 33.25 & 46.35 & 70.93 & 46.50 & 19.69\\
        DiHT$^\dagger$ \cite{diht} & 79.70 & 93.94 & 80.41 & 57.92 & 57.90 & 81.86 & 58.33 & 29.01\\
        \hline
        \end{tabular}
    }
\end{table}

\begin{table}[h!]
    \caption{Retrieval accuracies on diverse objects. We report overall mAP@1 and $t$-mAP@1. \textbf{Bolded} numbers indicate the \textbf{best} results. $\dagger$ means the model training data may overlap with FORB and the retrieval accuracy can be interpreted as an ``upper bound'' performance.}
    \label{tab:breakdown_extra_tmp}
    \resizebox{1.0\columnwidth}{!}{
        \centering
        \begin{tabular}{l|cccccccc}
        \hline
        & \multicolumn{8}{c}{mAP@1 (\%) / $t$-mAP@1 (\%)}  \\
        \hline
        Method & \specialcell{Animated\\Card} & \specialcell{Photorealistic\\Card} & \specialcell{Book\\Cover} & \specialcell{Painting}  & Currency & Logo & \specialcell{Packaged\\Goods} & \specialcell{Movie\\Poster} \\
        \hline
        BoW \cite{rootsift}            & 82.07 / 67.92 & 75.95 / 60.72 & 85.90 / 71.45 & 69.94 / 52.21 & 63.46 / 48.50 & 25.73 / 18.01 & 86.12 / 69.58 & 70.31 / 52.04 \\
        BoW (+ rerank) \cite{rootsift} & 87.45 / 74.88 & 83.81 / 69.99 & 88.98 / 76.78 & 77.53 / 61.97 & 70.71 / 57.72 & 18.94 / 15.39 & 81.13 / 69.94 & 76.37 / 60.90\\
        FIRe \cite{fire} & \textbf{93.03} / \textbf{82.99} & \textbf{94.56} / \textbf{84.57} & 89.53 / \textbf{79.89} & 86.64 / \textbf{77.19} & 79.82 / 68.99 & 37.19 / 30.31 & 91.37 / \textbf{80.64} & \textbf{83.20} / \textbf{70.21}\\
        \hline
        DELG \cite{delg} & 49.44 / 40.46 & 39.37 / 23.45 & 55.51 / 38.42 & 26.32 / 15.07 & 61.87 / 46.03 & 10.75 / ~~6.87 & 66.25 / 45.00 & 39.65 / 24.34 \\
        DELG (+ rerank) \cite{delg} & 64.68 / 50.24 & 55.42 / 28.31 & 67.62 / 42.33 & 39.37 / 18.11 & 73.48 / 50.71 & 17.69 / ~~9.25 & 75.50 / 47.95 & 49.61 / 26.97\\
        \hline
        CLIP$^\dagger$ \cite{clip} & 88.17 / 70.45 & 66.58 / 49.27 & \textbf{98.63} / 71.18 & 90.18 / 62.27 & \textbf{82.06} / \textbf{69.92} & \textbf{68.72} / \textbf{45.21} & 96.75 / 78.14 & \textbf{83.20} / 52.57\\
        SLIP \cite{slip} & 28.37 / 21.13 & 39.14 / 29.25 & 40.31 / 20.65 & 47.57 / 24.25 & 23.35 / 19.05 & 11.58 / ~~2.69 & 53.50 / 26.34 & 32.81 / 17.63\\
        BLIP$^\dagger$ \cite{blip} & 59.07 / 47.65 & 67.63 / 54.20 & 92.13 / 55.18 & 74.39 / 43.40 & 62.40 / 46.49 & 66.37 / 24.77 & 94.50 / 50.38 & 65.04 / 31.43\\
        BLIP2$^\dagger$ \cite{blip2} & 71.62 / 59.69 & 71.15 / 53.58 & 95.35 / 61.61 & 80.06 / 41.47 & 70.98 / 51.32 & 64.26 / 26.78 & 96.25 / 62.70 & 68.36 / 32.33\\
        DINO \cite{dino} & 48.86 / 39.10 & 77.37 / 61.95 & 45.17 / 32.00 & 62.85 / 50.62 & 50.13 / 39.59 & ~~4.70 / ~~2.75 & 72.25 / 53.99 & 52.73 / 39.84\\
        DINOv2 \cite{dinov2} & 65.46 / 42.88 & 84.82 / 74.67 & 65.71 / 37.06 & 76.52 / 56.31 & 61.21 / 52.59 & ~~4.47 / ~~1.18 & 89.62 / 67.13 & 61.33 / 35.07\\
        DiHT$^\dagger$ \cite{diht} & 78.36 / 64.58 & 72.15 / 58.53 & 94.32 / 64.64 & \textbf{90.79} / 47.41 & 78.23 / 56.19 & 63.68 / 23.02 & \textbf{97.38} / 72.63 & 75.78 / 36.77\\
        \hline
        \end{tabular}
    }
\end{table}

\begin{table}[h]
    \caption{Comparison of different image retrieval methods on reduced FORB dataset. We measure the performance in terms of mAP@5 and $t$-mAP@5. \textbf{Bolded} numbers indicate the \textbf{best} results. $\dagger$ means the model training data may overlap with FORB and the retrieval accuracy can be interpreted as an ``upper bound'' performance.}
    \label{tab:exp_results_reduced_overall_5}
    \resizebox{1.0\columnwidth}{!}{
        \centering
        \begin{tabular}{l|c|ccc|c|ccr}
        \hline
        & \multicolumn{4}{c|}{mAP@5 (\%)} & \multicolumn{4}{c}{$t$-mAP@5 (\%)} \\
        \hline
        Method & Overall & Easy & Medium & Hard  & Overall & Easy & Medium & Hard \\
        \hline
        BoW \cite{rootsift}            & 81.20 & 92.90 & 82.00 & 57.92 & 67.06 & 82.58 & 67.68 & 41.35 \\
        BoW (+ rerank) \cite{rootsift} & 83.35 & 95.18 & 84.32 & 58.03 & 70.43 & 84.60 & 71.26 & 43.93 \\
        FIRe \cite{fire} & \textbf{90.22} & 98.18 & \textbf{91.97} & 60.47 & \textbf{79.32} & \textbf{89.83} & \textbf{81.00} & \textbf{47.34} \\
        \hline
        DELG \cite{delg} & 51.57 & 83.22 & 50.43 & 26.96 & 36.64 & 69.18 & 35.01 & 16.69 \\
        DELG (+ rerank) \cite{delg} & 62.31 & 91.58 & 61.63 & 35.35	& 41.60 & 74.72 & 40.07 & 19.94 \\
        \hline
        CLIP$^\dagger$ \cite{clip} & 89.59 & \textbf{98.55} & 90.14 & \textbf{72.51} & 67.64 & 87.03 & 67.67 & 44.10 \\
        SLIP \cite{slip} & 41.43 & 66.33 & 40.89 & 18.00 & 25.92 & 51.59 & 24.78 & ~~8.59 \\
        BLIP \cite{blip} & 74.31 & 95.00 & 74.80 & 44.04 & 50.31 & 81.17 & 49.73 & 20.21 \\
        BLIP2 \cite{blip2} & 82.62 & 94.94 & 83.52 & 57.58 & 58.29 & 82.12 & 58.40 & 28.63 \\
        DINO \cite{dino} & 58.36 & 85.61 & 58.45 & 24.85 & 44.56 & 74.51 & 44.03 & 14.97\\
        DINOv2 \cite{dinov2} & 72.27 & 93.88 & 72.89 & 39.41 & 49.90 & 70.27 & 50.23 & 21.75\\
        DiHT \cite{diht} & 85.48 & 97.39 & 85.98 & 65.50 & 62.14 & 84.35 & 62.50 & 31.41 \\
        \hline
        \end{tabular}
    }
\end{table}

\begin{table}[h]
    \caption{Comparison of different image retrieval methods on reduced FORB dataset. We measure the performance in terms of mAP@1 and $t$-mAP@1. \textbf{Bolded} numbers indicate the \textbf{best} results. $\dagger$ means the model training data may overlap with FORB and the retrieval accuracy can be interpreted as an ``upper bound'' performance.}
    \label{tab:exp_results_reduced_overall_1}
    \resizebox{1.0\columnwidth}{!}{
        \centering
        \begin{tabular}{l|c|ccc|c|ccr}
        \hline
        & \multicolumn{4}{c|}{mAP@1 (\%)} & \multicolumn{4}{c}{$t$-mAP@1 (\%)} \\
        \hline
        Method & Overall & Easy & Medium & Hard  & Overall & Easy & Medium & Hard \\
        \hline
        BoW \cite{rootsift}            & 77.69 & 90.81 & 78.45 & 53.29 & 65.53 & 81.20 & 66.15 & 39.68 \\
        BoW (+ rerank) \cite{rootsift} & 82.04 & 94.09 & 83.02 & 56.36 & 69.40 & 83.69 & 70.23 & 42.76 \\
        FIRe \cite{fire} & \textbf{88.81} & \textbf{98.01} & \textbf{90.60} & 57.04 & \textbf{78.53} & \textbf{89.69} & \textbf{80.19} & \textbf{45.98} \\
        \hline
        DELG \cite{delg} & 47.36 & 79.05 & 46.10 & 24.20 & 34.46 & 66.35 & 32.81 & 15.59 \\
        DELG (+ rerank) \cite{delg} & 61.90 & 91.45 & 61.18 & 35.07 & 41.39 & 74.60 & 39.84 & 19.80 \\
        \hline
        CLIP$^\dagger$ \cite{clip} & 85.40 & 97.62 & 85.90 & \textbf{65.08} & 65.13 & 86.26 & 65.07 & 40.62 \\
        SLIP \cite{slip} & 34.78 & 60.60 & 34.00 & 13.02 & 22.45 & 47.65 & 21.22 & ~~6.67 \\
        BLIP \cite{blip} & 69.03 & 92.74 & 69.38 & 36.75 & 47.38 & 79.50 & 46.66 & 17.40 \\
        BLIP2 \cite{blip2} & 76.91 & 91.97 & 77.73 & 49.54 & 55.03 & 79.76 & 55.05 & 25.43 \\
        DINO \cite{dino} & 54.87 & 83.48 & 54.74 & 22.36 & 42.91 & 73.08 & 42.29 & 14.16 \\
        DINOv2 \cite{dinov2} & 68.64 & 92.35 & 69.10 & 35.07 & 48.08 & 69.54 & 48.29 & 20.10 \\
        DiHT \cite{diht} & 81.05 & 95.69 & 81.43 & 59.11 & 59.72 & 83.06 & 59.95 & 29.34 \\
        \hline
        \end{tabular}
    }
\end{table}

\begin{table}[h!]
    \caption{Object retrieval accuracies on reduced FORB dataset. We report overall mAP@5 and $t$-mAP@5. \textbf{Bolded} numbers indicate the \textbf{best} results. $\dagger$ means the model training data may overlap with FORB and the retrieval accuracy can be interpreted as an ``upper bound'' performance.}
    \label{tab:breakdown_reduced_detail_5}
    \resizebox{1.0\columnwidth}{!}{
        \centering
        \begin{tabular}{l|cccccccc}
        \hline
        & \multicolumn{8}{c}{mAP@5 (\%) / $t$-mAP@5 (\%)}  \\
        \hline
        Method & \specialcell{Animated\\Card} & \specialcell{Photorealistic\\Card} & \specialcell{Book\\Cover} & \specialcell{Painting}  & Currency & Logo & \specialcell{Packaged\\Goods} & \specialcell{Movie\\Poster} \\
        \hline
        BoW \cite{rootsift}            & 86.57 / 73.77 & 80.09 / 63.65 & 89.48 / 76.53 & 68.78 / 55.51 & 74.92 / 59.27 & 30.22 / 15.83 & 89.36 / 73.45 & 72.81 / 55.99 \\
        BoW (+ rerank) \cite{rootsift} & 89.75 / 76.51 & 85.22 / 70.63 & 90.99 / 79.08 & 72.46 / 58.85 & 76.66 / 62.60 & 20.18 / 16.09 & 84.52 / 73.16 & 74.70 / 59.63 \\
        FIRe \cite{fire} & \textbf{93.67} / \textbf{83.15} & \textbf{96.02} / \textbf{85.16} & 91.04 / \textbf{80.53} & 91.59 / \textbf{80.11} & 82.37 / 69.69 & 41.58 / 32.22 & 93.25 / \textbf{81.43} & 85.26 / \textbf{70.60} \\
        \hline
        DELG \cite{delg} & 54.92 / 43.82 & 44.54 / 24.91 & 60.37 / 39.59 & 35.77 / 18.60 & 65.88 / 46.42 & 12.46 / ~~7.85 & 71.32 / 48.02 & 40.58 / 23.56 \\
        DELG (+ rerank) \cite{delg} & 66.61 / 51.06 & 56.55 / 28.27 & 70.17 / 42.61 & 46.96 / 21.55 & 74.33 / 49.27 & 18.66 / 10.04 & 77.55 / 49.96 & 50.84 / 25.98 \\
        \hline
        CLIP$^\dagger$ \cite{clip} & 92.78 / 73.34 & 75.98 / 55.09 & \textbf{99.38} / 71.89 & 92.91 / 62.60 & \textbf{89.74} / \textbf{74.85} & \textbf{86.57} / \textbf{52.03} & 98.42 / 78.95 & \textbf{87.42} / 51.99 \\
        SLIP \cite{slip} & 37.03 / 26.20 & 48.21 / 33.57 & 46.90 / 21.46 & 67.82 / 28.96 & 32.85 / 26.74 & 16.27 / ~~3.35 & 65.99 / 28.48 & 39.02 / 16.75 \\
        BLIP \cite{blip} & 66.55 / 51.71 & 74.98 / 57.87 & 94.05 / 52.80 & 83.22 / 43.47 & 72.98 / 51.29 & 83.33 / 25.35 & 97.18 / 47.54 & 68.69 / 28.02 \\
        BLIP2 \cite{blip2} & 79.64 / 64.86 & 80.00 / 58.24 & 96.66 / 59.46 & 88.18 / 40.03 & 82.22 / 55.48 & 82.56 / 27.26 & 97.98 / 59.97 & 73.71 / 29.30 \\
        DINO \cite{dino} & 54.57 / 42.35 & 80.09 / 62.91 & 49.40 / 33.36 & 69.49 / 54.88 & 54.38 / 41.49 & ~~5.96 / ~~2.89 & 77.50 / 56.48 & 52.29 / 38.04 \\
        DINOv2 \cite{dinov2} & 71.75 / 44.66 & 88.51 / 77.37 & 67.69 / 35.03 & 81.59 / 58.65 & 67.44 / 57.02 & ~~7.34 / ~~1.55 & 92.26 / 66.26 & 62.27 / 31.68 \\
        DiHT \cite{diht} & 84.56 / 67.97 & 80.11 / 63.44 & 96.21 / 63.69 & \textbf{94.04} / 45.86 & 87.69 / 59.59 & 78.65 / 22.42 & \textbf{98.88} / 71.26 & 80.65 / 33.23 \\
        \hline
        \end{tabular}
    }
\end{table}

\begin{table}[h!]
    \caption{Object retrieval accuracies on reduced FORB dataset. We report overall mAP@1 and $t$-mAP@1. $\dagger$ means the model training data may overlap with FORB and the retrieval accuracy can be interpreted as an ``upper bound'' performance. \textbf{Bolded} numbers indicate the \textbf{best} results.}
    \label{tab:breakdown_reduced_detail_1}
    \resizebox{1.0\columnwidth}{!}{
        \centering
        \begin{tabular}{l|cccccccc}
        \hline
        & \multicolumn{8}{c}{mAP@1 (\%) / $t$-mAP@1 (\%)}  \\
        \hline
        Method & \specialcell{Animated\\Card} & \specialcell{Photorealistic\\Card} & \specialcell{Book\\Cover} & \specialcell{Painting}  & Currency & Logo & \specialcell{Packaged\\Goods} & \specialcell{Movie\\Poster} \\
        \hline
        BoW \cite{rootsift}            & 83.14 / 71.69 & 75.62 / 62.18 & 87.51 / 76.26 & 67.15 / 55.51 & 69.64 / 56.92 & 26.33 / 15.17 & 86.90 / 72.88 & 69.92 / 55.48 \\
        BoW (+ rerank) \cite{rootsift} & 87.75 / 74.93 & 84.54 / 70.15 & 90.93 / 79.02 & 72.46 / 58.85 & 74.33 / 60.82 & 19.51 / 15.56 & 83.82 / 72.57 & 74.09 / 59.20 \\
        FIRe \cite{fire} & \textbf{92.43} / \textbf{82.37} & \textbf{95.19} / \textbf{84.66} & 89.87 / \textbf{80.00} & 89.86 / \textbf{79.23} & 80.80 / 69.20 & 36.71 / 29.40 & 91.91 / \textbf{80.78} & 83.01 / \textbf{69.69} \\
        \hline
        DELG \cite{delg} & 50.16 / 40.67 & 40.00 / 23.45 & 57.36 / 38.57 & 31.88 / 17.44 & 62.50 / 44.95 & ~~9.81 / ~~6.77 & 67.63 / 46.58 & 37.88 / 22.69 \\
        DELG (+ rerank) \cite{delg} & 66.26 / 50.83 & 56.32 / 28.21 & 69.96 / 42.51 & 46.38 / 21.48 & 73.88 / 48.99 & 17.13 / ~~9.37 & 76.69 / 49.52 & 50.70 / 25.96 \\
        \hline
        CLIP$^\dagger$ \cite{clip} & 89.57 / 71.29 & 68.65 / 50.63 & \textbf{99.06} / 71.67 & 89.86 / 61.57 & \textbf{84.60} / \textbf{71.04} & \textbf{75.91} / \textbf{46.76} & 97.11 / 78.03 & \textbf{83.57} / 50.70 \\
        SLIP \cite{slip} & 30.76 / 22.36 & 40.16 / 28.92 & 41.81 / 20.03 & 59.90 / 26.88 & 26.34 / 21.63 & 12.22 / ~~2.65 & 55.30 / 25.64 & 34.26 / 15.55 \\
        BLIP \cite{blip} & 60.97 / 48.12 & 68.49 / 53.77 & 92.93 / 52.59 & 79.23 / 42.38 & 66.96 / 48.36 & 72.60 / 22.94 & 95.76 / 47.10 & 65.18 / 27.27 \\
        BLIP2 \cite{blip2} & 73.31 / 60.43 & 73.68 / 54.69 & 95.64 / 59.31 & 84.06 / 39.13 & 75.00 / 52.17 & 71.32 / 24.59 & 96.72 / 59.43 & 69.36 / 28.34 \\
        DINO \cite{dino} & 50.61 / 40.26 & 77.51 / 61.76 & 45.82 / 32.13 & 67.15 / 53.93 & 50.45 / 39.87 & ~~4.41 / ~~2.43 & 73.03 / 54.55 & 48.75 / 36.44 \\
        DINOv2 \cite{dinov2} & 67.40 / 42.53 & 85.78 / 75.26 & 64.66 / 34.52 & 78.26 / 57.62 & 62.95 / 53.59 & ~~5.15 / ~~1.31 & 89.40 / 64.81 & 59.05 / 31.05 \\
        DiHT \cite{diht} & 80.25 / 65.21 & 74.38 / 59.78 & 95.29 / 63.37 & \textbf{91.79} / 45.28 & 82.14 / 57.16 & 68.05 / 20.11 & \textbf{98.07} / 70.82 & 76.04 / 32.36 \\
        \hline
        \end{tabular}
    }
\end{table}

\subsection{Index File Size}

In Table~\ref{tab:index_size} we provide the size of index files for each baseline method. In our implementation, we do not use approximate nearest neighbors (ANN) search for top-only methods, but instead simply use matrix multiplication to compute cosine similarities between image embeddings and sort the database candidates accordingly. In fact, we observe that naive GPU-based matrix multiplication already achieves decent search latency. As can be seen in Table~\ref{tab:index_size}, compared to top-only methods, both the bottom-up and top-down methods have a relatively larger size of index file. In particular, the index file size of BoW and DELG is orders of magnitude larger than others, making it difficult to scale up to larger database in practice. In contrast, FIRe requires a much smaller index file whose size is similar to those of top-only methods. Given the superior inference speed and retrieval accuracy, FIRe overall outperforms other methods and is suitable for deployment in real applications. Based on the aforementioned observations, we believe one promising future direction is to develop methods that combine the benefits of FIRe and CLIP-like approaches.

\begin{table}
  \caption{The index file sizes of different image retrieval methods.}
  \label{tab:index_size}
  \resizebox{1.0\columnwidth}{!}{
      \centering
      \begin{tabular}{c|ccccc}
        \hline
        Method        & BoW (+ rerank) \cite{bow} & FIRe \cite{fire} & DELG (+ rerank) \cite{delg} & CLIP \cite{clip} & SLIP \cite{slip} \\
        \hline
        Index file size & 22.141 GB              & 0.563 GB & 15.222 GB & 0.082 GB & 0.107 GB\\
        \hline
        \hline
        Method        & BLIP \cite{blip} & BLIP2 \cite{blip2} & DINO \cite{dino} & DINOv2 \cite{dinov2} & DiHT \cite{diht} \\
        \hline
        Index file size & ~~0.214 GB & \textbf{0.054 GB}  & ~~0.082 GB & 0.214 GB & 0.160 GB\\
        \hline
      \end{tabular}
    }
\end{table}

\subsection{Data Format}

The query and database images are available in JPEG format, which can be easily read by many existing (Python) libraries. It is worth mentioning that the original images on the Internet are not necessarily in JPEG format. We standardize them to JPEG format with a script, which is also available on our GitHub repository. For metadata, including annotations and lists of images, they are organized in newline delimited JSON files, which can be loaded with (Python) json library. 

\subsection{Licensing and Maintenance Schedule}

The dataset link and downloaders of FORB are maintained by the authors on GitHub. 
In particular, the data (including images and metadata) are accessible under the CC BY-NC-SA license. All the supporting code is available on the same GitHub repository, licensed under the MIT license. 
Any issues or discussions regarding technical or other concerns can be submitted to the GitHub repository, and the authors will reply as soon as possible. Community forks and pull requests will be welcome and reviewed by the repository maintainers.

Our FORB benchmark is a growing project. In the future we expect to include more object types as well as to increase the quantities of both query and database images. New versions of FORB dataset will be shared and announced on the GitHub page (https://github.com/pxiangwu/FORB/). We maintain the history of versions and will provide the download link to each version. Finally, we expect to also include new emerging baseline methods to establish up-to-date benchmark results.

\subsection{Author Statement}
In accordance with the CC BY-NC-SA and MIT license, the authors bear all responsibility in case of violation of rights. The descriptions made in the paper and its supplementary material are accurate and agreed upon by all authors.

\bibliographystyle{plain}
\bibliography{appendix}